\documentclass[journal]{IEEEtran}
\ifCLASSINFOpdf
\else
\fi

\usepackage[utf8]{inputenc} 
\usepackage[T1]{fontenc}    
\usepackage{hyperref}       
\usepackage{url}            
\usepackage{booktabs}       
\usepackage{threeparttable}
\usepackage{amsfonts}       
\usepackage{nicefrac}       
\usepackage{microtype}      
\usepackage[small]{caption}
\usepackage[pdftex]{graphicx}
\usepackage{amsmath}
\usepackage{algorithm}
\usepackage{algorithmic}
\usepackage{xcolor}
\usepackage{soul}
\usepackage{amssymb}
\usepackage{cases}
\usepackage{comment}
\usepackage{latexsym}
\usepackage{multirow}
\usepackage{subfigure}
\usepackage{wrapfig}
\usepackage{amsthm,amsmath,amssymb}
\usepackage{times}
\usepackage[marginal]{footmisc}
\usepackage[figuresright]{rotating}
\usepackage{multirow}
\usepackage{mathrsfs}
\usepackage[colorinlistoftodos]{todonotes}

\newcommand{\tabincell}[2]{\begin{tabular}{@{}#1@{}}#2\end{tabular}}

\begin{document}
%
\title{Exploration in Deep Reinforcement Learning: From Single-Agent to Multi-Agent Domain}
%
%
%

\author{
Jianye~Hao~\IEEEmembership{Member,~IEEE},
Tianpei~Yang,
Hongyao~Tang,
Chenjia~Bai,
Jinyi~Liu,
Zhaopeng~Meng,
Peng~Liu,~and~Zhen~Wang~\IEEEmembership{Senior~Member,~IEEE}
\thanks{Manuscript received February 20, 2022; revised June 8, 2022 and October 23, 2022; accepted January 9, 2023. This work was supported in part by the National Science Fund for Distinguished Young Scholars (Grants No. 62025602); in part by the National Natural Science Foundation of China (Grants No. U22B2036, 11931015, U1836214); in part by the New Generation of Artificial Intelligence Science and Technology Major Project of Tianjin (Grants No. 19ZXZNGX00010); and in part by the Tencent Foundation and XPLORER PRIZE. \textsl{(Corresponding author: Zhen Wang)}.}
\thanks{J. Hao, T. Yang, H. Tang, J. Liu, and Z. Meng are with the College of Intelligence and Computing, Tianjin University, Tianjin 300350, China. (e-mail: \{jianye.hao, tpyang, bluecontra, jyliu, mengzp\}@tju.edu.cn).\\ 
C. Bai is with Shanghai Artificial Intelligence Laboratory, Shanghai 200232, China (e-mail: baichenjia@pjlab.org.cn). \\
P. Liu is with the School of Computer Science and Technology, Harbin Institute of Technology, Harbin 150001, China. (e-mail: pengliu@hit.edu.cn). \\
Zhen Wang is with School of Artificial Intelligence, OPtics and Electronics (iOPEN) \& School of Cyberspace, Northwestern Polytechnical University, Xi'an 710072, China. (e-mail: w-zhen@nwpu.edu.cn).}
}

%
%

\markboth{Exploration in Deep Reinforcement Learning: From Single-Agent to Multi-Agent Domain}%
{Hao \MakeLowercase{\textit{et al.}}: Exploration in DRL: From Single-Agent to Multi-Agent Domain}
%



\maketitle
\begin{abstract}
Deep Reinforcement Learning (DRL) and Deep Multi-agent Reinforcement Learning (MARL) have achieved significant success across a wide range of domains, 
including game AI, autonomous vehicles, robotics 
and so on. 
However, DRL and deep MARL agents are widely known to be sample inefficient that millions of interactions are usually needed
even for relatively simple problem settings, 
thus preventing the wide application and deployment in real-industry scenarios. 
One bottleneck challenge behind is the well-known exploration problem, 
i.e., how efficiently exploring the environment and collecting informative experiences that could benefit policy learning towards the optimal ones. 
This problem becomes more challenging in complex environments with sparse rewards, noisy distractions, long horizons, and non-stationary co-learners. In this paper, we conduct a comprehensive survey on existing exploration methods for both single-agent and multi-agent RL.
We start the survey by identifying several key challenges to efficient exploration.
Then we provide a systematic survey of existing approaches by classifying them into two major categories: uncertainty-oriented exploration and intrinsic motivation-oriented exploration.
Beyond the above two main branches, we also include other notable exploration methods with different ideas and techniques.
In addition to algorithmic analysis, we provide a comprehensive and unified empirical comparison of different exploration methods for DRL on a set of commonly used benchmarks.
According to our algorithmic and empirical investigation, we finally summarize the open problems of exploration in DRL and deep MARL and point out a few future directions.
\end{abstract}

\begin{IEEEkeywords}
Deep Reinforcement Learning, Multi-Agent Systems, Exploration, Uncertainty, Intrinsic Motivation.
\end{IEEEkeywords}

%
\IEEEpeerreviewmaketitle

%
\IEEEpeerreviewmaketitle

\section{Introduction}\label{sec1}
In recent years, Deep Reinforcement Learning (DRL) and deep Multi-agent Reinforcement Learning (MARL) have achieved huge success in a wide range of domains, including Go  \cite{alphaGo-2016}, 
Atari \cite{Badia20Agent57}, StarCraft  \cite{RashidSWFFW18Qmix}, 
robotics \cite{LillicrapHPHETS15DDPG}, and other control problems \cite{PathakAED17,tnnls-9410239,KongHYLK21}. This reveals the tremendous potential of DRL to solve real-world sequential decision-making problems. 
Despite the successes in many domains,
there is still a long way to apply DRL and deep MARL in real-world problems because of the sample inefficiency issue, which requires millions of interactions even for some relatively simple problem settings.
For example, Agent57  \cite{Badia20Agent57} is the first DRL algorithm that outperforms the human average performance on all 57 Atari games;
but the number of interactions it needs is several orders of magnitude larger than that of humans needed.
The issue of sample inefficiency naturally becomes more severe in multi-agent settings 
since the state-action space grows exponentially with the number of agents involved.
One bottleneck challenge behind is the exploration, 
i.e., how to efficiently explore the unknown environment and collect informative experiences that could benefit the policy learning most.
This can be even more challenging in complex environments with 
sparse rewards, noisy distractions, long horizons, and non-stationary co-learners.
Therefore, how to efficiently explore the environment is significant and fundamental for DRL and deep MARL.

In recent years, considerable progress has been made in exploration from different perspectives. 
However, 
a comprehensive survey on exploration in DRL and deep MARL is currently missing.
A few prior papers contain the investigation on exploration methods for DRL  \cite{survey-drl-1,survey-drl-2,survey-drl-3,survey-drl-4,phdthesis:Dann2020StrategicEI} and MARL  \cite{survey-madrl-1,survey-madrl-2,survey-madrl-3}.
However, these works focus on general topics of DRL and deep MARL,
thus lack of systematic literature review of exploration methods and in-depth analysis.
Aubret et al.~ \cite{aubret2019survey} conduct a survey on intrinsic motivation in DRL, investigating how to leverage this idea to learn complicated and generalizable behaviors
for problems like exploration  \cite{BellemareSOSSM16}, hierarchical learning
\cite{tnnls-9119863},
skill discovery  \cite{tnnls-9205265} and so on.
Since exploration is not their focus, the ability of intrinsic motivation in addressing different exploration problems is not well studied. 
Besides, several works study the exploration methods only for multi-arm bandits  \cite{bandit-book-2020},
while they
are often incompatible with deep neural networks directly and have the issue of scalability in complex problems.
\cite{ladosz2022exploration} is a concurrent work. This work only considers single-agent exploration while we give a unified view of single-agent and multi-agent exploration.
Due to the lack of a comprehensive survey on exploration,
the advantages and limitations of existing exploration methods are seldom compared and studied in a unified scheme, which prevents further understanding of the exploration problem in the RL community.

\begin{figure}[t]
    \centering
    \includegraphics[width=1.0\linewidth]{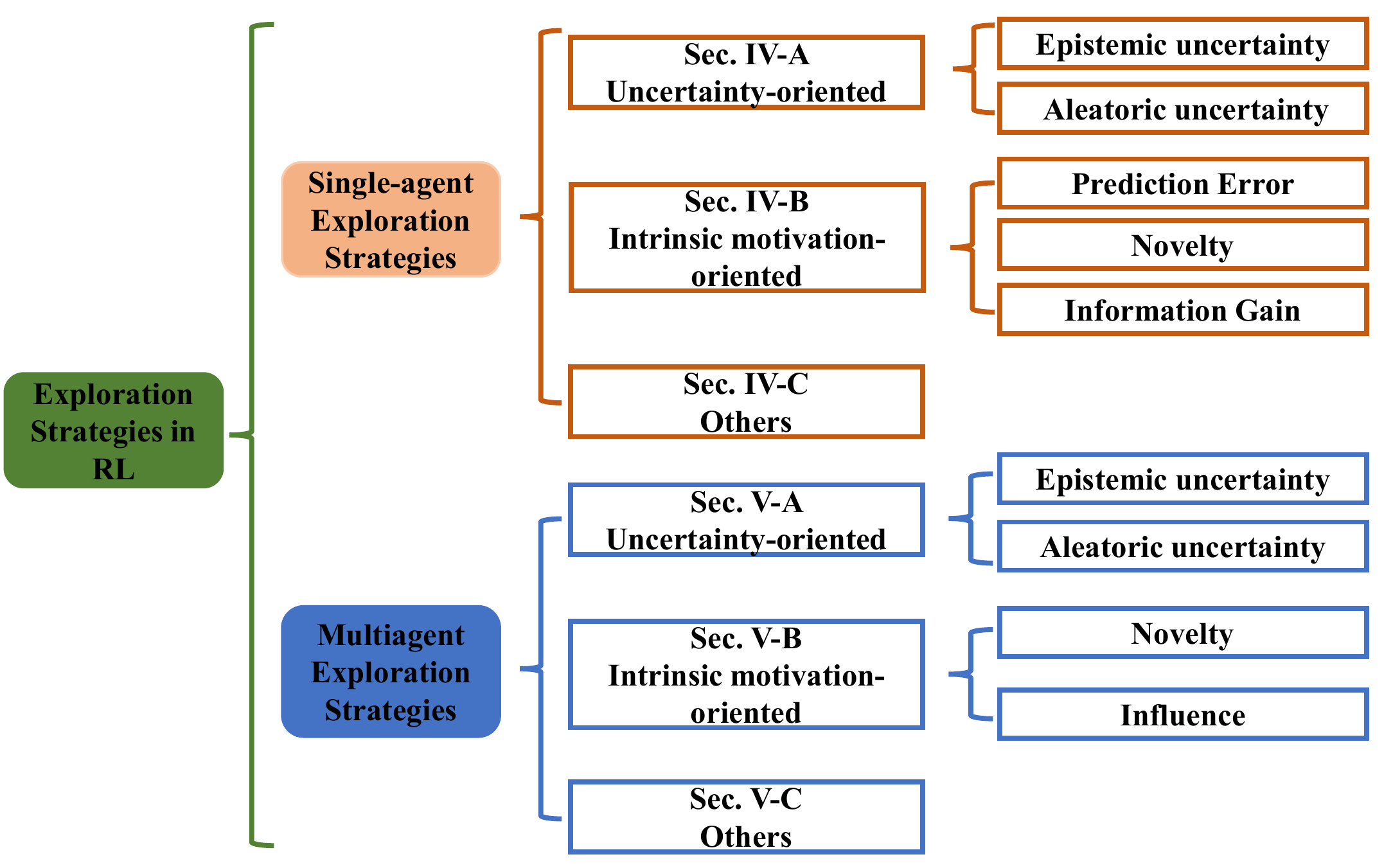}
    \caption{Illustration of our taxonomy of the current literature on methods for DRL and deep MARL.
    }
    \label{fig:framework}
\vspace{-2em}
\end{figure}

In this paper,
we propose a comprehensive survey of exploration methods in DRL and deep MARL, analyzing these methods in a unified view from both algorithmic and experimental perspectives.
We focus on the model-free setting where the agent learns its policy without access to the environment model. 
The general principles of exploration studied in the model-free setting are also shared by model-based methods in spite of different ways to realize.
We start our survey by identifying the key challenges to achieving efficient exploration in DRL and deep MARL.
The abilities in addressing these challenges also serve as criteria when we analyze and compare the existing exploration methods.
The overall taxonomy of this survey is given in Fig.~\ref{fig:framework}.
For both exploration in DRL and deep MARL, 
we classify the existing exploration methods into two major categories based on their core ideas and algorithmic characteristics.
The first category is uncertainty-oriented exploration that originates from the principle of \emph{Optimism in the Face of Uncertainty} (OFU).
The essence of this category is to leverage the quantification of epistemic and aleatoric uncertainty to measure the sufficiency of learning and the intrinsic stochasticity, 
based on which efficient exploration can be derived.
The second category is intrinsic motivation-oriented exploration.
In developmental psychology, intrinsic motivation is regarded as the primary driver in the early stages of human development  \cite{ryan2000intrinsic,barto2013intrinsic},
e.g., children often employ less goal-oriented exploration but use curiosity to gain knowledge about the world.
Taking such inspiration, this kind of method heuristically
makes use of various reward-agnostic information as the intrinsic motivation of exploration.
Note that to some degree, methods in one of the above two categories may have underlying connections to some methods in the other one, usually from specifically intuitive or theoretical perspectives.
In our taxonomy, we classify each method by sticking to its origination in motivation and algorithm as we describe above, as well as referring to the conventional literature it anchors.
Beyond the above two main streams, we also include a few other advanced exploration methods, which show potential in solving hard-exploration tasks.

In addition to analysis and comparison from the algorithmic perspective,
we provide a unified empirical evaluation of representative DRL exploration methods among several typical exploration environments,
in terms of cumulative rewards and sample efficiency.
The benchmarks demonstrate the successes and failures of compared methods, showing the efficacy of corresponding algorithmic techniques.
Uncertainty-oriented exploration methods show general improvements on
exploration in most environments.
Nevertheless, it is nontrivial to estimate the uncertainty with high quality in complex environments. 
In contrast, intrinsic motivation can significantly boost exploration in environments with sparse, delayed rewards, 
but may also cause deterioration in conventional environments due to the deviation of learning objectives.
At present, the research on exploration in deep MARL is at an early stage.
Most current methods for deep MARL exploration share similar ideas of single-agent exploration and extend the techniques.
In addition to the issues in the single-agent setting,
these methods face difficulties in addressing the larger joint state-action space, the inconsistency of individual exploration behaviors, etc.
Besides, the lack of common benchmarks prevents a unified empirical comparison.
Finally, we 
highlight several significant open questions of exploration in DRL and deep MARL,
followed by
some potential future directions.

We summarize our main contributions as follows:
\begin{itemize}
\item We give a comprehensive survey on exploration in DRL and deep MARL 
with a novel taxonomy for the first time.

\item We analyze the strengths and weaknesses of representative articles on exploration for DRL and deep MARL, along with their abilities in addressing different challenges.
   
\item We provide a unified empirical comparison of representative exploration methods in DRL among several typical exploration benchmarks.

\item We highlight existing exploration challenges, open problems, and future directions for DRL and deep MARL.
\end{itemize}

The following of this paper is organized as follows. 
Sec.~\ref{sec2} describes preliminaries on basic RL algorithms and exploration methods. 
Then we introduce the challenges for exploration in Sec.~\ref{sec3}. 
According to our taxonomy, we present the exploration methods for DRL and deep MARL in Sec.~\ref{sec4} and Sec.~\ref{sec5}, respectively. 
In Sec.~\ref{sec6}, we provide a unified empirical analysis of different exploration methods in commonly adopted exploration environments;
moreover, we discuss several open problems in this field and several promising directions for future research.
Finally, Sec.~\ref{sec7} gives the conclusion. 

\section{Preliminaries}\label{sec2}


\subsection{Markov Decision Process and Markov Game}

\textbf{Markov Decision Process (MDP).}
An MDP is generally defined as 
$\left< S, A, T, R, \rho_0, \gamma \right>$, 
with a set of states $S$, a set of actions $A$, a stochastic transition function $T: S \times A \rightarrow P(S)$, which represents the probability distribution over possible next states, given the current state and action,
a reward function $R: S \times A \rightarrow \mathbb{R}$, 
an initial state distribution $\rho_0: S \rightarrow \mathbb{R}_{\in [0,1]}$, 
and a discounted factor $\gamma \in [0,1)$. 
An agent interacts with the environment 
by performing its policy $\pi: S \rightarrow P(A)$, receiving a reward $r$.
The agent's objective is to 
maximize the expected cumulative discounted reward:
$J(\pi) = \mathbb{E}_{\rho_0, \pi, T} \left[\sum_{t=0}^{\infty}\gamma^{t} r_t \right]$.

\textbf{Markov Game (MG).}
MG is a multi-agent extension of MDP, which is defined as $\langle S, N, \{A^i\}_{i=1}^N,T, \{R^i\}_{i=1}^N, \rho_0, \gamma, Z, $\\ $\{O^i\}_{i=1}^N \rangle$, 
additionaly with action sets for each of $N$ agents, $A^{1},...,A^{N}$, a state transition function, 
$T: S \times A^{1} \times ... \times A^{N} \rightarrow P(S)$, 
a reward function for each agent
$R^{i}:S \times A^{1} \times ... \times A^{N} \rightarrow \mathbb{R}$.
For partially observable Markov games, each agent $i$ receives a local observation $o^{i}: Z(S, i) \rightarrow O^{i}$
and interacts with environment with its policy $\pi^{i}: O^{i} \rightarrow P(A^{i})$. 
The goal of each agent is to learn a policy that maximizes its expected discounted return,
i.e., $J^{i}(\pi^{i})=\mathbb{E}_{\rho_0,\pi^{1},...,\pi^{N}, T} \left[ \sum_{t=0}^{\infty}\gamma^{t}r^{i}_{t} \right]$,
where $r^i_t = R^i(s_{t},a^{1}_{t},...,a^{N}_{t})$.

Real-world problems formalized by MDPs and MGs can have immense state(-action) spaces.
Highly efficient exploration is indispensable to realizing effective decision-making.

\subsection{Reinforcement Learning}

RL is a learning paradigm of learning from interactions with the environment  \cite{SuttonB98RLAI}.
One central notion of RL is value function, which defines the expected return
obtained by following a policy.
Formally,
\emph{value function} $v^{\pi}(s)$ and \emph{action-value function} $q^{\pi}(s,a)$ are defined as, $v^{\pi}(s) =  \mathbb{E}_{\pi}  \left[\sum_{t=0}^{\infty} \gamma^{t} r_{t+1}|s_0=s \right]$
and
$q^{\pi}(s,a) = \mathbb{E}_{\pi} \left[\sum_{t=0}^{\infty}\gamma^{t} r_{t+1}|s_0=s, a_0=a \right]$.
Based on the above definition, most RL algorithms can be divided into two categories: value-based methods and policy-based methods.
Value-based methods generally learn the value functions from which the policies are derived implicitly.
In contrast, policy-based methods maintain explicit policies and optimize to maximize the RL objective $J(\pi)$.
Value-based methods are well suited to off-policy learning and usually performant in discrete action space;
while policy-based methods are capable of both discrete and continuous control and often equipped with appealing performance guarantees.

\textbf{Value-based Methods.} 
Deep $Q$-Network (DQN)  \cite{DBLP:journals/nature/MnihKSRVBGRFOPB15} is the most representative algorithm in DRL that derived from $Q$-learning.
The $Q$-function $Q(s,a;\theta)$ parameterized with $\theta$ is learned by minimizing Temporal Difference (TD) loss: 
$\mathcal{L}^{\rm DQN}(\theta)  =  \mathbb{E}_{(s_t,a_t,r_t,s_{t+1}^{\prime}) \sim D}  \left[y - Q(s_t,a_t;\theta) \right]^2$, 
where $y=r_t + \gamma \max_{a^{\prime}} Q(s_{t+1}, a^{\prime}; \theta^{-})$ is the target value, 
$D$ is the replay buffer and $\theta^{-}$ denotes the parameters of the target network.
Further, a variety of variants are proposed to improve the learning performance of DQN,
from different perspectives such as addressing the approximation error  \cite{HasseltGS16DDQN}, 
modeling value functions in distributions  \cite{BellemareDM17C51,DabneyRBM18QRDQN}, 
and other advanced structures and training techniques  \cite{WangSHHLF16Dueling,SchaulQAS15PER}.

\textbf{Policy Gradients Methods.} 
Policy-based methods optimize a parameterized policy $\pi_{\phi}$
by performing gradient ascent 
with regard to policy parameters $\phi$.
According to the \emph{Policy Gradients Theorem}  \cite{SuttonB98RLAI}, $\phi$ is updated as below:
\begin{equation}
\begin{aligned}
    \nabla_{\phi} J(\phi) & =  \mathbb{E}_{\pi_{\phi}} \left[ \nabla_{\phi} \log \pi_{\phi}(a_t|s_t) q^{\pi_{\phi}}(s_t,a_t) \right].
\end{aligned}
\label{eqation:REINFORCE_PG}
\end{equation}
One typical policy gradient algorithm is REINFORCE  \cite{Williams92REINFORCE} that uses the complete return $G_t=\sum_{i=0}^{\infty} \gamma^{i}r_{t+i}$ as estimates $\hat{Q}(s_t,a_t)$ of $q^{\pi_{\phi}}(s,a)$, i.e., Monte Carlo value estimation.


\textbf{Actor-Critic Methods.}
Actor-Critic methods typically approximate value functions and estimate $q^{\pi_{\phi}}(s,a)$ in Eq.~\eqref{eqation:REINFORCE_PG} with bootstrapping  \cite{SuttonB98RLAI}, conventionally $\hat{Q}(s_t,a_t) = r_t + \gamma V(s_t)$. 
Apart from the stochastic policy gradients (Eq.~\eqref{eqation:REINFORCE_PG}), \emph{Deterministic Policy Gradients} (DPG)  \cite{SilverLHDWR14DPG} allows deterministic policy, formally $\mu_{\phi}$, to be updated via maximizing an approximated $Q$-function.
In DDPG  \cite{LillicrapHPHETS15DDPG}, the policy is updated as:
\begin{equation}
\begin{aligned}
    \nabla_{\phi} J(\phi) = \mathbb{E}_{\mu_{\phi}} \left[ \nabla_{\phi} \mu_{\phi}(s) \nabla_{a} Q^{\mu_{\phi}}(s,a)|_{a=\mu_{\phi}(s)}\right].
\end{aligned}
\label{eqation:DDPG}
\end{equation}

\textbf{MARL Algorithms.}
In MARL, agents learn their policies in a shared environment whose dynamics are influenced by all agents. 
The most straightforward way is to let each agent learn a decentralized policy independently, treating other agents as part of the environment, i.e., \emph{Independent Learning} (IL);
while another way is to learn a joint policy for all agents, i.e., \emph{Joint Learning} (JL).
However, IL and JL are known to suffer from non-stationary and poor scalability respectively.
To address the issues of IL and JL, \emph{Centralized Training and Decentralized Execution} (CTDE) becomes the most popular MARL paradigm in recent years.
CTDE allows agents to fully utilize global information (e.g., global states, other agents' actions) during training 
while only local information (e.g., local observations) is required during execution  \cite{RashidSWFFW18Qmix,LoweWTHAM17MADDPG}.
providing good training efficiency and practical execution policies.

One representative CTDE method is Multi-Agent Deep Deterministic Policy Gradient (MADDPG)  \cite{LoweWTHAM17MADDPG}.
Concretely, consider a game with $N$ agents with deterministic policies $\{\mu_i\}_{i=1}^N$ parameterized by $\{\phi_i\}_{i=1}^N$,
the deterministic policy gradient for each agent $i$ can be:
$
    \nabla_{\phi_i} J(\phi_i) = \mathbb{E}_{x, \vec{a}} \big[  \nabla_{\phi_i} \mu_{\phi_i}(o_i) 
     \nabla_{a_i} Q_{i}^{\vec{\mu}} (x, a_1,..., a_N)|_{a_i=\mu_{\phi_i}(o_i)} \big],
\label{eqation:MADDPG}
$
where $x$ is a concatenation each agent's local observation $(o_1,..., o_N)$ and $\vec{a}$, $\vec{\mu}$ denote the concatenation of corresponding variables.
Each agent maintains its own critic $Q_i$ which estimates the joint observation-action value function and uses the critic to update its decentralized policy. 

Exploration is a critical topic in RL. A longstanding and fundamental problem in RL is the \textit{exploration-exploitation dilemma}:
choose the best action to maximize rewards or acquire more information by exploring the novel states and actions.
Despite many prior efforts made on this problem, it remains non-trivial to deal with, especially in practical problems.


\subsection{Basic Exploration Techniques}
\label{sec:basicexploration}

\textbf{$\epsilon$-Greedy.} 
The most basic exploration method is
$\epsilon$-greedy:
with a probability of $1-\epsilon$, the agent chooses the action greedily (i.e., exploitation); and a random choice is made otherwise (i.e., exploration).
Albeit the popularity and simplicity, $\epsilon$-greedy is inefficient in complex problems with large state-action space.

\textbf{\textit{Boltzmann} Exploration.} 
Another exploration method in RL is
\textit{Boltzmann} (\textit{softmax}) exploration: 
agent draws actions from a Boltzmann distribution over its $Q$-values.
Formally, the probability of choosing action $a$ is as,
$p(a)=\frac{e^{Q(s,a)/\tau}}{\sum_{i}e^{Q(s,a^i)/\tau}}$,
where the temperature parameter $\tau$ controls the degree of the selection strategy towards a purely random strategy, i.e., the higher value of the $\tau$, the more randomness the selection strategy. The drawback of Boltzmann exploration is that it cannot be directly applied in continuous state-action spaces.

\textbf{Upper Confidence Bounds (UCB).} 
UCB is a classic exploration method originally from 
Multi-Armed Bandits (MABs) problems  \cite{lai1985asymptotically}. 
In contrast to performing unintended and inefficient exploration as for
naive random exploration methods (e.g., $\epsilon$-greedy),
the methods among UCB family measure the potential of each action by upper confidence bound of the reward expectation. 
Formally, the selected action can be calculated as: 
$
    a = \arg \max_a Q(a) + \sqrt{{-\ln k}/{2N(a)}},
$ 
where $Q(a)$ is the reward expectation of action $a$, $N(a)$ is the count of selecting action $a$ and $k$ is a constant.


\textbf{Entropy Regularization.} 
While value-based RL methods add randomness in action selection based on $Q$-values, 
entropy regularization is widely used to promote exploration for RL algorithms  \cite{DBLP:conf/icml/MnihBMGLHSK16} with stochastic policies, by adding the policy entropy $H(\pi(a|s))$ to the objective function as a regularization term, encouraging the policy to take diverse actions.
Such a regularization may deviate from the original optimization objective.
One solution is to gradually decay the influence of entropy regularization during the learning process.

\textbf{Noise Perturbation.} 
As to deterministic policies, noise perturbation is a natural way to induce exploration.
Concretely, by adding noise perturbation sampled from a stochastic process $\mathcal{N}$ to the deterministic policy $\pi(s)$, an exploration policy $\pi'(s)$ is constructed, i.e., 
$\pi'(s)=\pi(s)+\mathcal{N}$.
The stochastic process can be selected to suit the environment,
e.g.,
an Ornstein-Uhlenbeck process is preferred in physical control problems  \cite{LoweWTHAM17MADDPG};
more generally, a standard Gaussian distribution is simply adopted in many works  \cite{FujimotoHM18TD3}.
However, such vanilla noise-based exploration is unintentional exploration which can be inefficient in complex learning tasks with exploration challenges.



\subsection{Exploration based on Bayesian Optimization}
In this section, we take a brief view with respect to Bayesian Optimization (BO)~\cite{tutorialofbo}, which could perform efficient exploration to find where the global optimum locates. The objective of BO is to solve the problem: 
$
    x^* = \arg\max_{x\in X} f(x),
$ 
where $x$ is what going to be optimized, $X$ is the candidate set, and $f$ is the objective function: $x \to \mathbb{R}$, which is always black-box.

BO strategies treat the objective function $f$ as a random function, thus a Bayesian statistic model for $f$ is the basic component of BO. 
The Bayesian statistic model is then used to construct a tractable acquisition function $\alpha(x)$, which provides the selection metric for each sample $x$. 
The acquisition function should be designed for more efficient exploration, to seek a probably better selection that has not been attempted yet. 
In the following, we mainly introduce two acquisition functions
, and explain how they facilitate efficient exploration.

\textbf{Gaussian Process-Upper Confidence Bound (GP-UCB)} models the Bayesian statistic model as Gaussian Process (GP), and uses the mean $ \mu(x)$ and standard deviation $\sigma(x) $ of the GP posterior to determine the upper confidence bound~(UCB), making decisions in such optimistic estimation~\cite{DBLP:conf/icml/SrinivasKKS10}. Acquisition function of GP-UCB is $\alpha(x) = \mu_{t-1}(x) + \beta_t^{1/2}\sigma_{t-1}(x)$, 
where the bonus $\sigma(x)$ guides optimistic exploration.

\textbf{Thompson Sampling (TS)}
\cite{thompson1933likelihood}, also known as posterior sampling or probability matching, samples a function $f$ from its posterior $P(f)$ each time, and takes actions greedily with respect to such randomly drawn belief $f$ as follows: 
$
    \alpha(x) = f(x),~s.t.~f \sim P(f).
$
TS captures the uncertainty according to the posterior and enables deep exploration. 

In DRL, the above methods take a primitive role, prompting many uncertainty-oriented exploration methods based on a Bayesian statistical model about the optimization objective, which we discuss in Sec.~\ref{sec:uncertainty-oriented exploration}.


\section{What Makes Exploration Hard in RL} 
\label{sec3}
In this section, we identify several typical environmental challenges to efficient exploration for DRL and deep MARL. We analyze the causes and characteristics of each challenge by illustrative examples 
and also point out some key factors of potential solutions. Most exploration methods are proposed to address one or some of these challenges.
Therefore, 
these challenges serve as significant touchstones 
for the analysis and evaluation in the following sections.

\label{sec:challenge}

\textbf{Large State-action Space.}
The difficulty of DRL naturally increases with the growth of the state-action space.
For example, 
real-world robots often have high-dimensional sensory inputs like images or high-frequency radar signals and have numerous degrees of freedom for delicate manipulation.
Another practical example is the recommendation system which has graph-structured data as states and a large number of discrete actions.
In general, when facing a large state-action space,
exploration can be inefficient, since it can take an unaffordable budget to well explore the state-action space.
Moreover, 
the state-action space may have complex underlying structures:
the states are not accessible at equal efforts where causal dependencies exist among states;
the actions can be combinatorial and hybrid of discrete and continuous components.
These practical problems make efficient exploration more challenging.

Most exploration methods compatible with deep neural networks are able to deal with high-dimensional state space.
With low-dimensional compact state representations learned, exploration and learning can be conducted efficiently.
Towards efficient exploration in complex state space,
sophisticated strategies may be needed to quantify the degree of familiarity of states, avoid unnecessary exploration, reach the `bottleneck' states, and so on. 
By contrast, 
the exploration issues in large action space are lack study currently.
Though several specific algorithms that are tailored for complex action structures are proposed \cite{HausknechtS15aPDDPG,ChandakTKJT19ActionRep,Li21HyAR},
how to efficiently explore in a large action space is still an open question.
A further discussion can be seen in Sec.~\ref{sec:open_problems}.

\begin{figure}[ht]
\centering
\vspace{-1em}
\subfigure[Chain MDP \cite{DBLP:conf/icml/Strens00}]{\includegraphics[width = 0.7\linewidth]{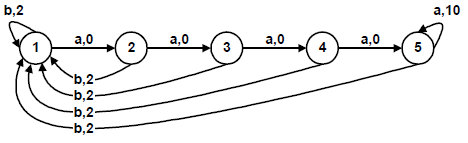}}\\
\subfigure[The first room in Montezuma's Revenge \cite{Brockman2016Gym}]{\includegraphics[width = 0.65\linewidth,height=1.4in]{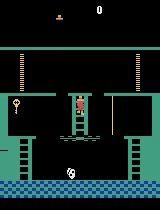}}
\caption{Examples of environments with sparse delayed rewards. (a) An illustration of the Chain MDP. The optimal behavior is to always go right to receive a reward of $+10$ when reaching the state $5$. (b) The first room in Montezuma's Revenge. An ideal agent needs to climb down the ladder, move left and collect the key where obtains a reward ($+100$);
then the agent backtracks and navigates to the door and opens it with the key,
resulting in a final reward ($+300$). }
\label{fig:first_scene_montezuma}
\vspace{-0.2cm}
\end{figure}

\textbf{Sparse, Delayed Rewards.} 
Another major challenge is to explore environments with sparse, delayed rewards. 
Basic exploration strategies
can rarely discover meaningful states or obtain informative feedback in such scenarios.
One of the most representative environments is Chain MDP  \cite{DBLP:conf/icml/Strens00,DBLP:conf/nips/OsbandAC18},
as the instance with five states and two actions (\textit{left} and \textit{right}) illustrated in Fig.~\ref{fig:first_scene_montezuma} (a).
Even with finite states and simple state representation,
Chain MDP is a classic hard exploration environment 
since the degree of reward sparsity and the difficulty of exploration increase as the chain becomes longer.
Montezuma's Revenge \cite{Brockman2016Gym} is another notorious example with sparse, delayed rewards. 
Fig. \ref{fig:first_scene_montezuma} (b) shows the first room of Montezuma's Revenge.
Without effective exploration, the agent would not be able to finish the task in such an environment.
Intuitively,
to solve such problems,
an effective exploration method is to
leverage reward-agnostic information as dense signals to explore the environment.
In addition, the ability to perform temporally extended exploration (also called temporally consistent exploration or deep exploration) is another key point in such circumstances. 
From the above two perspectives,
several recent works  \cite{BurdaESK19RND,DBLP:conf/iclr/ChoiGMOWNL19,Badia20NGU,DBLP:conf/nips/OsbandAC18}
have shown promising results, which will be discussed later.

\begin{figure}[t]
\centering
\subfigure[All 24 rooms in Montezuma's Revenge \cite{montezuma24rooms}]{\includegraphics[width = 0.97\linewidth,height=1in]{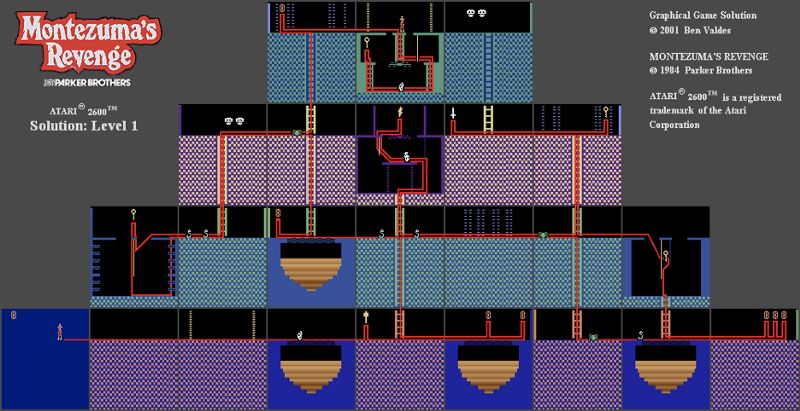}
\label{fig:overview_montezuma}}\\
\subfigure[An illustration of four domains in Minecraft \cite{TesslerGZMM17HieLifelong}]{\includegraphics[width = \linewidth,height=1.9in]{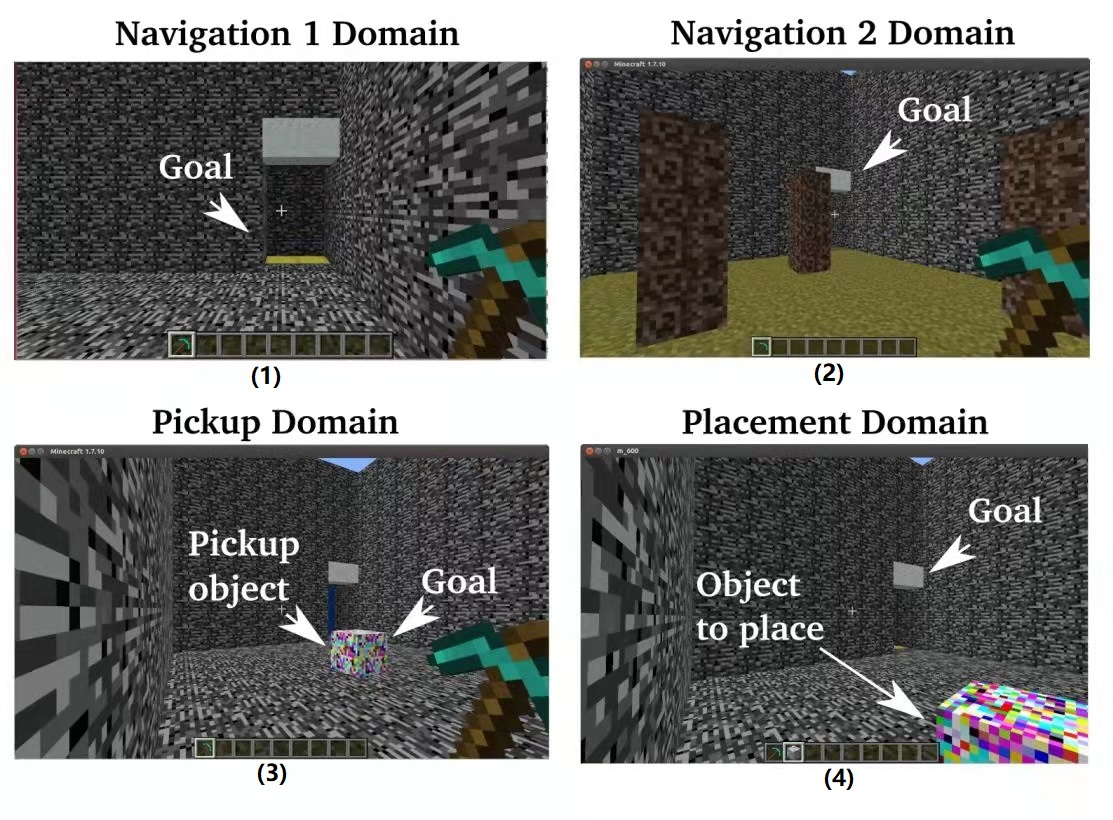}
\label{fig:overview_minecraft}}
\caption{Examples of environments with long-horizon and extremely sparse, delayed rewards. 
(a) A panorama of 24 rooms in Montezuma's Revenge. 
(b) A four-domain environment in Minecraft. Both domains contain several rooms with different objects and textures, which makes the problem have a long horizon. In each room, the agent needs to trigger a series of critical events (e.g., pickup, carry, placement), which makes the rewards of the environment sparse and delayed.}
\label{fig:extremely_sparse_envs}
\vspace{-1.5em}
\end{figure}

Although progress has been made in several representative environments,
unfortunately, 
more practical environments with long horizons and extremely sparse, delayed rewards remain far away from being solved.
Reconsider the whole game of Montezuma's Revenge and a panorama of 24 different rooms \cite{montezuma24rooms} is shown in Fig.~\ref{fig:overview_montezuma}.
The agent needs to collect different objects (e.g., keys and treasures) 
and 
solve relations among objects (e.g, the correspondence between keys and doors)
to 
trigger a series of critical events (e.g., discovering the sword).
This long-horizon task requires a more complicated policy than the one-room scene mentioned above.
Another example is the four-domain environment in Minecraft  \cite{TesslerGZMM17HieLifelong} shown in Fig.~\ref{fig:overview_minecraft}.
The agent with a first-person view is expected to traverse the four domains with different obstacles and textures, to pick up the object and place it on the target location,
then obtain the final reward. 
At present, few approaches are able to learn effective policies in such problems even with specific heuristics and prior knowledge. 
This is also regarded as a significant gap between DRL and real-world applications, which will be discussed as an open problem in Sec.~\ref{sec:open_problems}.


\begin{figure}[t]
\centering
\vspace{-0.1cm}
\includegraphics[width=\linewidth]{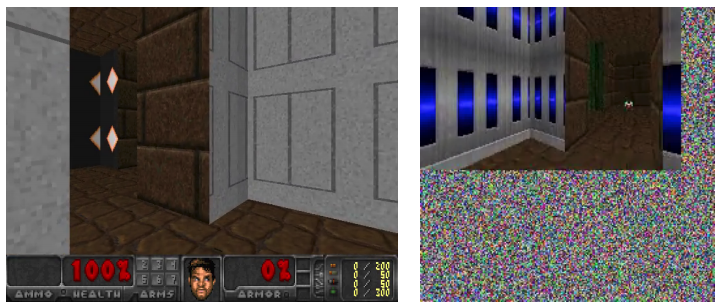}
\caption{The Noisy-TV problem \cite{DBLP:conf/cig/KempkaWRTJ16}. We show a typical VizDoom observation (left) and a noisy variant (right). The Gaussian noise is added to the observation space, which attracts the agent to stay in the current room and prevents it from passing through more rooms.}
\label{fig:noisy-tv}
\vspace{-1.5em}
\end{figure}

\textbf{White-noise Problem.}\label{sec:noise-tv} 
The real-world environments often have high randomness, where usually unpredictable things appear in the observation or action spaces. For example, visual observations of autonomous cars contain predominantly irrelevant information, like continuously changing positions and shapes of clouds. In exploration literature, white noise is often used to generate high entropy states and inject randomness into the environment. Because the white-noise is unpredictable, the agent cannot build an accurate dynamics model to predict the next state. `Noisy-TV' is a typical kind of white-noise environment. The noise of `Noisy-TV' includes constantly changing backgrounds, irrelevant objects in observations, Gaussian noise in pixel space, and so on. 
The high entropy of TV becomes an irresistible attraction to the agent. In Fig.~\ref{fig:noisy-tv}, we show a similar `Noisy-TV' in VizDoom~ \cite{DBLP:conf/cig/KempkaWRTJ16} on the right. The uncontrollable Gaussian noise is added to the observation space, which attracts the agent to stay in the current room and prevents it from passing through more rooms. Exploration methods that measure the novelty through predicting the future become unstable when confronting `Noisy-TV' or similarly unpredictable stimuli. Existing methods mainly focus on learning state representation  \cite{BurdaESK19RND,PathakAED17,BurdaEPSDE19,DBLP:conf/icml/KimNKKK19} to improve robustness when faced with the white-noise problem. How an agent can explore robustly in stochastic environments is an important research branch in DRL. We will discuss this problem further in Sec.~\ref{sec:open_problems}.

\textbf{Multi-agent Exploration.} 
Except for the above challenges, 
explorations in multi-agent settings are more arduous.  
1) Exponential increase of the state-action space. 
The straightforward challenge is that the joint state-action space increases exponentially with the increase in the number of agents, making it much more difficult to explore the environment. Therefore, the crucial point is how to carry out a comprehensive and effective exploration strategy while reducing the cost of exploration. A naive exploration strategy is that agents execute individual exploration based on their local information.

\begin{figure}[t]
\centering
\subfigure[Pass \cite{DBLP:conf/iclr/0001WWZ20}]{\includegraphics[height=1.4in,width=1.65in]{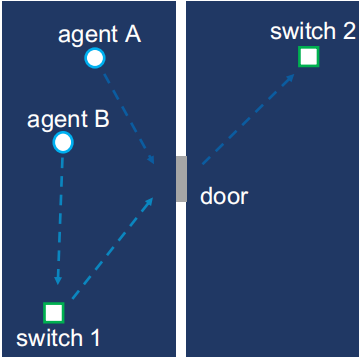}
\label{fig:pass}}
\subfigure[Cooperative Navigation]{\includegraphics[height=1.4in,width=1.65in]{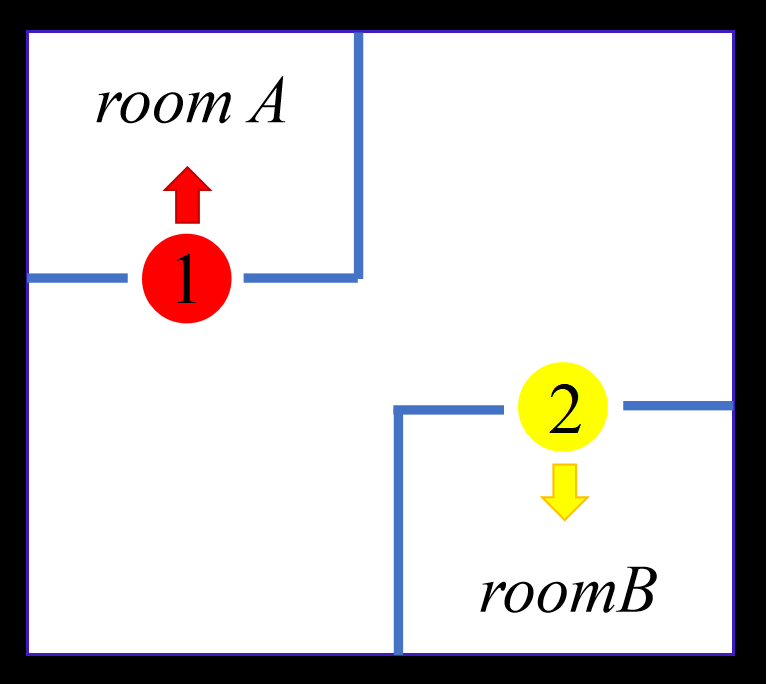}
\label{fig:navigation}}
\caption{Tasks with multi-agent exploration problem. (a) An illustration of the pass environment. Two agents need to pass the door and reach another room, while the door opens only if one agent steps at one of the switches. (b) An illustration of cooperative navigation with two homogeneous agents $1$ (red dot) and agent $2$ (yellow dot). Agents are required to search the big room with two small rooms, (i.e., room $A$ and room $B$) and then will be rewarded.}
\label{multi-agent env}
\vspace{-1em}
\end{figure}

2) Coordinated exploration. 
Although individual exploration avoids the exponential increase of the joint state-action space, it induces extra difficulties in the exploration measurement due to partial observation and non-stationary problems. The estimation based on local observation is biased and cannot reflect global information about the environment. Furthermore, an agent's behaviors are influenced by other coexisting agents, which induces extra stochasticity. Thus, such an approach always fails in tasks that need agents to behave cooperatively. 
For example, Fig.~\ref{fig:pass} illustrates 
a typical environment that requires agents to perform coordinated exploration.
Two agents are only rewarded when both of them go through the door and reach the right room; while the door will be open only when one of the two switches is covered by one agent. Thus, the optimal strategy should be as follows: one agent first steps on switch 1 to open the door, then the other agent goes to the right room to step on switch 2 to hold the door open and lets the remaining agent in. 
Challenges arise in such an environment where agents should visit some critical states in a cooperative manner, through which the optimal policy can be achieved.

3) Local- and global-exploration balance. To achieve such coordinated exploration strategies, only local information is insufficient, the global information is also necessary. 
However, one main challenge is the inconsistency between local information and global information, which requires agents to make a balance between local and global perspectives, otherwise, it may lead to inadequate or redundant exploration. 
Fig.~\ref{fig:navigation} shows such inconsistency. 
From the global perspective, either agent $1$ visits room $A$, agent $2$ visits room $B$ or agent $1$ visits room $B$, agent $2$ visits room $A$ are equally treated, which means the latter does not make an increase of the known global states. However, if agents explore relying more on local information, each agent will still try to visit another room to increase its known local states, which will cause redundant exploration from the global view. Furthermore, agents cannot simply explore by relying on global information, since how much each agent contributes to the global exploration is unknown. The trade-off of exploring locally and globally is the key problem that needs to be addressed to facilitate more efficient multi-agent exploration, which will be further discussed in Sec.~\ref{sec:open_problems}.

\section{Exploration in Single-agent DRL}\label{sec4}

We first investigate exploration works in single-agent DRL that leverage different theories and heuristics to address or alleviate the above challenges. 
Based on different key ideas and principles of these methods, we classify them into two major categories (shown in Fig.~\ref{fig:framework}). 
The first category is uncertainty-oriented exploration, which originates from the OFU principle. 
The second category is intrinsic motivation-oriented exploration, which is inspired by intrinsic motivation in psychology  \cite{ryan2000intrinsic,barto2013intrinsic} that intrinsically rewards exploration activities to encourage such behaviors. 
In addition, we also conclude other techniques beyond these two mainstreams.
\subsection{Uncertainty-oriented Exploration}
\label{sec:uncertainty-oriented exploration}
\begin{figure}[t]
\center
\includegraphics[width=1.0\linewidth]{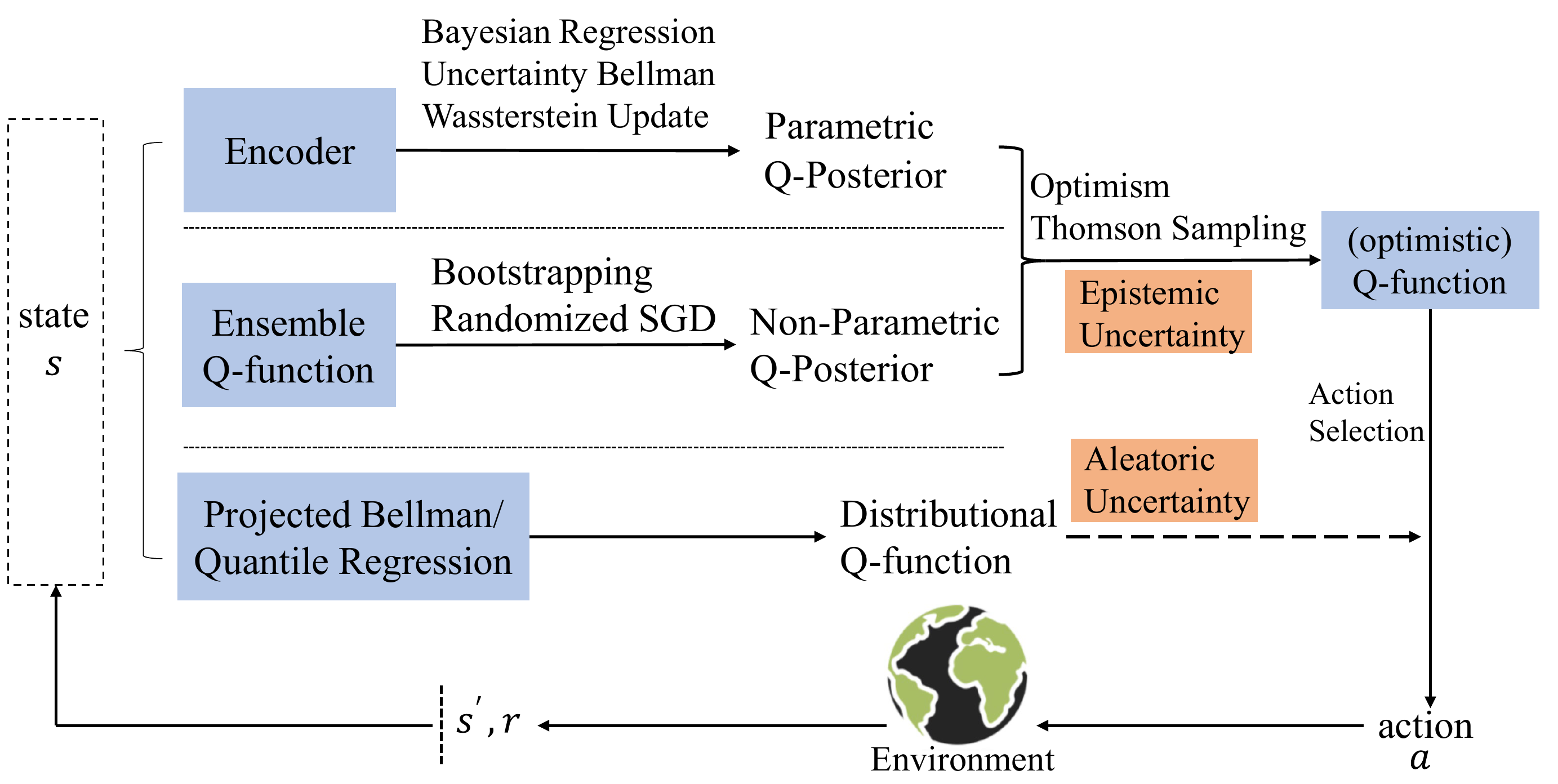}
\caption{The principle of uncertainty-oriented exploration methods.}
\label{fig:uncertainty}
\vspace{-1em}
\end{figure}

\begin{table*}[ht]
\centering
\caption{Characteristics of uncertainty-oriented exploration algorithms in RL. We use whether the algorithm uses parametric or non-parametric posterior to distinguish methods. The last two properties correspond to the challenges shown in Sec.~\ref{sec3}. We use a blank, partially, high and $\checkmark$ to indicate the extent to which the method addresses a specific problem.}
\label{table4.1}
\centering
\label{tab:uncertainty-exploration}
\begin{tabular}{p{3.0cm}|p{2.0cm}|p{3.6cm}|p{1.5cm}p{1.5cm}p{1.5cm}p{1.5cm}}

\toprule
\toprule
\rule{0pt}{12pt}
 & & Method  & Large State Space & Continuous Control & Long-horizon & White-noise\tabularnewline
\midrule
\multirow{10}{3.0cm}{Epistemic Uncertainty}
 & \multirow{5}{1.8cm}{Parametric Posterior}
 & RLSVI \cite{osband2019deep}  \cite{DBLP:conf/icml/OsbandRW16} &            &               & partially & partially \tabularnewline
 & & Bayesian DQN \cite{DBLP:conf/ita/Azizzadenesheli18}     &               &               & partially & partially \tabularnewline
 & & Successor Uncertainty  \cite{DBLP:conf/nips/JanzHMHHT19}                    & $\checkmark$  &               & partially & partially \tabularnewline
 & & Wasserstein DQN  \cite{DBLP:conf/nips/MetelliLR19}      & $\checkmark$ &                & partially & partially \tabularnewline 
 & & UBE  \cite{ODonoghueOMM18}                              & $\checkmark$ &                & partially & partially \tabularnewline \\
 & \multirow{4}{1.8cm}{Non-parametric Posterior}
 & Bootstrapped DQN \cite{DBLP:conf/nips/OsbandBPR16}        & $\checkmark$   &               & partially & partially\tabularnewline
 & & OAC \cite{DBLP:conf/nips/CiosekVLH19}                   & $\checkmark$    & \checkmark & partially                & partially \tabularnewline
 & & SUNRISE  \cite{kimin2020sunrise}                        & $\checkmark$ & $\checkmark$   & partially & partially \tabularnewline
 & & OB2I  \cite{OB2I-2021}                                  & $\checkmark$ &                & partially & partially \tabularnewline
\midrule 
\multirow{4}{3.0cm}{Epistemic \& Aleatoric Uncertainty} 
 & \multirow{4}{2.0cm}{Non-parametric Posterior}
 & DUVN \cite{DBLP:journals/corr/abs-1711-10789}             & $\checkmark$ &                & partially & partially \tabularnewline
 & & IDS \cite{DBLP:conf/colt/KirschnerK18}                  & $\checkmark$ &                &  high             & high \tabularnewline
 & & DLTV with QR-DQN  \cite{DBLP:conf/icml/MavrinYKWY19}    & $\checkmark$ &                &               & high \tabularnewline
 & & DLTV with NC-QR-DQN \cite{zhou2020non}                  & $\checkmark$  &               &               & high \tabularnewline

\bottomrule
\bottomrule
\end{tabular}
\vspace{-1em}
\end{table*}
In model-free RL, uncertainty-oriented exploration methods often measure the uncertainty of the value function. We conclude the principle of uncertainty-oriented exploration as shown in Fig.~\ref{fig:uncertainty}. 
There exist two types of uncertainty-oriented exploration. 1)~\emph{Epistemic uncertainty}, which represents the errors that arise from insufficient and inaccurate knowledge about the environment  \cite{DeardenFR98, DBLP:journals/corr/abs-1711-10789}, is also named parametric uncertainty. Many strategies provide efficient exploration by following the principle of OFU \cite{auer2002finite, DBLP:conf/colt/Azizzadenesheli16}, where higher epistemic uncertainty is considered as insufficient knowledge of the environment. 
These methods can be grouped into two categories based on the formulation of the $Q$-posterior, which is used to estimate the epistemic uncertainty. The first group models the $Q$-posterior parametrically on the encoder of the $Q$-estimator, and the second group models it in a non-parametric way by an ensemble, etc.
The second kind of uncertainty named 
2) \emph{Aleatoric uncertainty} represents the intrinsic randomness of the environment and can be captured by the return distribution estimated by distributional $Q$-function \cite{BellemareDM17C51, DBLP:journals/corr/abs-1711-10789}, which is also known as intrinsic uncertainty or return uncertainty.




Based on the uncertainty estimation, there exist mainly two ways to utilize the uncertainty. 1) Following the idea of UCB  \cite{lai1985asymptotically, DBLP:journals/jmlr/Auer02}, a direct method for exploring states and actions with high uncertainty is \textbf{performing optimistic action-selection} by choosing the action to maximize the optimistic value function $Q^+$ in each time step, as
\begin{equation}
\quad a_t = \arg\max_a\ Q^+(s_t,a),
\label{eq:action-section}
\end{equation}
where $
Q^+(s_t,a_t)={Q(s_t, a_t)+{\rm{Uncertainty}}(s_t,a_t)}
$ indicates the ordinary Q-value added by an exploration bonus based on a specific uncertainty measurement, which is usually derived by the posterior estimation of value function or dynamics  \cite{DBLP:conf/nips/OsbandRR13}. 
2) Following the idea of Thompson Sampling  \cite{thompson1933likelihood} to perform exploration. The action selection is greedy to a \textbf{sampled value function $Q_\theta$ from the $Q$-posterior}, as
\begin{equation}
a_t = \arg\max_a Q_{\theta}(s_t,a), \quad Q_{\theta} \sim {\rm{Posterior}} {(Q)},
\label{eq:sampling}
\end{equation}
that is, to first estimate the posterior distribution of Q-function through a parametric or non-parametric posterior, then sample a Q-function $Q_\theta$ from this posterior and use $Q_\theta$ for action-selection when interacting with the environment for a whole episode. Compared with UCB-based exploration, the Thompson sampling method uses the same Q-function in the whole episode rather than performing optimistic action-selection in each time step, thus the agent is enabled to perform \emph{deep} exploration and has advantages in long-horizon exploration tasks.  
Table~\ref{tab:uncertainty-exploration} shows the characteristics of uncertainty-oriented exploration in terms of whether it addresses the challenges in Sec.~\ref{sec:challenge}.

\subsubsection{Exploration via Epistemic Uncertainty}
Estimating the epistemic uncertainty usually needs to maintain a posterior of the value function. We categorize existing methods based on which type of posterior is used, including the parametric posterior and the non-parametric posterior.

\textit{Parametric Posterior-based Exploration.} Parametric posterior is typically learned by Bayesian regression in linear MDPs, where the transition and reward functions are assumed to be linear to state-action features. Randomized Least-Squares Value Iteration (RLSVI)  \cite{DBLP:conf/icml/OsbandRW16} perform Bayesian regression in linear MDPs so that it is able to sample the value function through Thompson Sampling as Eq.~\eqref{eq:sampling}. RLSVI is shown to attain a near-optimal worst-case regret bound \cite{zanette2020frequentist} in linear settings.
In DRL that uses neural networks as function approximators, Bayesian DQN  \cite{DBLP:conf/ita/Azizzadenesheli18}
extends the idea of RLSVI to generalized function approximations. 
BLR constructs the parametric posterior of the value function by approximately considering the value iteration of the last-layer Q-network as a linear MDP problem. 
Usually, BLR can only be applied for fixed inputs with linear function approximation and cannot be directly used in DRL. Bayesian DQN solves this problem by approximately considering the feature mapping before the output layer as a fixed feature vector, and then performing BLR based on this feature vector. However, in DRL tasks, the features of high-dimensional state-action pairs are trained in the learning process, which violates the hypothesis of BLR that the features are fixed and may cause unstable performance in large-scale tasks.
Based on a similar idea, Successor Uncertainty  \cite{DBLP:conf/nips/JanzHMHHT19} approximates the posterior through successor features  \cite{DBLP:journals/neco/Dayan93a, DBLP:conf/nips/BarretoDMHSSH17}. Because the successor feature contains the discounted representation in an episode, the $Q$-value is linear to the successor feature of the corresponding state-action pair. BLR can be applied to measure the posterior of the value function based on the successor representation. RLSVI, Bayesian DQN, and Successor Uncertainties address the long-horizon problem since Thomson sampling captures the long-term uncertainties.

Another approach that stands in a novel view to use epistemic uncertainty is Wasserstein Q-Learning (WQL)  \cite{DBLP:conf/nips/MetelliLR19}, which uses a Bayesian framework to model the epistemic uncertainty, and propagates the uncertainty across state-action pairs. Instead of applying a standard Bayesian update, WQL approximates the posterior distribution based on Wasserstein barycenters  \cite{DBLP:journals/siamma/AguehC11}. However, the use of the Wasserstein-Temporal-Difference update makes it computationally expensive.
Uncertainty Bellman Equation and Exploration (UBE)  \cite{ODonoghueOMM18} proposes another Bayesian posterior framework that uses an upper bound on the Q-posterior variance. 
However, without the whole posterior distribution, UBE cannot perform Thomson sampling and instead uses the posterior variance as an exploration bonus.



\textit{Non-Parametric Posterior-based Exploration.} Beyond parametric posterior, bootstrap-based exploration  \cite{DBLP:journals/corr/OsbandR15} constructs a non-parametric posterior based on the bootstrapped value functions, which has theoretical guarantees in tabular and linear MDPs. In DRL, Bootstrapped DQN  \cite{DBLP:conf/nips/OsbandBPR16} maintains several independent $Q$-estimators and randomly samples one of them, 
which enables the agent to conduct temporally-extended exploration since the agent considers long-term effects of exploration from the $Q$-function and follows the same exploratory policy in the entire episode. 
Bootstrapped DQN is similar to RLSVI, but samples value function via bootstrapping instead of Gaussian distribution. Bootstrapped DQN is easy to implement and performs well, thus becoming a common baseline for \emph{deep} exploration. Subsequently, Osband et al.  \cite{DBLP:conf/nips/OsbandAC18} show that via injecting a `prior' for each bootstrapped $Q$-function, the bootstrapped posterior can further increase the diversity of bootstrapped functions in regions with fewer observations, and thus improves the generalization of uncertainty estimation. Considering uncertainty propagation, OB2I  \cite{OB2I-2021} performs backward induction of uncertainty to capture the long-term uncertainty in an episode. 




The parametric posterior-based methods can only handle discrete control problems since the update of LSVI and other Bayesian methods require the action space to be countable. However, the non-parametric posterior  \cite{DBLP:conf/nips/OsbandAC18,DBLP:conf/nips/OsbandBPR16, OB2I-2021} based methods can be applied in continuous control to choose actions that maximize $Q_\theta(s,a)$ from the posterior.
SUNRISE  \cite{kimin2020sunrise} integrates bootstrapped sampling to provide a bonus for optimistic action selection, and additionally adopts a weighted Bellman backup to prevent instability in error propagation. Optimistic Actor Critic (OAC) \cite{DBLP:conf/nips/CiosekVLH19} builds the upper bound of $Q$-value through two bootstrapped networks and explores by choosing optimistic actions based on the upper bound. 

The above methods perform optimistic action selection or posterior sampling based solely on epistemic uncertainty. There exist several other methods  \cite{DBLP:journals/corr/abs-1711-10789, DBLP:conf/iclr/NikolovKBK19, DBLP:journals/corr/abs-1905-09638, DBLP:conf/icml/MavrinYKWY19} that consider both the epistemic uncertainty and the aleatoric uncertainty in exploration. Beyond estimating the epistemic uncertainty, additionally preserving the aleatoric uncertainty enables to prevent the agent from exploring areas with high randomness. We introduce these methods in the following.


\subsubsection{Exploration under Both Types of Uncertainty} 
For an environment with large randomness, the estimated uncertainty may be disturbed by aleatoric uncertainty, e.g., the ensemble estimators may be uncertain about a state-action pair not because that is seldom visited, but due to its large environment randomness. Meanwhile, since the aleatoric uncertainty cannot be reduced during training, being optimistic about aleatoric uncertainty may lead the agent to favor actions with higher variances, which hurts the performance. To consider both epistemic uncertainty and aleatoric uncertainty for exploration, Double Uncertain Value Network (DUVN)  \cite{DBLP:journals/corr/abs-1711-10789} firstly proposes to use Bayesian dropout  \cite{GalG16} to measure the epistemic uncertainty and return distribution to estimate the aleatoric uncertainty. However, DUVN does not tackle the negative impact of being optimistic to aleatoric uncertainty. 

Inspired by Information Directed Sampling (IDS)  \cite{DBLP:conf/colt/KirschnerK18} in bandit settings, Nikolov et al.  \cite{DBLP:conf/iclr/NikolovKBK19} extend the idea of IDS to general MDPs for efficient exploration by considering both epistemic and aleatoric uncertainties, and try to avoid the impact of aleatoric uncertainty. 
This method combines distributional RL  \cite{BellemareDM17C51} to measure aleatoric uncertainty and bootstrapped $Q$-values to approximate epistemic uncertainty. 
Then the behavior policy is designed to balance instantaneous regret and information gain.
To improve the computational efficiency of IDS,
Carmel et al.  \cite{DBLP:journals/corr/abs-1905-09638} estimate both types of uncertainty on the expected return through two networks. Specifically, aleatoric uncertainty is captured via the learned return distribution using QR-DQN  \cite{DabneyRBM18QRDQN}, and the epistemic uncertainty is estimated on the Bayesian posterior by sampling parameters from two QR-DQN networks. However, considering the two types of uncertainties need more computation caused by the use of distributional value functions. 
Without explicitly estimating epistemic uncertainty, Decaying Left Truncated Variance (DLTV)  \cite{DBLP:conf/icml/MavrinYKWY19} 
uses the variance of quantiles as bonuses and applies a decay schedule to drop the effect of aleatoric uncertainty along with the training process. NC-QR  \cite{zhou2020non} then improves DLTV through non-crossing quantile regression.

We discuss the uncertainty-oriented exploration methods in dealing with the \emph{white-noise problem} shown in Sec.~\ref{sec3} and Table~\ref{tab:uncertainty-exploration}. Theoretically, if the posterior of value function can be solved accurately (e.g., a closed-form solution), the epistemic uncertainty will be accurate and the exploration will be robust to the white-noise. However, since the Q-functions are typically estimated through parametric or non-parametric methods, exploration based on the posterior can still be affected by the randomness of the environment. The lack of prior function also leads to an inaccurate estimation of the true posterior. For methods that also estimate the aleatoric uncertainty, the exploration will be more robust since the agent can avoid exploring the noisy states, and these methods are also more suitable for solving tasks where there is randomness in the interaction process. Nevertheless, estimating the noise is unstable and computationally expensive with projected Bellman update or quantile regression.

\subsection{Intrinsic Motivation-oriented Exploration}
\begin{figure}[ht]
\vspace{-1em}
\centering
\includegraphics[width=1.0\linewidth]{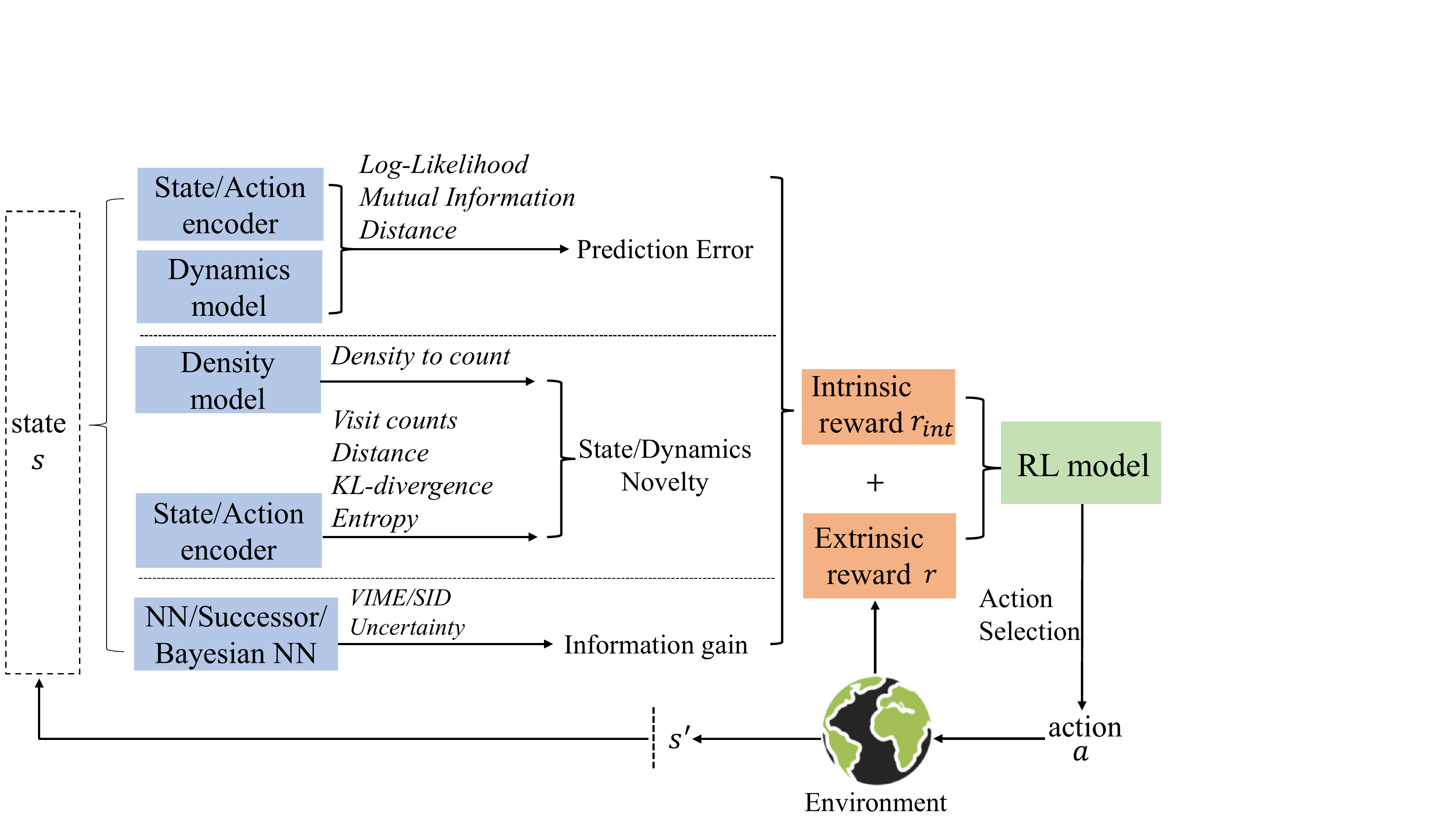}
\caption{The principle of intrinsic motivation-oriented methods.}\label{fig:intrinsic}
\end{figure}

Intrinsic motivation originates from humans' inherent tendency to interact with the world in an attempt to have an effect, and to feel a sense of accomplishment \cite{ryan2000intrinsic,barto2013intrinsic}. It is usually accompanied by positive effects (rewards), thus intrinsic motivation-oriented exploration methods often design intrinsic rewards to create a sense of accomplishment for agents. In this section, we investigate previous works tackling the exploration problem based on intrinsic motivation. These works can be technically classified into three categories (as shown in Figure~\ref{fig:intrinsic}): 1) methods that estimate prediction errors of the environmental dynamics; 2) methods that estimate the state novelty; 3) methods based on the information gain. Table \ref{table2} presents the characteristics of all reviewed intrinsic motivation-oriented exploration algorithms in terms of whether they can apply to continuous control problems, and whether they can solve the white-noise and long-horizon problems described in Section~\ref{sec:challenge}. 

\subsubsection{Prediction Error} 
The first class of works is based on prediction errors, which encourages agents to explore states with higher prediction errors. Specifically, for each state,  the intrinsic reward is designed using its prediction error for the next state, which can be measured as the distance between the predicted next state and the true one:
$
R(s_t, s_{t+1})=\text{dist}\big(\phi(s_{t+1}), \hat{f}\left( \phi(s_t),a_t\right) \big), 
$
where $\text{dist}(\cdot)$ can be any distance measurement function, $\phi$ is an encoder network that maps the raw state space to a latent space. $\hat{f}$ is a dynamics model that predicts the next latent state with the current latent state and action as input. In this direction, how to learn a suitable encoder $\phi$ is the main challenge.

\begin{table}[t]
\centering
\caption{Characteristics of reviewed intrinsic motivation-oriented exploration algorithms in RL, where blank, partially, high and $\checkmark$ denote the extent that this method addresses this problem.}\label{table2}
\begin{tabular}{p{1.3cm}|p{3.0cm}p{1.0cm}p{0.7cm}p{0.7cm}}
\toprule
\bottomrule
\rule{0pt}{12pt}
 &  Method &  Continuous Control & Long-horizon & White-noise\\ 
  \cline{1-5}
 \rule{0pt}{12pt}
\multirow{5}{1.1cm}{\tabincell{c}{Prediction \\error}} & Dynamic-AE \cite{StadieLA15}& &  &     \\
 &   ICM  \cite{PathakAED17}  & & & high  \\
  &   \tabincell{c}{Curiosity-Driven \cite{BurdaEPSDE19}}  & &   & \\
  &   VDM \cite{vdm-2020}  & \checkmark & &  high  \\
   &   AR4E \cite{OhC19}  & & &  high  \\
    &   EMI  \cite{KimKJLS19}  & $\checkmark$ & &   \\
   \cline{1-5}
 \rule{0pt}{12pt}
 \multirow{18}{1.1cm}{Novelty} &  TRPO-AE-hash \cite{TangHFSCDSTA17}  & $\checkmark$  &  &  partially\\
   &  A3C$+$ \cite{BellemareSOSSM16}  & $\checkmark$ & & partially \\
   &  DQN-PixelCNN \cite{OstrovskiBOM17} & & & partially \\
   &  $\phi$-EB \cite{MartinSEH17}  & $\checkmark$ & & partially  \\
   & VAE$+$ME \cite{abs-1905-12621} &$\checkmark$& &\\
   &  DQN+SR \cite{MachadoBB20}  & & & high \\
   &  DORA \cite{DBLP:conf/iclr/FoxCL18}  &  & & high \\
   &  A2C$+$CoEX \cite{DBLP:conf/iclr/ChoiGMOWNL19} & $\checkmark$ & &  \\
   &  RND \cite{BurdaESK19RND}  & $\checkmark$ & & partially \\
   &  Action balance RND \cite{DBLP:conf/cig/SongCHF20}  & $\checkmark$ & & partially \\
      &  MADE \cite{zhang2021made} & $\checkmark$& &  \\
   & Informed exploration \cite{DBLP:conf/nips/OhGLLS15}  &$\checkmark$ & partially&  \\
   &  $\text{EX}^2$ \cite{DBLP:conf/nips/FuCL17}  & & & high \\
   &  SFC \cite{DBLP:journals/corr/abs-1903-07400} &$\checkmark$ &partially & partially \\
   &  CB \cite{DBLP:conf/icml/KimNKKK19}  &$\checkmark$ &partially &high \\
   &  VSIMR \cite{klissarovvariational} &$\checkmark$ & & high \\
   &  SMM \cite{DBLP:journals/corr/abs-1906-05274} &$\checkmark$ & &  \\
   &  DeepCS \cite{stanton2018deep} & & &  \\
   &  Novelty Search \cite{DBLP:conf/nips/TaoFP20} & $\checkmark$& &high  \\
    & EC \cite{DBLP:conf/iclr/SavinovRVMPLG19} &$\checkmark$ & partially &high \\
   &  NGU \cite{Badia20NGU} & $\checkmark$& high& high \\
   &  NovelD \cite{zhang2021noveld} & $\checkmark$& &high  \\
   \cline{1-5}
 \rule{0pt}{12pt}
 \multirow{3}{1.1cm}{\tabincell{c}{Information\\ gain}} & VIME \cite{DBLP:conf/nips/HouthooftCCDSTA16}  &$\checkmark$  &  &high\\
   & AKL \cite{DBLP:journals/corr/AchiamS17} & $\checkmark$ &  &high\\
   & Disagreement \cite{DBLP:conf/icml/PathakG019} &$\checkmark$ &  &high\\
   & MAX \cite{DBLP:conf/icml/ShyamJG19} &$\checkmark$ &  &high\\
\bottomrule
\bottomrule
\end{tabular}
\vspace{-1em}
\end{table}

To address this problem, Dynamic Auto-Encoder (Dynamic-AE)  \cite{StadieLA15} learns an auto-encoder to compute the distance between the predicted state and the true state in the latent space. However, this approach is unable to handle the white-noise problem (Sec.~\ref{sec:challenge}) since the encoder does not remove the noisy distractions existing in the environment. Furthermore, it cannot handle the long-horizon problem since it does not consider temporally-extended information. A solution to improve the robustness of $\phi$ is Intrinsic Curiosity Module (ICM)  \cite{PathakAED17} 
which learns $\phi$ by a self-supervised inverse model using states pair $(s_t, s_{t+1})$ to predict the action $a_t$ done between them. Thus, $\phi$ ignores the uncontrollable aspects in the environment, so it can handle the white-noise problem to some degree. However, one major drawback is that it only considers the influence caused by one-step action, thus it cannot handle the long-horizon problem. 
Burda et al.  \cite{BurdaEPSDE19} propose a detailed analysis of previous works that use different latent spaces to compute the prediction error. They show that using random features is sufficient for many typical RL environments, while its generalization is worse than learned features. 
Later, VDM~\cite{vdm-2020} improves the previous method by learning the stochasticity in the dynamics through a variational dynamics model.

Instead of learning a state representation using an encoder $\phi$, there are some works  \cite{KimKJLS19,OhC19} aiming to learn both state and action representations. Similar to ICM, AR4E~\cite{OhC19} also learns the state transition model via a self-supervised reverse model. In addition, AR4E learns an action encoder that expands low-dimension actions to high-dimension action representation.
By increasing the representation power of the dynamics model, the results show improvements over ICM. EMI \cite{KimKJLS19} learns both the state and action representations $\phi_{s}(s)$ and $\phi_{a}(a)$ maximizing the Mutual Information (MI) of forward dynamics $I([\phi_{s}(s); \phi_{a}(a)]; \phi_{s}(s'))$ and inverse dynamics $I([\phi_{s}(s);\phi_{s}(s')]; \phi_{a}(a))$. Then, EMI computes the prediction error in the latent space as an intrinsic reward. Nevertheless, EMI does not address the white-noise and the long-horizon problems.


\subsubsection{Novelty} 
The second category focuses on motivating agents to visit states they visited less or have never visited.  
The first type of method estimating novelty is count-based. Formally, the intrinsic reward is set as the inverse proportion to the visit counts of state $N(s_t)$: $R(s_t)=1/N(s_t)$. It is normally hard to measure counts in a large or continuous state space. 
To address this problem, Tang et al.  \cite{TangHFSCDSTA17} propose TRPO-AE-hash, which uses SimHash 
function in the latent space of an auto-encoder.
There are other attempts being proposed to deal with the large or continuous state space, like A3C$+$  \cite{BellemareSOSSM16} and DQN-PixelCNN  \cite{OstrovskiBOM17}, which rely on density models \cite{OordKEKVG16} to compute the pseudo-count $\hat{N}(s_t)$, 
defined as the generalized visit count:
$
\hat{N}(s_t) = \rho(s_t)(1-\rho'(s_t))/(\rho'(s_t)-\rho(s_t)),
$
where $\rho(s_t)$ is the density model which produces the probability of observing $s_t$, and $\rho'(s_t)$ is the probability of observing $s_t$ after a new occurrence of $s_t$. 
Although these methods work well in environments with sparse, delayed rewards, extra computational complexity is caused by estimating the density model. In order to reduce the computational costs, $\phi$-EB  \cite{MartinSEH17} models the density in a latent space rather than the raw state space. Their results on Montezuma's revenge show significant advantages considering the decrease in computational costs. 

Other indirect count-based methods are proposed, e.g., DQN+SR  \cite{MachadoBB20} uses the norm of the successor representation  \cite{DBLP:conf/nips/BarretoDMHSSH17} as the intrinsic reward. DORA  \cite{DBLP:conf/iclr/FoxCL18} proposes a generalization of counters, called $E$-values, that can be used to evaluate the propagating exploratory value over trajectories. 
DORA does not handle the long-horizon problem since it does not consider the temporally-extended information. Choi et al. \cite{DBLP:conf/iclr/ChoiGMOWNL19} propose Attentive Dynamics Model (ADM) to discover the contingent region for state representation and exploration purposes. The intrinsic reward is defined as the visit count of the state representations consisting of the contingent region. Experiments on Montezuma's Revenge show ADM successfully extracts the location of the character and achieves a high score of 11618 combined with PPO  \cite{SchulmanWDRK17PPO}.

Random Network Distillation (RND)  \cite{BurdaESK19RND} estimates the state novelty by distilling a fixed random network into the other network. The intrinsic reward is set as the prediction error of a neural network predicting features of each state produced by the fixed random network. 
RND does not handle the long-horizon problem because it only considers the visit counts of each state and ignores the temporally-extended information. Furthermore, random features may be insufficient to represent the environment. Later, Song et al.  \cite{DBLP:conf/cig/SongCHF20} propose the action balance exploration strategy, which is based on RND and concentrates on finding unknown states. This approach aims to balance the frequency of each action selection to prevent an agent from paying too much attention to individual actions, thus encouraging the agent to visit unknown states. 
However, like RND, it also cannot address the long-horizon problem. Recently, Zhang et al.  \cite{zhang2021made} propose a new exploration method, called MADE, which maximizes the deviation of the occupancy of the policy from explored regions. This term is added as a regularizer to the original RL objective, which results in an intrinsic reward that can be incorporated to improve the performance of RL algorithms. 

Besides, another way to estimate the state novelty is to measure the distance between the current state $s_t$ and states frequently visited:
$
R(s_t) = \mathbb{E}_{s'\in \mathcal{B}}[\text{dist}(s_t;s')]
$,
where $\text{dist}(\cdot)$ can be any distance measurement and $\mathcal{B}$ is a distribution over recently visited states. Informed exploration  \cite{DBLP:conf/nips/OhGLLS15} uses a forward model to predict the action that leads the agent to the least often visited state in the last $d$ time steps. Then, an informed exploration strategy is built on $\epsilon$-greedy strategy, which has a probability of $\epsilon$ to select the action that leads to the least often visited state in the last $d$ time step, rather than random actions. Later, to improve the learning efficiency of the dynamics model, EX$^2$  \cite{DBLP:conf/nips/FuCL17} learns a classifier to differentiate each visited state from the other. Intrinsic rewards are assigned to those states that the classifier is not able to discriminate. 
Successor Feature Control (SFC)  \cite{DBLP:journals/corr/abs-1903-07400} is another kind of intrinsic reward, which takes statistics over trajectories into account, differing from previous works that use local information only to evaluate the intrinsic motivation. SFC can find the environmental bottlenecks where the dynamics change a lot and encourage the agent to go through these bottlenecks using intrinsic rewards. 

CB \cite{DBLP:conf/icml/KimNKKK19} learns compact state representation through the variational information bottleneck  \cite{DBLP:conf/iclr/AlemiFD017}. The intrinsic reward is defined as the KL-divergence between a fixed Gaussian prior and the posterior distribution of latent variables. CB ignores the task-irrelevant information and addresses the white-noise problem to a high degree. However, it requires extrinsic rewards in training. Similarly, VSIMR \cite{klissarovvariational} also adopts a KL-divergence intrinsic reward, but uses a variational auto-encoder (VAE) to learn the latent space. State Marginal Matching (SMM)  \cite{DBLP:journals/corr/abs-1906-05274} is another method using a KL-divergence intrinsic reward. It computes the KL-divergence between the state distribution derived by the policy and a target uniform distribution. Lastly, Tao et al. \cite{DBLP:conf/nips/TaoFP20} propose a new kind of intrinsic reward based on the distance between each state and its nearest neighboring states in the low dimensional feature space to solve tasks with sparse rewards. However, using low-dimensional features may cause a loss of information that is crucial for the exploration of the entire state space, which restricts its generalization to complex scenarios.

All the above methods can be classified as the inter-episode novelty, where the novelty of each state is considered from the perspective of across episodes. In contrast, the intra-episode novelty resets the state novelty at the beginning of each episode, which encourages the agent to visit more different states within an episode. Stanton and Clune  \cite{stanton2018deep} firstly distinguish the inter-episode novelty and intra-episode novelty and introduce Deep Curiosity Search (DeepCS) to improve the intra-episode exploration. The intrinsic reward is binary, which is set as 1 for unexplored states and 0 otherwise. Later, the Episodic Curiosity (EC) module  \cite{DBLP:conf/iclr/SavinovRVMPLG19} uses an episodic memory to form such a novelty bonus. 
To compute the state's intra-episode novelty, EC compares each state with states in the episodic memory and rewards states that are far from those states contained in the episodic memory. 
Besides, EC stores states with intrinsic rewards larger than a threshold, which means the stored states are more unreachable than other states, like bottleneck states. Thus, in fact, EC encourages the agent to visit bottleneck states. However, the use of episodic memory may restrict its scalability to the large state space.

Further, \emph{Never Give Up} (NGU)  \cite{Badia20NGU} proposes a new intrinsic reward mechanism that combines both inter-episode novelty and intra-episode novelty. An intra-episode intrinsic reward is constructed
by using k-nearest neighbors over the visited states stored in the episodic memory. This encourages the agent to visit as many as possible states within the current episode no matter how often the state is visited previously. The inter-episode novelty is driven by RND  \cite{BurdaESK19RND} and multiplicatively modulates the episodic similarity signal, which serves as a global measure with regard to the whole learning process.
This kind of novelty shows generalization among complex tasks and allows temporally-extended exploration. More recently, Zhang et al.  \cite{zhang2021noveld} propose a new criterion, called NovelD, which assigns intrinsic rewards to states at the boundary between already explored and unexplored regions. To avoid the agent exploiting the intrinsic reward by going back and forth between novel states and previous states, NovelD considers the episodic restriction that the agent is only rewarded for its first visit to the novel state in an episode. Their results show NovelD outperforms SOTA in Mini-Grid. 

\subsubsection{Information Gain} 
The last class of methods leads the agents toward unknown areas, as well as to prevent agents from paying much attention to stochastic areas. This is achieved by using the information gain as an intrinsic reward, which is computed based on the decrease in the uncertainty about environment dynamics  \cite{DBLP:conf/nips/IttiB05}. 
If the environment is deterministic, the transitions are predictable, so the uncertainty of dynamics can be decreased. On the contrary, when faced with stochastic environments, the agent is hardly capable to predict dynamics accurately, which implicitly restricts this kind of method. Specifically, the intrinsic reward based on information gain is defined as:
$
R(s_t,s_{t+k})=\text{Uncertainty}_{t+k}(\theta)-\text{Uncertainty}_t(\theta), 
$
where $\theta$ denotes the parameter of a dynamics model and $\text{Uncertainty}$ refers to the model uncertainty, which can be estimated in different ways as described in Sec.~\ref{sec:uncertainty-oriented exploration}.

Variational Information Maximizing Exploration (VIME)  \cite{DBLP:conf/nips/HouthooftCCDSTA16} encourages the agent to take actions that maximize the information gain about its belief of environment dynamics. 
In VIME, the dynamics are approximated using a Bayesian neural network (BNN)  \cite{DBLP:conf/nips/Graves11} and the reward is computed as the uncertainty reduction on weights of BNN. However, using a BNN as the dynamics model makes the VIME hard to apply to complex scenarios due to the high computation costs. Later, Achiam and Sastry  \cite{DBLP:journals/corr/AchiamS17} propose a more efficient way than VIME to learn the dynamics model, by replacing BNNs with a neural network followed by fully-factored Gaussian distributions. They design two kinds of rewards: the first one (NLL) is the cross-entropy, which approximates the KL-divergence of the true transition probabilities and the learned model. The second reward (AKL) is designed as the learning progress, which is the improvement of the prediction between the current time step $t$ and after $k$ improvements at $t + k$. 

Pathak et al.  \cite{DBLP:conf/icml/PathakG019} train an ensemble of dynamics models and use the mean of outputs as the final prediction. The intrinsic reward is designed as the variance over the ensemble of network output. The variance is high when dynamics models are not learned well, and low when the training process continues and all models will converge to the mean value finally, thus ignoring the noisy distractions since noises in the environment are task-irrelevant and do not affect the convergence. A similar idea is MAX \cite{DBLP:conf/icml/ShyamJG19}, while it uses the JS-divergence instead of the variance over the outputs of dynamics models. These methods handle the white-noise problem to some degree (Table~\ref{table2}) since the ensemble technique ignores the stochasticity. However, the main issue is the high computational complexity since it requires training an ensemble of models. 

\subsection{Other Advanced Methods for Exploration}
\label{subsec:other_method}

\begin{table*}[ht]
\centering
\caption{Characteristics of reviewed other advanced exploration algorithms (in Section IV.C), where blank, partially, high and $\checkmark$ denote the extent that this method addresses this problem. Note safe exploration methods are not included in this table due to their specialization in safety characteristics.}\label{tableother}
\begin{tabular}{p{2cm}|p{3cm}p{3.5cm}p{2.8cm}p{2.7cm}}
\toprule
\bottomrule
\rule{0pt}{12pt}
 &  Method &  Continuous Control & Long-horizon & White-noise\\ 
  \cline{1-5}
 \rule{0pt}{12pt}
\multirow{4}{1.1cm}{\tabincell{c}{Distributed\\ Exploration}} & Ape-X [116] &  &  &     \\
 &   R2D2 [117]  &  & &  \\
  &  NGU [57]  & \checkmark &  high  & high \\
  &   Agent57 [2]  & \checkmark & high & high  \\
   \cline{1-5}
 \rule{0pt}{12pt}
\multirow{2}{1.1cm}{\tabincell{c}{Parametric\\ Noise}} 
   &  ParamNoise [117]  & $\checkmark$ & & partially \\
   &  NoisyNet [118]  & $\checkmark$ & & partially \\
   \cline{1-5}
 \rule{0pt}{12pt}
 \multirow{3}{1.1cm}{\tabincell{c}{Others}} 
   & Go-Explore [140,141]  &$\checkmark$  & high &high\\
   & DTSIL [142] & $\checkmark$ & high &high\\
   & PotER [143] &$\checkmark$ & partially &high\\
\bottomrule
\bottomrule
\end{tabular}
\vspace{-1em}
\end{table*}

Beyond the two main streams introduced in the previous two sections, next we discuss several other branches of approaches
which cannot be classified into the previous two main streams exactly.
These methods provide different insights into how to achieve a general and effective exploration in DRL (Table \ref{tableother}).

\subsubsection{Distributed Exploration}
One straightforward idea to improve exploration is
using heterogeneous actors with diverse exploration behaviors to discover the environment.
One representative work is Ape-X  \cite{HorganQBBHHS18APEX}, in which a bunch of DQN workers perform $\epsilon$-greedy with different values of $\epsilon$ among independent environment instances in parallel.
The independent randomness of different environment instances and the distinct exploration degrees (i.e., $\epsilon$) of distributed workers allow efficient exploration of the environment regarding the wall time.
Prioritized Experience Replay (PER)  \cite{SchaulQAS15PER} is then applied to improve the learning efficiency from diverse experiences collected by different workers.
Later, R2D2 \cite{Kapturowski19R2D2} is proposed to further develop Ape-X architecture by integrating recurrent state information,
making the learning more efficient. 

Moreover, aim at solving hard-exploration games, distributed exploration is adopted to enhance the advanced exploration methods.
NGU \cite{Badia20NGU} is built on the R2D2 architecture and performs efficient exploration by incorporating both episodic novelty and life-long novelty.
Furthermore, Agent57  \cite{Badia20Agent57} improves NGU by adopting a separate parametrization of $Q$-functions and a meta-controller
for stable value approximation and 
adaptive selection of exploration policies, respectively.
Both NGU and Agent57 achieve significant improvement over most previous exploration methods in hard-exploration tasks of the Atari suite while maintaining good performance across the remaining Atari tasks.
This indicates the strong ability of advanced distributed methods in dealing with large state-action space and sparse reward.

\subsubsection{Exploration with Parametric Noise}

Another perspective to encourage exploration is to inject noise directly in the parameter space. 
Plappert et al.  \cite{PlappertHDSC0AA18ParameterSpaceNoise} propose to inject spherical Gaussian noise directly in network parameters. 
The noise scale is adaptively adjusted according to a heuristic mechanism which depends on a distance measure  
between the perturbed and non-perturbed policy.
Another similar work is NoisyNet  \cite{FortunatoAPMHOG18NoisyNet} which also injects noise in network parameters for effective exploration.
One thing that differs the most is that the noise scale (variance) in NoisyNet is learnable and updated from the RL loss function along with the other network parameters, in contrast to the heuristic adaptive mechanism in  \cite{PlappertHDSC0AA18ParameterSpaceNoise}.
Compared with naively adding noise to the outputs of the policy, parametric noise is more likely to encourage diverse and consistent exploration.
However, unlike action noise, it is difficult for parametric noise to realize specific and aimed exploration behaviors.
Therefore, parametric noise is an effective branch of methods to improve exploration in usual environments;
however, it may not do a great favor
in dealing with the exploration challenges like sparse, delayed rewards.

\subsubsection{Safe Exploration}

Another branch of exploration is safe exploration which pertains to the requirement of Safe RL.
The methods in this branch aim at exploring efficiently
while avoiding stepping into unsafe states or making dangerous behaviors.
This is especially significant to training RL agents in real-world applications since the induced hazards can be unbearable and devastating, e.g., in autonomous driving.
Two categories of Safe RL are proposed in the survey~\cite{GarciaF15SRLsurvey}: Optimization Criterion and Exploration Process.
The former category modifies the original optimization criterion of RL.
One representative criterion is the Constrained MDP (CMDP), which becomes a standard formalism in modern safe RL methods.
In this direction, Constrained Policy Optimization (CPO)~\cite{AchiamHTA17CPO} is a notable work and it solves the CMDP with a constrained form of TRPO~\cite{SchulmanLAJM15TRPO},
following which several improvements and extensions are proposed~\cite{TesslerMM19RewardCPO,StookeAA20ResponsiveSafety,BharadhwajKRLSG21Conser}.
A benchmark for safe exploration is established later~\cite{Achiam2019BenchmarkingSE} based on the formalism of CMDP.
Notably, the safety of the exploration process in this category is not explicitly addressed, since only the learning objective is modified for safe exploitation.

As to the other category of modifying the exploration process,
it is typically realized by either incorporating external knowledge (i.e., prior knowledge, demonstrations, teacher advice) or risk-directed exploration.
For prior knowledge, existing methods make use of a variety of forms, e.g., pre-trained classifiers of dangerous objects
~\cite{HuntFMHDS21VSRL}, special safety layer~\cite{Dalal2018SafeECAS}, human intervention~\cite{SaundersSSE18HumanInter}.
From another angle,
some works make use of given demonstrations and devise safe exploration methods,
e.g., by fitting a density model ~\cite{ThananjeyanBRLM20SAVED}, learning a dynamics model~\cite{ThomasLM21NearFuture}.
For risk-directed exploration,
the identification of safe or undesired states offers such useful guidance,
e.g., by maintaining a Gaussian Process~\cite{TurchettaB016SafeEGP}, obtaining from the optimal value functions of related tasks~\cite{KarimpanalR00V20TransferableDP}, or fitting a classifier of pre-designated safe and dangerous states~\cite{LiptonGLCD16CombatingIF}.
In a distinct way,
some works propose learning an explicit policy.
\cite{FatemiSSK19DeadEnd} derives a secure exploration policy by learning in a designated exploration MDP;
while \cite{Yu2022SafetyEditor} proposes a safety editor policy to correct the exploration actions.
Besides, safe control and constraint satisfaction for dynamical systems are widely studied in control theory~\cite{BajcsyBBTT19Efficient,HerbertBGT19Reach}. 
Typically, these methods require the access of regularity conditions and system dynamics,
which is not available in the setting of RL.
Although incorporating external knowledge is highly effective, such knowledge can be expensive and may not be available in general cases.

Beyond the above three branches, we introduce several other remarkable works with different exploration ideas.
Arguably, Go-Explore \cite{Ecoffet19GoExp,Ecoffet20FirstReturn} may be the most powerful method to solve Montezuma's Revenge and Pitfall, the most notorious hard-exploration problems in 57 Atari games. 
The recipe of Go-Explore is \textit{return-then-explore}: policy first arrives at the states of interest (called the \textit{go} step) and then explores from them (called the \textit{explore} step).
This significantly reduces redundant exploration among the state region already visited and encourages pursuing novel states.
The states of interest are selected heuristically (e.g., with visitation count) from a state archive that stores the visited states;
and the arrival of the selected
states can be achieved by resetting the simulation or performing an optionally trained goal-conditioned policy (the corresponding variant is called policy-based Go-Explore~\cite{Ecoffet20FirstReturn}).
After the arrival, random exploration is conducted and the archive is updated with newly visited states.
Finally, a robustification phase is carried out to learn a robust policy from high-performing trajectories.
Despite the amazing ability in solving extremely hard exploration problems,
the superiority of Go-Explore comes from sophisticated and specific designs, and it may not be a general method for other hard-exploration problems.
This is relaxed to some extent by DTSIL~\cite{GuoCMFB0L20DTSIL} which presents a similar idea to Go-Explore.
Another different work is Potentialized Experience Replay (PotER)  \cite{ZhaoDZKLX20Potential}.
PotER defines a potential energy function for each state in experience replay, 
including attractive potential energy that encourages the agent to be far away from the initial state and a repulsive one that prevents it from going towards obstacles, i.e., death states.
This allows the agent to learn from both superior and inferior experiences using intrinsic potential signals.
Although PotER also relies on task-specific designs, potential-based exploration is seldom studied in DRL, thus is worthwhile for further study.

\section{Exploration in Multi-agent DRL}\label{sec5}


\begin{table*}[ht]
\centering
\caption{Characteristics of reviewed multi-agent exploration algorithms, where blank, partially, high and $\checkmark$ denote the extent that this method addresses this problem. Note that the surveyed works on exploration in multi-agent MAB are not included for the 'Others' category.}\label{tablema}
\begin{tabular}{p{2cm}|p{3cm}p{4cm}p{3.5cm}p{3.5cm}}
\toprule
\bottomrule
\rule{0pt}{12pt}
 &  {\tabincell{c}Method} &  {\tabincell{c}Large joint state-action space} & {\tabincell{c}Coordinated exploration} & {\tabincell{c}Local-global exploration}\\ 
  \cline{1-5}
 \rule{0pt}{12pt}
\multirow{3}{1.1cm}{\tabincell{c}Uncertainty-oriented} & 
   MSQA~[145]    & partially &  &     \\
  & TS strategy~[146]   & &   & partially \\
 &  Bayes-UCB~[146]   & & & partially   \\
   \cline{1-5}
 \rule{0pt}{12pt}
 \multirow{5}{1.1cm}{\tabincell{c}Intrinsic Motivation-oriented}  & LIIR [156]& $\checkmark$&  partially&  partially   \\
 &  EDTI [159]   & $\checkmark$& partially&   \\
  &  Iqbal and Sha [153]    &partially & partially  & \\
  &  Chitnis et al. [160]  &partially &partially &    \\
  &     CMAE [154]& $\checkmark$& partially&  partially  \\
   \cline{1-5}
 \rule{0pt}{12pt}
 \multirow{4}{1.1cm}{\tabincell{c}{Others }} 
 & Seed Sampling [165] &  & partially & partially    \\
 & Seed TD [166]  & $\checkmark$ & partially & partially  \\
  &  CTEDD [167] & $\checkmark$ & high & partially \\
  &  MAVEN [168] & $\checkmark$ & high & partially   \\
\bottomrule
\bottomrule
\end{tabular}
\vspace{-1em}
\end{table*}

After investigating the exploration methods for single-agent DRL,
we move to multi-agent exploration methods.
At present, the study on exploration for deep MARL is roughly at the preliminary stage.
Most of them extend the ideas in the single-agent setting and propose different mechanisms by integrating the characteristics of deep MARL (Table \ref{tablema}).
Recall the challenges faced by MARL exploration (Sec.~\ref{sec:challenge}),
the dimensionality of joint state-action space of multiple agents
scales up the difficulty of quantifying the uncertainty and computing various forms of intrinsic motivation.
Beyond the large exploration space, multi-agent interaction is another critical characteristic
to be considered:
1) due to partial observations and non-stationary dynamics, cooperative and coordinated behaviors among agents are nontrivial to achieve;
2) multiple agents jointly influence the environmental dynamics and often share reward signals,
raising a challenge in reasoning and inferring 
the effects of joint exploration consequences;
3) more complex balance between local (individual) and global (joint) exploration and exploitation are to be dealt with;
4) 
multi-agent interactions induce mutual influence among agents, 
providing richer reward-agnostic information for potential utilization. 
In the following, we investigate MARL exploration methods by following the similar taxonomy adopted in the single-agent setting.

\subsection{Uncertainty-oriented Exploration}
\label{sec:uncertainty-oriented-exploration-mul}
In the multi-agent domain, estimating the uncertainty is difficult since the joint state-action space is significantly large. Meanwhile, quantifying the uncertainty has special difficulties due to partial observation and non-stationary problems. 
1) Each agent only draws a local observation from the joint state space. This makes the uncertainty measurement a kind of local uncertainty. Exploration based on local uncertainty is unreliable since the estimation is biased and cannot reflect the 
global information of the environment. 
2) The 
the non-stationary problem leads to a noisy local uncertainty measurement since an agent cannot obtain other agents' policies \cite{BolanderA11}. 
Both problems increase the stochasticity of the environment from the views of individual agents and make uncertainty estimation difficult.
Meanwhile, the agent should balance the local and global uncertainty to explore the novel states concerning the local information, and also avoid duplicate exploration by considering the other cooperative agents' uncertainty.


The epistemic uncertainty-based approach can be directly extended to the multi-agent problem. Following the OFU principle, Zhu et al. \cite{maspwgpzhu} propose Multi Safe Q-Agent, which formulates the posterior of the $Q$ function using the Gaussian process. 
Then, an upper bound of the $Q$ function ($Q^+$) can be obtained as Eq.~\eqref{eq:action-section}, in which the variance of the Gaussian process portrays the epistemic uncertainty. The agent follows a Boltzmann policy to explore according to $Q^+$. To overcome the non-stationary problems, they further constrain the actions to ensure low risk through the joint unsafety measurement.

There exist methods that use both epistemic uncertainty and aleatoric uncertainty in multi-agent exploration, where the use of aleatoric uncertainty models the stochasticity in the value function. Martin et al.~\cite{eeozsgzerosum} measure the aleatoric uncertainty following the idea of distributional value estimation in single-agent RL \cite{BellemareDM17C51}, and perform exploration based on both the aleatoric uncertainty and epistemic uncertainty. Specifically, they extend several single-agent exploration methods  \cite{auer2002finite,russo2018tutorial,KaufmannCG12} to zero-sum stochastic games, and find the most effective approaches among them are Thomson sampling and Bayes-UCB-based methods. The Thompson sampling-based method samples from the posterior of the value function.
The Bayes-UCB-based method extends the Bayes-UCB  \cite{KaufmannCG12} to a zero-sum stochastic form game, which samples several payoff matrices from the posterior distribution and chooses the action with the highest mean payoff quantile.
Both strategies maintain a posterior distribution to measure the epistemic uncertainty, thus they can perform \emph{deep} exploration. In addition,  LH-IRQN~\cite{LyuA20}, DFAC~\cite{DFAC}, and MCMARL~\cite{DQMIX} parameterize value function via a categorical distribution or a quantile distribution. 
Those methods consider aleatoric uncertainty only for better value estimation, nevertheless, it is more desirable to use the aleatoric uncertainty to improve the robustness in exploration by explicitly avoiding the agent overly exploring states with high aleatoric uncertainty.




\subsection{Intrinsic motivation-oriented Exploration}
Intrinsic motivation is widely used as the basis of exploration bonuses to encourage agents to explore unseen regions. Following the great success in single-agent RL, a number of works  \cite{DBLP:journals/corr/abs-1906-02138,DBLP:journals/corr/abs-1905-12127,came,DBLP:conf/nips/ZhengCWHHCFGZ21} tend to apply intrinsic motivation in the multi-agent domain. 
However, the difficulty of measuring intrinsic motivation increases exponentially with the number of agents increasing. 
Furthermore, assigning intrinsic rewards to agents has special difficulties in the multi-agent setting due to partial observation and non-stationary problems.
Similarly to the discussions in the previous subsection,
such difficulties contain the local and noisy estimation of intrinsic reward and the balance between local and global exploration.
A potential fortune is the profuse reward-agnostic information in multi-agent interactions, which can be utilized to devise diverse kinds of intrinsic rewards.
We investigate previous methods that aim to apply intrinsic motivation in multi-agent domains. Some works have addressed some of the above challenges to a certain degree, such as coordinated exploration \cite{DBLP:conf/nips/DuHFLDT19,DBLP:conf/iclr/0001WWZ20}. Nevertheless, there still exist open questions, such that how to decrease the difficulties in estimating the intrinsic motivation in large-scale multi-agent systems, and how to balance the inconsistency between local and global intrinsic motivation.

Some works assign agents extra bonuses based on novelty to encourage exploration. For example, Wendelin B{\"o}hmer et al. \cite{DBLP:journals/corr/abs-1906-02138} introduce an intrinsically rewarded centralized agent to interact with the environment and store experiences in a shared replay buffer while decentralized agents update their local policies using the experience from this buffer. They find that although only the centralized agent is rewarded, decentralized agents can still benefit from this and improve exploration efficiency. Instead of simply designing intrinsic rewards according to global states' novelty, Iqbal and Sha  \cite{DBLP:journals/corr/abs-1905-12127} define several types of intrinsic rewards by combining the decentralized curiosity of each agent. One type is selected adaptively for each episode which is controlled by a meta-policy. However, these types of intrinsic rewards are domain-specific and cannot be extended to other scenarios. 

Instead of designing intrinsic rewards based on state novelty, Learning Individual Intrinsic Reward (LIIR) is proposed
 \cite{DBLP:conf/nips/DuHFLDT19} which learns the individual intrinsic reward and uses it to update an agent's policy with the objective of maximizing the team reward. LIIR extends a similar idea in single-agent domains \cite{zheng2018learning,zheng2019can} that learns an extra proxy critic for each agent, with the input of intrinsic rewards and extrinsic rewards. The intrinsic rewards are learned by building the connection between the parameters of intrinsic rewards and the centralized critic based on the chain rule, thus achieving the objective of maximizing the team reward. Jaques et al.  \cite{DBLP:conf/icml/JaquesLHGOSLF19} define the intrinsic reward function from another perspective called "social influence", which measures the influence of one agent's actions on others' behavior.
By maximizing this function, agents are encouraged to take actions with the strongest influence on the policies of other agents, those joint actions lead agents to behave cooperatively. Instead of using intrinsic rewards as in  \cite{DBLP:conf/icml/JaquesLHGOSLF19}, Wang et al. integrate such influence as a regularizer into the learning objective \cite{DBLP:conf/iclr/0001WWZ20}. They measure the influence of one agent on other agents' transition function (EITI) and rewarding structure (EDTI) and encourage agents to visit critical states in the state-action space (Fig.~\ref{fig:pass}), through where agents can transit to potentially important unexplored regions. 

Chitnis et al.  \cite{DBLP:conf/iclr/ChitnisT0020} tackle the coordinated exploration problem from a different view by considering that the environment dynamics caused by joint actions are different from that caused by individually sequential actions. Therefore, this method incentivizes agents to take joint actions whose effects cannot be achieved via a composition of the predicted effect into individual actions executed sequentially. However, they need manually modify the environment to disable some agents, which is unrealistic for real-world scenarios. To summarize, the research of the intrinsic motivation-oriented exploration in MARL is mainly extended from single-agent domains, like state novelty estimation and so on. Meanwhile, there exist some works that leverage the mutual influence among agents to guide coordinated exploration. However, with the inconsistency between local and global information, how to obtain a robust and accurate intrinsic motivation estimation that balances the local and global information to derive coordinated exploration is a promising direction that needs to be studied. 

\subsection{Others Methods for Multi-agent Exploration}
Different from the works introduced previously which are derived from uncertainty estimation or intrinsic motivation, 
there are a few notable multi-agent exploration works 
which cannot be exactly classified into the former two categories.
We introduce these works in chronological order below. 

In the literature of multi-agent repeated matrix games and multi-agent MAB problems, 
the previous works propose exploration methods from different perspectives 
\cite{DBLP:journals/aamas/CarmelM99,DBLP:conf/atal/ChalkiadakisB03,VerbeeckNPT05ESRL,DBLP:conf/ijcai/ChakrabortyCDJ17}.
These works
are studied in environments with relatively small state-action space, which is far from the scales often considered in deep MARL.
Towards complex multi-agent environments with large state-action space,
Dimakopoulou and Van Roy  \cite{DBLP:conf/icml/DimakopoulouR18,DBLP:conf/nips/DimakopoulouOR18} identify
three essential properties for efficient coordinated exploration: 
adaptivity, commitment, and diversity.
They present the failures of straightforward extensions of single-agent posterior sampling approaches in satisfying the above properties.
For a practical method to meet these properties,
they propose a Seed Sampling
by incorporating randomized value networks for scalability.

From different angles, Chen  \cite{DBLP:conf/atal/000220} proposes a new framework to address the coordinated exploration problem under the paradigm of CTDE.
The key idea of the framework is to train a centralized policy first and then derive decentralized policies via policy distillation.
Efficient exploration and learning are conducted by optimizing a global maximum entropy RL objective with global information.
Decentralized policies distill cooperative behaviors from the centralized policy in the favor of agent-to-agent communication protocols.
Another notable exploration method that follows the CTDE paradigm is MAVEN  \cite{DBLP:conf/nips/MahajanRSW19}.
MAVEN utilizes a hierarchical policy to control a shared latent variable as a signal of a coordinated exploration mode, in which the value-based agents condition their policies.
By maximizing the mutual information between the latent variable and the induced episodic trajectory, 
MAVEN achieves diverse and temporally-extended exploration.
Both the above two methods leverage the centralized training mechanism to enable coordinated exploration,
and they demonstrate the potential of such exploration improvements in benefiting deep MARL.


\section{Discussion}
\label{sec6}
\subsection{Empirical Analysis}

\begin{table*}[th]
\centering
\caption{A benchmark of experimental results of exploration methods in DRL. The results are from those reported in their original papers.}\label{table:benchmark}
\centering
\begin{threeparttable}

\scalebox{0.85}{
\begin{tabular}{c|cccccccc}
\toprule
\bottomrule
\rule{0pt}{12pt}
 \tabincell{c}{Benchmark \\scenarios}& Method  & \tabincell{c}{Basic Algorithm}& \multicolumn{3}{c}{\tabincell{c}{Convergence Time (frames)}} & \multicolumn{3}{c}{\tabincell{c}{Convergence Return}} \\
   \cline{1-9}
 \rule{0pt}{12pt}
\multirow{16}{1.7cm}{\tabincell{c}{Montezuma's \\Revenge}}
 & Successor Uncertainty  \cite{DBLP:conf/nips/JanzHMHHT19}  & DQN  \cite{DBLP:journals/nature/MnihKSRVBGRFOPB15} & \multicolumn{3}{c}{200M} & \multicolumn{3}{c}{0} \\
 & Bootstrapped DQN  \cite{DBLP:conf/nips/OsbandBPR16}  & Double DQN  \cite{HasseltGS16DDQN} & \multicolumn{3}{c}{20M} & \multicolumn{3}{c}{100} \\
 & Randomized Prior Functions  \cite{DBLP:conf/nips/OsbandAC18}  & Bootstrapped DQN  \cite{DBLP:conf/nips/OsbandBPR16} & \multicolumn{3}{c}{200M} & \multicolumn{3}{c}{2500}  \\
 & Uncertainty Bellman Equation  \cite{ODonoghueOMM18}  & DQN  \cite{DBLP:journals/nature/MnihKSRVBGRFOPB15} & \multicolumn{3}{c}{500M} & \multicolumn{3}{c}{$\sim$2750} \\
 & IDS  \cite{DBLP:conf/iclr/NikolovKBK19}  & Bootstrapped DQN  \cite{DBLP:conf/nips/OsbandBPR16} \& C51  \cite{BellemareDM17C51} & \multicolumn{3}{c}{200M} & \multicolumn{3}{c}{0} \\
 & DLTV  \cite{DBLP:conf/icml/MavrinYKWY19}  & QR-DQN  \cite{DabneyRBM18QRDQN} & \multicolumn{3}{c}{40M}  & \multicolumn{3}{c}{187.5} \\
 &EMI  \cite{KimKJLS19} &TRPO  \cite{SchulmanLAJM15TRPO}& \multicolumn{3}{c}{50M} & \multicolumn{3}{c}{387}\\
 &A2C$+$CoEX \cite{DBLP:conf/iclr/ChoiGMOWNL19} & A2C  \cite{DBLP:conf/icml/MnihBMGLHSK16}& \multicolumn{3}{c}{400M} & \multicolumn{3}{c}{6635}\\
 &RND  \cite{BurdaESK19RND} &PPO  \cite{SchulmanWDRK17PPO}& \multicolumn{3}{c}{1.6B} & \multicolumn{3}{c}{8152}\\
  &Action balance RND  \cite{DBLP:conf/cig/SongCHF20} & RND  \cite{BurdaESK19RND}& \multicolumn{3}{c}{20M} & \multicolumn{3}{c}{4864}\\
 & PotER  \cite{ZhaoDZKLX20Potential}  & SIL  \cite{OhGSL18SIL} & \multicolumn{3}{c}{50M} & \multicolumn{3}{c}{6439} \\
 & NGU  \cite{Badia20NGU}  & R2D2  \cite{Kapturowski19R2D2} & \multicolumn{3}{c}{35B} & \multicolumn{3}{c}{10400} \\
 & Agent57  \cite{Badia20Agent57}   & NGU  \cite{Badia20NGU} & \multicolumn{3}{c}{35B} & \multicolumn{3}{c}{9300} \\
 & DTSIL \cite{GuoCMFB0L20DTSIL}   & PPO  \cite{SchulmanWDRK17PPO} + SIL~\cite{OhGSL18SIL} & \multicolumn{3}{c}{3.2B} & \multicolumn{3}{c}{22616} \\
 & Go-Explore (no domain knowledge)  \cite{Ecoffet19GoExp}   & PPO  \cite{SchulmanWDRK17PPO} + Backward Algorithm  \cite{Salimans18Backward} & \multicolumn{3}{c}{5.55B (Hypothetically over 70B)\tnote{1}} & \multicolumn{3}{c}{43763} \\
 & Go-Explore (domain knowledge)  \cite{Ecoffet19GoExp}  & PPO  \cite{SchulmanWDRK17PPO} + Backward Algorithm  \cite{Salimans18Backward} & \multicolumn{3}{c}{5.2B (Hypothetically over 150B)} & \multicolumn{3}{c}{666474}\\
 & Go-Explore (no horizon limit)\tnote{2})  \cite{Ecoffet19GoExp}  & PPO  \cite{SchulmanWDRK17PPO} + Backward Algorithm  \cite{Salimans18Backward} & \multicolumn{3}{c}{5.2B (Hypothetically over 150B)} & \multicolumn{3}{c}{18003200} \\
 & Go-Explore (best in Nature version) \cite{Ecoffet20FirstReturn} & PPO  \cite{SchulmanWDRK17PPO} + Backward Algorithm  \cite{Salimans18Backward} & \multicolumn{3}{c}{11B (Hypothetically over 300B)\tnote{1}} & \multicolumn{3}{c}{$>$40000000} \\
   \cline{1-9}
 \rule{0pt}{12pt}
 \multirow{12}{1.7cm}{\tabincell{c}{Overall \\Atari Suite}}
 & Bootstrapped DQN  \cite{DBLP:conf/nips/OsbandBPR16}  & Double DQN  \cite{HasseltGS16DDQN} & \multicolumn{3}{c}{200M (55 games)} & \multicolumn{3}{c}{553\%(mean), 139\%(median)} \\
 & Uncertainty Bellman Equation  \cite{ODonoghueOMM18} (1-step)  & DQN  \cite{DBLP:journals/nature/MnihKSRVBGRFOPB15} & \multicolumn{3}{c}{200M (57 games)} & \multicolumn{3}{c}{776\%(mean), 95\%(median)} \\
 & Uncertainty Bellman Equation  \cite{ODonoghueOMM18} (n-step)  & DQN  \cite{DBLP:journals/nature/MnihKSRVBGRFOPB15} & \multicolumn{3}{c}{200M (57 games)} & \multicolumn{3}{c}{440\%(mean), 126\%(median)} \\
 &A3C$+$  \cite{BellemareSOSSM16} & A3C \cite{DBLP:conf/icml/MnihBMGLHSK16}& \multicolumn{3}{c}{200M (57 games)} & \multicolumn{3}{c}{273\%(mean), 81\%(median)}\\
 & Noisy-Net  \cite{FortunatoAPMHOG18NoisyNet}  & DQN  \cite{DBLP:journals/nature/MnihKSRVBGRFOPB15} & \multicolumn{3}{c}{200M (55 games)} & \multicolumn{3}{c}{389\%(mean), 123\%(median)} \\
 & Noisy-Net  \cite{FortunatoAPMHOG18NoisyNet}  & Dueling DQN  \cite{WangSHHLF16Dueling} & \multicolumn{3}{c}{200M (55 games)} & \multicolumn{3}{c}{608\%(mean), 172\%(median)} \\
 & IDS  \cite{DBLP:conf/iclr/NikolovKBK19}  & Bootstrapped DQN  \cite{DBLP:conf/nips/OsbandBPR16} & \multicolumn{3}{c}{200M (55 games)} & \multicolumn{3}{c}{651\%(mean), 172\%(median)} \\
 & IDS  \cite{DBLP:conf/iclr/NikolovKBK19}  & Bootstrapped DQN  \cite{DBLP:conf/nips/OsbandBPR16} \& C51  \cite{BellemareDM17C51} & \multicolumn{3}{c}{200M (55 games)} & \multicolumn{3}{c}{1058\%(mean), 253\%(median)} \\
& Randomized Prior Functions  \cite{DBLP:conf/nips/OsbandAC18}  & Bootstrapped DQN  \cite{DBLP:conf/nips/OsbandBPR16} & \multicolumn{3}{c}{200M (55 games)} & \multicolumn{3}{c}{444\%(mean), 124\%(median)} \\
& Randomized Prior Functions  \cite{DBLP:conf/nips/OsbandAC18}  & Bootstrapped DQN  \cite{DBLP:conf/nips/OsbandBPR16}+Dueling  \cite{WangSHHLF16Dueling} & \multicolumn{3}{c}{200M (55 games)} & \multicolumn{3}{c}{608\%(mean), 172\%(median)} \\
 & NGU  \cite{Badia20NGU}  & R2D2  \cite{Kapturowski19R2D2} & \multicolumn{3}{c}{35B (57 games)} & \multicolumn{3}{c}{3421\%(mean), 1354\%(median)} \\
 & Agent57  \cite{Badia20Agent57}   & NGU  \cite{Badia20NGU} & \multicolumn{3}{c}{35B (57 games)} & \multicolumn{3}{c}{4766\%(mean), 1933\%(median)} \\
\cline{1-9}
 \rule{0pt}{12pt}
\multirow{6}{1.7cm}{\tabincell{c}{Vizdoom\\(MyWayHome)}}
& & & Dense & Sparse& Very sparse & Dense & Sparse& Very sparse\\
& ICM  \cite{PathakAED17}   &A3C  \cite{DBLP:conf/icml/MnihBMGLHSK16} &300M & 500M & 700M & 1.0 & 1.0 &0.8  \\
 & AR4E  \cite{OhC19}   & ICM  \cite{PathakAED17} &300M & 400M & 600M & 1.0 & 1.0 &1.0  \\
 & SFC  \cite{DBLP:journals/corr/abs-1903-07400}  &Ape-X DQN  \cite{HorganQBBHHS18APEX} &250M & - & - & 1.0 &-  & - \\ 
 &$\text{EX}^2$ \cite{DBLP:conf/nips/FuCL17}  &TRPO  \cite{SchulmanLAJM15TRPO} &200M & - & - & 0.8 &-  & -\\
 &  EC \cite{DBLP:conf/iclr/SavinovRVMPLG19}  &PPO  \cite{SchulmanWDRK17PPO}  &100M & 100M & 100M & 1.0 & 1.0  & 1.0\\

\cline{1-9}
 \rule{0pt}{12pt}
\multirow{6}{1.7cm}{\tabincell{c}{SMAC}} 
& & & \multicolumn{2}{c}{2S vs 3Z (easy)} & \multicolumn{2}{c}{3M (easy)}& \multicolumn{2}{c}{3S vs 5Z (hard)}\\
& LIIR  \cite{DBLP:conf/nips/DuHFLDT19}   & Actor-critic \cite{SuttonB98RLAI} &\multicolumn{2}{c}{1.0} & \multicolumn{2}{c}{0.9}& \multicolumn{2}{c}{0.9} \\
& & & \multicolumn{2}{c}{2S vs 3Z (easy)} & \multicolumn{2}{c}{6H vs 8Z (Super hard)}& \multicolumn{2}{c}{Corridor (Super hard)}\\
& MAVEN \cite{DBLP:conf/nips/MahajanRSW19} & QMIX \cite{RashidSWFFW18Qmix} &\multicolumn{2}{c}{1.0} & \multicolumn{2}{c}{0.6}& \multicolumn{2}{c}{0.8} \\ 
& & & \multicolumn{2}{c}{3M (Sparse)} & \multicolumn{2}{c}{2S vs 1Z (Sparse)}& \multicolumn{2}{c}{3S vs 5Z (Sparse)}\\
& CMAE  \cite{came}   & QMIX \cite{RashidSWFFW18Qmix} &\multicolumn{2}{c}{0.5} & \multicolumn{2}{c}{0.5}& \multicolumn{2}{c}{0.1} \\
  
 \bottomrule
\bottomrule
\end{tabular}
}
\begin{tablenotes}
\footnotesize
\item [1] The hypothetical frame denotes the number of game frames that would have been played if Go-Explore replayed trajectories instead of resetting the\\ emulator state as done in their original paper. For Go-Explore (best in Nature version), the frame number is estimated by us since it is not provided.
\item[2] Removing the maximum limit of 400,000 game frames imposed by default in OpenAI Gym, the best single run of Go-Explore with domain knowledge \\achieved a score of 18,003,200 and solved 1441 levels during 6,198,985 game frames, corresponding to 28.7 hours of gameplay (at 60 game frames per \\second, Atari's original speed) before losing all its lives.
\end{tablenotes}
\end{threeparttable}
\vspace{-1em}
\end{table*}

For a unified empirical evaluation of different exploration methods, we summarize the experimental results of some representative methods on four representative benchmarks: 
Montezuma's Revenge, the overall Atari suite, Vizdoom, SMAC, 
which are almost the most popular benchmark environments used by exploration methods. Each of them has different characteristics and evaluation focus on different exploration challenges. Minecraft \cite{TesslerGZMM17HieLifelong} also has been used in several works \cite{TesslerGZMM17HieLifelong,trott2019keeping}. However, it is not commonly used by exploration methods, thus we omit it in this section.

Recall the introduction in Sec.~\ref{sec:challenge}, Montezuma's Revenge is notorious due to its sparse, delayed rewards and long horizon. 
On the contrary,
the overall Atari suite focuses on a more general evaluation of exploration methods in improving the learning performance of RL agents.
Vizdoom is another representative task with multiple reward configurations (from dense to very sparse).
Distinct from the previous two tasks, Vizdoom is a navigation (and shooting) game with a first-person view.
This simulates a learning environment with severe partial observability and underlying spatial structure,
which is more similar to real-world ones faced by humans. SMAC \cite{RashidSWFFW18Qmix} is the representative MARL benchmark with a decentralized multiagent control in which each learning agent controls an individual army entity. SMAC has a lot of scenarios from easy to very hard with dense and sparse reward configurations.
We collect the reported results on the four tasks from the original papers.
Thus, the exploration methods without such results are not included.

The results are shown in Table \ref{table:benchmark}, in terms of the exploration method, the basic RL algorithm, the convergence time, and return.
The table provides a quick glimpse into the performance comparison:
1) For Montezuma's Revenge, Go-Explore \cite{Ecoffet19GoExp,Ecoffet20FirstReturn}, DTSIL~\cite{GuoCMFB0L20DTSIL} and NGU  \cite{Badia20NGU} outperform human-level performance (4756 in average) by large margins.
Especially, Go-Explore achieves the best results that exceed the human world record of 1.2 million.
The obvious drawback is the convergence time, as the requirement of billions of frames is extremely far away from sample efficiency.
2) For the overall Atari suite with 200M training frames, 
IDS  \cite{DBLP:conf/iclr/NikolovKBK19} achieves the best results and outperforms its basic methods (i.e., Bootstrapped DQN) significantly, 
showing the success of the sophisticated utilization of uncertainty
in improving exploration and learning generally.
Armed with a more advanced distributed architecture like R2D2  \cite{Kapturowski19R2D2}, Agent57  \cite{Badia20Agent57} and NGU  \cite{Badia20NGU} achieve extremely high scores in 35B training frames.
3) For Vizdoom (MyWayHome), EC  \cite{DBLP:conf/iclr/SavinovRVMPLG19} outperforms other exploration methods especially in achieving higher sample efficiency across all reward settings.
This demonstrates the effectiveness of intrinsic motivation for another time and also reveals the potential of episodic memory in dealing with partial observability and spatial structure of navigation tasks. We conclude the results as follows:

(1) Results
    show that uncertainty-oriented methods achieve better results on the overall Atari suite,
    demonstrating the effectiveness in general cases.
    Meanwhile, their performance in Montezuma's Revenge is generally lower than intrinsic motivation-oriented methods. 
    This is because uncertainty quantification is mainly based on well-learned value functions,
    which are hard to learn if the extrinsic rewards are almost absent.
    In principle, the effects of uncertainty-oriented exploration methods heavily rely on the quality of uncertainty estimates.
    
    
(2) The intrinsic motivation-oriented methods usually focus on hard-exploration tasks like Montezuma's Revenge. 
    For example, RND  \cite{BurdaESK19RND} outperforms human-level performance by using intrinsic rewards to constantly look for novel states to help the agent pass more room. 
    However, according to a recent empirical study  \cite{Taiga2020On}, although the existing methods greatly improve the performance in several hard-exploration tasks,
    they may have no positive effect on or even hinder the learning performance in other tasks.
    This can be attributed to the introduction of intrinsic motivation, which often alters the original learning objective and may deviate from the optimal policies.
    This raises a requirement for improving
    the versatility of intrinsic motivation-oriented methods.
    A preliminary success is achieved by NGU  \cite{Badia20NGU} that learns a series of $Q$ functions corresponding to different coefficients of the intrinsic reward,
    among which the exploitative $Q$ function is ready for execution.
    Such a technique is further improved by separate parameterization in Agent57  \cite{Badia20Agent57}.
    
(3) Other advanced exploration methods also pursue efficient exploration from different perspectives. 
    1) Distributed training greatly improves exploration and learning performance generally.
    Distributed training is one of the key components of Agent57  \cite{Badia20Agent57}, the first DRL agent that achieves superhuman performance on all 57 Atari games.
    2) $Q$-network with parametric noise brings stable performance improvement compared to a deterministic $Q$-network. 
    For an instance, 
    Noisy-Net  \cite{FortunatoAPMHOG18NoisyNet} significantly outperforms the intrinsic motivation-oriented methods evaluated by the overall Atari suite. 
    3) Another notable concept is potential-based exploration. 
    As a representative, PotER  \cite{ZhaoDZKLX20Potential} achieves good performance in Montezuma's Revenge with significantly higher sample efficiency than other methods,
    making potential-based exploration promising in addressing hard-exploration environments.
    

(4) MARL exploration methods have cutting edges in the SOTA benchmark, SMAC\cite{smac}. 1) Exploration is more necessary for super hard tasks of SMAC. MAVEN \cite{DBLP:conf/nips/MahajanRSW19} greatly improves the exploration of these tasks by using a hierarchical policy to control the shared latent variable as a coordination signal. 2) CMAE \cite{came} significantly improves exploration on sparse reward tasks of SMAC, which indicates it is more critical and promising to reduce the exploration space with both large state-action space and sparse reward challenges. 


\subsection{Open Problems}
\label{sec:open_problems}

Although encouraging progress has been achieved,
efficient exploration remains a challenging problem for DRL and deep MARL. 
Moreover, we discuss several open problems which are fundamental yet not well addressed by existing methods,
and point out a few potential solutions and directions.

\textbf{Exploration in Large State-action Space.} 
The difficulty of exploration escalates as the growth of scale and complexity of state-action space.
To deal with large state space, exploration methods often need a high-capacity neural network to measure the uncertainty and novelty. 
For representative uncertainty-oriented methods, Bootstrapped DQN  \cite{DBLP:conf/nips/OsbandBPR16}, OAC  \cite{DBLP:conf/nips/OsbandAC18} and IDS  \cite{DBLP:conf/colt/KirschnerK18} use 
Bayesian network to approximate the posterior of $Q$-function, which is computationally intensive.
Theoretically, accurately estimating the $Q$-posterior in large state space requires infinite bootstrapped $Q$-networks, which is obviously infeasible in practice, and instead an ensemble of $10$ $Q$-networks is often used.
Intrinsic motivation-oriented methods often need additional auxiliary models such as forward dynamics  \cite{DBLP:conf/icml/PathakG019} and density model  \cite{OstrovskiBOM17} to measure the novelty of states or transitions. 
For intrinsic motivation-oriented methods, learning accurate density estimation and effective auxiliary models such as forward dynamics in large state space is also nontrivial within a practical budget.
The consequent quality of uncertainty estimation and intrinsic guidance achieved thus in turn affects the exploration performance.

Another major limitation of existing works is the incapability of learning and exploring in large and complex action spaces.
Most methods consider a relatively small discrete action space or low-dimensional continuous action space.
Nevertheless, in many real-world scenarios, the action space can consist of a large number of discrete actions or has a complex underlying structure such as a hybrid of discrete and continuous action spaces.
Conventional DRL algorithms have scalability issues and even are infeasible to be applied.
A few recent works attempt to deal with large and complex action spaces in different ways. 
For example, 
 \cite{ChandakTKJT19ActionRep} proposes to learn a compact action representation of large discrete action spaces and convert the original policy learning into a low-dimensional space.
Another idea is to factorize the large action space  \cite{FarquharGLWUS20Growing}, e.g., into a hierarchy of small action spaces with different abstraction levels.
Besides, for structured action space like discrete-continuous hybrid action space, several works are proposed with sophisticated algorithms  \cite{HausknechtS15aPDDPG,Xiong18PDQN,FuTHLCF19MAHybrid,Li21HyAR}.
Despite the attempts made in the aforementioned works,
how efficient exploration can be realized with a large and complex action space remains unclear.

Since the main challenge is the large-scale and complex structure of state-action spaces, a natural solution is to construct an abstract and well-behaved space as a surrogate,
among which exploration can be conducted efficiently.
Thus, one promising way is to leverage the representation learning of states and actions.
The potential of representation learning in RL has been demonstrated
by some recent works in improving policy performance in environments with image states  \cite{LaskinSA20CURL} and hybrid actions  \cite{Li21HyAR}, as well as generalization across multiple tasks  \cite{YaratsFLP21ProtoRL}.
The representations in these works are learned by following specific criteria, e.g., reconstruction, instance-discriminative contrast, and dynamics prediction.
However, how an exploration-oriented state and action representations can be obtained is unclear yet.
To our knowledge, some efforts have been made in this direction.
For example, DB  \cite{db-2021} learns dynamics-relevant representations for exploration through the information bottleneck. SSR  \cite{MachadoBB20} makes use of successor representation upon which count-based exploration is then performed.
The central problem is 
what exploration-favorable information should be retained by the representation to learn. 
To fulfill exploration-oriented state representation,
we consider that the information of both the environment to explore and the agent's current knowledge are pivotal.
One feasible approach to leverage useful environment information is learning state abstraction based on 
the topology of the state space with actionable connectivity  \cite{GhoshGL19Actionable},
thus unnecessary exploration of redundant states can be avoided.
Taking into account agent's current knowledge, a further abstraction of state space can be achieved for more efficient exploration.
A potential way is establishing an equivalence relation based on the familiarity of states,
following which the boundary of exploration can be characterized.
In this manner, we expect highly targeted and efficient exploration can be realized.
For action representation, one key point may be the utilization of action semantics,
i.e., how the action affects the environment especially on the critical states.
The similarity of action semantics between actions enables effective generalization of learned knowledge,
e.g., the value estimate of an action can be generalized to other actions that have similar impacts on the environment.
At the same time, the distinction of action semantics can be made use of to select potential actions to seek for novel information.




\textbf{Exploration in Long-horizon Environments Extremely Sparse, Delayed Rewards.}
For exploration in environments with sparse, delayed rewards,
some promising results have been achieved by a few exploration methods 
from the perspective of intrinsic motivation  \cite{BurdaESK19RND,DBLP:conf/cig/SongCHF20}
or uncertainty guidance
 \cite{DBLP:conf/nips/OsbandAC18,ODonoghueOMM18}.
However, most current methods also reveal their incapability when dealing with sparser or extremely sparse rewards, typically in an environment with a long horizon.
As a typical example,
the whole game of Montezuma's Revenge has not been solved by DRL agents except for
Go-Explore \cite{Ecoffet19GoExp,Ecoffet20FirstReturn}.
However, Agent57 achieves over 9.3k scores by taking advantage of a two-scale novelty mechanism and an advanced distributed architecture base (i.e., R2D2  \cite{Kapturowski19R2D2});
Go-Explore  \cite{Ecoffet19GoExp,Ecoffet20FirstReturn} achieves superhuman performance
through imitating superior trajectory experiences collected in a return-and-explore fashion
with access to controlling the simulation.
These methods are far from sample efficiency and rely much on task-specific designs, thus lacking generality.
For real-world navigation scenarios, 
the practical methods often need to combine prior knowledge and intrinsic motivation to perform exploration in a long horizon. 
In such environments, the prior knowledge usually includes representative landmarks  \cite{mirowski2018learning} and topological graphs  \cite{savinov2018semi} which are designated by humans.
Apparently, current exploration methods for navigation rely heavily on prior knowledge.
However, this is often expensive and nontrivial to obtain in general.

Overall, learning in such an environment with extremely sparse, delayed rewards is an unsolved problem at present,
which is of significance to developing RL towards practical applications.
Intuitively, beyond sparse and delayed rewards, solving such problems involves higher-level requirements 
on long-time memorization of environmental context and versatile control of complex environmental semantics. 
These aspects can be the desiderata of effective exploration methods in future studies.
To take a step in this direction,
long-time memorization of the environmental context may be realized by more sophisticated models, e.g., Transformer  \cite{VaswaniSPUJGKP17Transformer} and Episodic Memory, in an implicit or explicit manner.
Another promising yet challenging solution is 
establishing a universal approach to extract the hierarchical structure of different environments.
Learning an ensemble of sub-policies (i.e., skills) is a possible way to fulfilling the versatile control of the environment,
based on which 
temporal abstracted exploration can be performed at higher levels.
In addition, incorporating general-form prior knowledge to reduce unnecessary exploration is also a promising perspective.


\textbf{Exploration with White-noise Problem.}
The stochasticity inherent in dynamics or manually injected in environments usually distracts agents in exploration. Several works like RND  \cite{BurdaESK19RND}, ICM  \cite{PathakAED17}, and EMI  \cite{KimKJLS19} all focus on solving state-space noise by constructing a compact feature representation to discard task-irrelevant features. Count-based exploration handles stochasticity in state-space through attention  \cite{DBLP:conf/iclr/ChoiGMOWNL19} and state-space VAE  \cite{klissarovvariational}. The state representation discards the task-irrelevant information in exploration, which is promising to overcome the white-noise problem. 
However, most methods require a dynamics model or state-encoder, which increases the computational cost. Although CB  \cite{DBLP:conf/icml/KimNKKK19} does not learn a state-encoder, it 
needs environmental rewards to remove the noisy information and thus cannot work in extremely sparse reward tasks.
Other methods to solve this problem include Bayesian disagreement  \cite{DBLP:conf/icml/PathakG019} and active world model learning  \cite{kim2020active}. They use the Bayesian uncertainty and information gain-based model that is insensitive to white-noise to overcome the state-space noise. Although they are promising to handle the state space noise, the action-space noise has not been rigorously discussed in the community. The sticky Atari  \cite{machado2018revisiting} injects action-space noise in discrete action space, while it is hard to design and represent the realistic noise in real-world applications. How to construct a more realistic distraction is worth exploring direction in the future.
One possible direction in solving the white-noise problem is using adversarial training to learn a robust policy. Assuming that $s+\epsilon$ is a noisy state with a noise $\epsilon$ sampled from a distribution, we can find the adversarial examples by solving a min-max problem and then reduce the sensibility of the model to the adversaries. Specifically, we can enforce an adversarial smooth loss on value function as,
$\min_\theta \max_{\epsilon} \big[V_\theta(s+\epsilon)-V_\theta(s)\big]^2$,
where the inner maximum finds a noisy state that is easy to attack, while the outer minimum enforces smoothness on the value function.
Another promising direction is using direct regularization. For example, enforcing Lipschitz constraints for policy networks to make the predictions provably robust to small perturbations produced by noise; using bisimulation constraints to reduce the effects of noise by learning the dynamics-relevant representations.

\textbf{Convergence.}
For uncertainty-oriented exploration, optimism and Thomson sampling-based methods need the uncertainty to converge to zero in the training process to make the policy converge. Theoretically, the epistemic uncertainty enables convergence to zero in tabular  \cite{jin2018q} and linear MDPs  \cite{OPPO-2020,jin2020provably} according to the theoretical results. In general, MDPs, as the agent learns more about the environment, the uncertainty that encourages exploration gradually decreases to zero, then the confidence set of the MDP posterior will contain the true MDP with a high probability  \cite{auer2009near,bound-2016}.
However, due to the curse of dimensionality, the practical uncertainty estimation mostly relies on function approximation like neural networks, which makes the estimation errors hard to converge. Meanwhile, the bootstrapped sampling  \cite{DBLP:conf/nips/OsbandBPR16}, linear final layer approximation  \cite{DBLP:conf/ita/Azizzadenesheli18} or variational inference  \cite{FortunatoAPMHOG18NoisyNet,lipton2018bbq} only provide the approximated Bayesian posterior rather than the accurate posterior, which may be hard to represent the true confidence of the value function with the changing experiences in exploration. 
The recently proposed Single Model Uncertainty  \cite{tagasovska2018single} is a kind of uncertainty measurement through a single model, which may provide more stable rewards in exploration. 
Combining out-of-distribution detection methods  \cite{thudumu2020comprehensive} with RL exploration is also a promising direction.

For intrinsic motivation-oriented exploration, a regular operation is to add an intrinsic reward to the extrinsic reward, which is a kind of reward shaping mechanism. However, these reward transformation methods are usually heuristically designed while not theoretically justified. Unlike potential-based reward  \cite{ng1999policy} that does not change the optimal policy, the reward transformations may hinder the performance and converge to the suboptimal solution. There are several efforts that propose novel strategies to combine extrinsic and intrinsic policy without reward transformation. For example, scheduled intrinsic drive  \cite{DBLP:journals/corr/abs-1903-07400} and MuleX  \cite{beyer2019mulex} learn intrinsic and extrinsic policies independently, and then use a high-level scheduler or random heuristic to decide which one to use in each time step. This method improves the robustness of the combined policy while also does not make the theoretical guarantee of optimality. The two-objective optimization  \cite{zhang2019learning} calculates the gradient of intrinsic and extrinsic reward independently and then combines the gradient directions according to their angular bisectors.
However, this technique does not work when extrinsic rewards are almost absent. Recently, meta-learned rewards   \cite{zheng2018learning,zheng2019can} are proposed to learn an optimal intrinsic reward by following the gradient of extrinsic reward, which guarantees the optimality of intrinsic rewards. However, how to combine this method with existing bonus-based exploration remains an open question. Another promising research direction is designing intrinsic rewards following the potential-based reward shaping principle to ensure the optimality of the combined policy.

 
\textbf{Multi-agent Exploration.}
The research of multi-agent exploration is still at an early stage and does not address most of the mentioned challenges well, such as partial observation, non-stationary, high-dimensional state-action space, and coordinated exploration. 
Since the joint state-action space grows exponentially as the increase in number of agents, the difficulty in measuring the uncertainty or intrinsic motivation escalates significantly.
Referring to the discussion of the large state-action space,
the scalability problem in multi-agent exploration can also be alleviated by representation learning.
Beyond the representation approaches in single-agent domain,
one special and significant information that can be taken advantage of is to build a graph structure of multi-agent interactions and relations.
It is of great potential while remaining open to incorporating such graph structure information in state and action representation learning for effective abstraction and reduction of the joint space.
Furthermore, with the inconsistency between local and global information (detailed in Section ~\ref{sec:challenge}), how to obtain a robust and accurate uncertainty (or intrinsic motivation) estimation that balances the local and global information to derive a coordinated exploration is worthwhile to be studied. One promising solution is to design exploration strategies by taking credit assignment into consideration, this technique has been successfully applied to achieve multi-agent coordination \cite{FoersterFANW18COMA}, which may help to assign a  reasonable intrinsic reward for each agent to solve this problem. 
Another problem is that there is still no well-accepted benchmark yet, most of the previous works have designed specific test beds to verify the effectiveness of their proposed corresponding solutions, such as influence-based exploration strategies \cite{DBLP:conf/iclr/0001WWZ20,DBLP:conf/icml/JaquesLHGOSLF19} which are tested on environments that need strong cooperative and coordinated behaviors (shown in Fig ~\ref{fig:pass}); novelty-based strategies  \cite{DBLP:journals/corr/abs-1905-12127, DBLP:journals/corr/abs-1906-02138} are tested on environments where novel states are well correlated with improved rewards. Therefore, how to construct a well-accepted benchmark and derive a general exploration strategy that is suited for the benchmark remains an open problem in multi-agent exploration. 

\textbf{Safe Exploration.}
Although a wide range of works have studied safe RL,
it remains challenging to achieve safe exploration during the training process.
In other words, the safety requirement is not only the ultimate goal to be achieved in evaluation and deployment, but also expected to be maximally satisfied during the sampling and interaction in the environments.
As we discussed in Sec.~\ref{subsec:other_method},
arguably the most effective way to realize safe exploration is to incorporate prior knowledge, which also includes all kinds of demonstrations and risk definitions in a general view.
However, such prior knowledge is often task-specific and not always available.
How to establish a general framework for defining or learning safety knowledge is an open question.
Moreover, another fundamental problem in safe exploration is the conflict between safety requirements and exploration effectiveness.
Without full knowledge of the environment or perfect prior knowledge,
it is inevitable to violate the safety requirement some times when performing exploration during the training process.
The principled approaches to balance such a trade-off are under-explored~\cite{BharadhwajKRLSG21Conser}.
One possible angle to deal with such conflicts between the safety requirement and the RL objective is Multi-objective Learning~\cite{NguyenNVNDL20multiobj}.
In addition, safe RL exploration methods, e.g., with theoretical guarantees~\cite{Bura21SEProvable} and for dynamical systems are other open angles.


\section{Conclusion}
\label{sec7}

In this paper, we conduct a comprehensive survey on exploration in DRL and deep MARL.
We identify the major challenges of exploration for both single-agent and multi-agent RL.
We investigate previous efforts on addressing these challenges by a taxonomy consisting of two major categories:
uncertainty-oriented exploration and intrinsic motivation-oriented exploration.
Besides, some other advanced exploration methods with distinct ideas are also concluded. 
In addition to the study on algorithmic characteristics, we provide a unified comparison of the most current exploration methods for DRL
on four representative benchmarks.
From the empirical results, we shed some light on the specialties  and limitations of different categories of exploration methods.
Finally, we summarize several open questions.
We discuss in depth what is expected to be solved and provide a few potential solutions that are worthwhile being further studied in the future.

At present,
exploration in environments with large state-action space, long-horizon, and complex semantics is still very challenging with current advances.
Moreover, exploration in deep MARL is much less studied than that in the single-agent setting.
Towards a broader perspective to address exploration problems in DRL and deep MARL, we highlight a few suggestions and insights below:
1) First, 
current exploration methods are evaluated mainly in terms of cumulative rewards and sample efficiency 
in only several well-known hard-exploration environments or manually designed environments.
The lack of multidimensional criteria and standard experimental benchmarks inhibits the calibration and evaluation of different methods.
In this survey, we identify several challenges to exploration and use them for qualitative analysis of different methods from the algorithmic aspect.
New environments specific to these challenges and corresponding quantitative evaluation criteria are of necessity.
2) Second, the essential connections between different exploration methods are to be further revealed.
For example, although intrinsic motivation-oriented exploration methods usually come from strong heuristics, 
some of them are closely related to uncertainty-oriented exploration methods, as the underlying connection between RND  \cite{BurdaESK19RND} and Random Prior Fitting  \cite{CiosekFTHT20FittingPrior}.
The study of such essential connections can promote the unification and classification of both the theory and methodology of RL exploration.
3) Third, exploration among large action space, exploration in long-horizon environments, and convergence analysis are relatively lacking studies.
The progress in solving these problems is significant to both the practical application of RL algorithms and RL theoretical study.
4) Lastly, multi-agent exploration can be even more challenging due to complex multi-agent interactions.
At present, many multi-agent exploration methods are proposed by integrating the exploration methods in single-agent settings and MARL algorithms.
Coordinated exploration with decentralized execution and exploration under non-stationarity may be the key problems to address.


%

\ifCLASSOPTIONcompsoc
\else
\fi

\ifCLASSOPTIONcaptionsoff
  \newpage
\fi

\appendices

\section{Model-based Exploration}

In the main text, we mainly conduct a survey on exploration methods in a model-free domain. In this section, we briefly introduce exploration methods for model-based RL. Model-based RL uses an environment model to generate simulated experience or planning. In exploration, RL algorithms can directly use this environment model for uncertainty estimation or intrinsic motivation. 

For uncertainty-based exploration, model-based RL is based on optimism for planning and exploration~\cite{nix1994estimating}. For example, Model-assisted RL \cite{kalweit2017uncertainty} uses an ensemble model to measure the dynamics uncertainty, and makes use of artificial data only in cases of high uncertainty, which encourages the agent to learn difficult parts of the environment. Planning to explore~\cite{sekar2020planning} also uses ensemble dynamics for uncertainty estimation, and seeks out future uncertainty by integrating uncertainty to Dreamer-based planning~\cite{hafner2019dream}. Noise-Augmented RL~\cite{pacchiano2020optimism} uses statistical bootstrap to generalize the optimistic posterior sampling~\cite{agrawal2017posterior} to DRL. Hallucinated UCRL~\cite{curi2020efficient} 
measures the epistemic uncertainty and reduces optimistic exploration to exploitation by enlarging the control space. 

For intrinsic motivation-based exploration, model-based RL usually uses the information gain of the dynamics to incentivize exploration. Ready Policy One~\cite{ball2020ready} uses Gaussian distribution to build the environment model and measures the information gain through the Gaussian process. Then they optimize the policies for both reward and model uncertainty reduction, which encourages the agent to explore high-uncertainty areas. Dynamics Bottleneck~\cite{db-2021} uses variational methods to learn the latent space of dynamics and uses the information gain measured by the difference between the prior and posterior distributions of the latent variables.

\section{Hindsight-based Exploration}

In this section, we discuss a specific exploration problem in goal-conditional RL settings. For example, the agent needs to reach the specific goal in mazes, but only receives a reward if it successfully reaches the desired goal. It is almost impossible to reach the goal by chance, even in the simplest environment. 

Hindsight experience replay (HER)~\cite{HER-2017} shows promising results in solving such problems by using hindsight goals to make the agent learn from failures. Specifically, the HER agent samples the already-achieved goals from a failed experience as the hindsight goals. The hindsight goals are used to substitute the original goals and recompute the reward functions. Because the hindsight goals lie close to the sampled transitions, the agent frequently receives the hindsight rewards and accelerates learning. On the basis of HER, hindsight policy gradient~\cite{HPG-2019} extends HER to on-policy RL by using importance sampling. RIG~\cite{vrl-2018} uses HER to handle image-based observation by learning a latent space from a variational autoencoder. Dynamic HER~\cite{dher-2019} solves dynamic goals in real robotics tasks by assembling successful experiences from two failures. Competitive experience replay~\cite{cer-2019} introduces self-play~\cite{Selfplay-2018} between two agents and generates a curriculum for exploration. Entropy-regularized HER~\cite{mep-2019} develops a maximum entropy framework to optimize the prioritized multi-goal RL. Curriculum-guided HER~\cite{curr-2019} selects the replay experiences according to the proximity to true goals and the curiosity of exploration. Hindsight goals generation~\cite{HGG-2019} creates valuable goals that are easy to achieve and valuable in the long term. Directed Exploration~\cite{guo2019directed} uses the prediction error of dynamics to choose high uncertain goals and learns a goal-conditional policy to reach them based on HER.

\section{Exploration via Skill Discovery}

Recent research converts the exploration problem to an unsupervised skill-discovery problem. That is, through learning skills in reward-free exploration, the agent collects information about the environment and learns skills that are useful for downstream tasks. We discuss this kind of learning method in the following.

The unsupervised skill-discovery methods aim to learn skills by exploring the environment without any extrinsic rewards. The basic principle of skill discovery is empowerment \cite{salge2014empowerment}, which describes what the agent can be done while learning how to do it. The skill is a latent variable $z\sim p(z)$ sampled from a skill space, which can be continuous or discrete. The skill discovery methods seeks to find a skill-conditional policy $\pi(a|s,z)$ to maximize the mutual information between $S$ and $Z$. Formally, the mutual information of $S$ and $Z$ can be solved in reverse and forward forms as follows,
\begin{equation}\label{eq:skill-rev}
I(S;Z)=H(Z)-H(Z|S).~~~{\rm (Reverse)}
\end{equation}\vspace{-1.5em}
\begin{equation}\label{eq:skill-for}
I(S;Z)=H(S)-H(S|Z).~~~{\rm (Forward)}
\end{equation}

Most existing skill discovery methods follow the reverse form defined in Eq.~\eqref{eq:skill-rev}, where $I(S;Z)=\mathbb{E}_{s,z\sim p(s,z)}[\log p(z|s)]-\mathbb{E}_{z\sim p(z)}[\log p(z)]$. A variational lower bound $I(S;Z)\geq\mathbb{E}_{s,z\sim p(s,z)}[\log q_\phi(z|s)]-\mathbb{E}_{z\sim p(z)}[\log p(z)]$ is derived by using $q_\phi(z|s)$ to approximate the posterior  $p(z|s)$ of skill, where $q_\phi(z|s)$ is trained by maximizing the likelihood of $(s,z)$ collected by the agent. The mutual information $I(S;Z)$ is maximized by using the variational lower bound as the reward function when training the policy. Following this principle, variational intrinsic control~\cite{gregor2016variational} considers $s$ as the last state in a trajectory and the skill is sampled from a prior distribution. DIAYN \cite{eysenbach2018diversity} trains the policy to maximize $I(S;Z)$ while minimizing $I(A;Z|S)$ additionally, which pushes the skills away from each other to learn distinguishable skills. VALOR~\cite{achiam2018variational} uses a trajectory $\tau$ to calculate the variational distribution $q_\phi(z|\tau)$ conditioned on the trajectory rather than $q_\phi(z|s)$ calculated by the individual states, which encourages the agent to learn dynamical modes. Skew-Fit \cite{pong2019skew} uses states sampled from the replay buffer as goals $G$ and the agent learns to maximize $I(S;G)$, where $G$ serves as the skill $Z$ in Eq. \eqref{eq:skill-rev}.

There are several recent works that use the forward form of $I(S,Z)$ defined in Eq. \eqref{eq:skill-for} to maximize the mutual information, where $I(S;Z)=\mathbb{E}_{s,z\sim p(s,z)}[\log p(s|z)]-\mathbb{E}_{s\sim p(s)}[\log p(s)]$. Similar to the reverse form, a variational distribution $q_\phi(s|z)$  that fits the $(s,z)$ tuple collected by the agent is used to form a variational lower bound as $I(S,Z)\geq \mathbb{E}_{s,z\sim p(s,z)}[\log q_\phi(s|z)]-\mathbb{E}_{s\sim p(s)}[\log p(s)]$. DADS \cite{sharma2019dynamics} maximizes the mutual information between skill and the next state $I(S',Z|S)$ that conditions on the current state. Then the variational distribution $q_\phi(s|z)$ becomes $q_\phi(s'|s,z)$, which represents a skill-level dynamics model and enables the use of model-based planning algorithms for downstream tasks. EDL \cite{campos2020explore} finds that existing skill discovery methods tend to visit known states rather than discovering new ones when maximizing the mutual information, which leads to a poor coverage of the state space. Based on the analyses, EDL uses State Marginal Matching (SMM) \cite{DBLP:journals/corr/abs-1906-05274} to perform maximum entropy exploration in the state space before skill discovery, which results in a fixed $p(s)$ that is agnostic to the skill. EDL succeeds at discovering state-covering skills in environments where previous methods failed.

\section{Self-imitation Learning}

In addition to the various methods we have discussed, Self-Imitation Learning (SIL) is another branch of methods that can facilitate RL in the face of learning and exploration difficulties.
The main idea of SIL is to imitate the agent's own experiences collected during the historical learning process.
A SIL agent tries to make the best use of the superior (e.g., successful) experiences encountered by occasion.

Oh et al. \cite{OhGSL18SIL} firstly propose the concept of SIL. They propose additional off-policy AC losses to ascend the policy towards non-negative advantages and make the value function approximate the corresponding returns.
They demonstrate the effectiveness of SIL when combined with A2C and PPO in 49 Atari games and the Key-Door-Treasure domain. Furthermore, they shed light on the theoretical connection between SIL and lower-bound Soft Q-Learning \cite{HaarnojaTAL17SQL}.
From a different angle, Gangwani et al. \cite{GangwaniLP19SIDP} derive a SIL algorithm by casting the divergence minimization problem of the state-action visitation distributions of the policy to optimize and (historical) good policies, to a policy gradient objective.
In addition, Gangwani et al. \cite{GangwaniLP19SIDP} utilize Stein Variation Policy Gradient (SVPG) \cite{LiuRLP17SVPG} for the purpose of optimizing an ensemble of diverse SI policies.
Later, SIL is developed from transition level to trajectory level with the help of trajectory \cite{Guo19ExpSIL}  or goal conditioned policies \cite{DAI20ESIL}. Diverse Trajectory-conditioned SIL (DTSIL) \cite{Guo19ExpSIL} is proposed to maintain a buffer of good and diverse trajectories based on their similarity which are taken as input of a trajectory-conditioned policy during SIL.
Episodic SIL (ESIL) \cite{DAI20ESIL} is proposed to also perform trajectory-level SIL with a trajectory-based episodic memory;
to train a goal-conditioned policy, HER \cite{HER-2017} is utilized to alter the original (failed) goals among which good trajectories to self-imitate are then determined and filtered.

Recently, a few works improve original SIL algorithm \cite{OhGSL18SIL} in different ways. 
Tang \cite{Tang20SILQ} proposes a family of SIL algorithms through generalizing the original return-based lower-bound Q-learning \cite{OhGSL18SIL} to $n$-step TD lower bound, 
balancing the trade-off between fixed point bias and contraction rate.
Chen et al. \cite{Chen20SILSR} propose SIL with Constant Reward (SILCR) to get rid of the need of immediate environment rewards (but make use of episodic reward instead).
Beyond policy-based SIL, Ferret et al. propose Self-Imitation Advantage Learning (SAIL) \cite{FerretPG21SIAL} which extends SIL to off-policy value-based RL algorithms through modifying Bellman optimality operator that connects to Advantage
Learning \cite{BellemareSOSSM16}.
Ning et al. \cite{Ning21CoIL} propose Co-Imitation Learning (CoIL) which learns two different agents via letting each of them alternately explore the environment and selectively imitate the heterogeneous good experiences from each other.

Overall, SIL is different from the two main categories of methods considered in our main text, i.e., intrinsic motivation and uncertainty. From some angle, SIL methods can be viewed as passive exploration methods which reinforce the exploration the informative reward signals in the experience buffer. 



\bibliographystyle{IEEEtran}
\bibliography{IEEEtran}

\begin{thebibliography}{100}
\providecommand{\url}[1]{#1}
\csname url@samestyle\endcsname
\providecommand{\newblock}{\relax}
\providecommand{\bibinfo}[2]{#2}
\providecommand{\BIBentrySTDinterwordspacing}{\spaceskip=0pt\relax}
\providecommand{\BIBentryALTinterwordstretchfactor}{4}
\providecommand{\BIBentryALTinterwordspacing}{\spaceskip=\fontdimen2\font plus
\BIBentryALTinterwordstretchfactor\fontdimen3\font minus
  \fontdimen4\font\relax}
\providecommand{\BIBforeignlanguage}[2]{{%
\expandafter\ifx\csname l@#1\endcsname\relax
\typeout{** WARNING: IEEEtran.bst: No hyphenation pattern has been}%
\typeout{** loaded for the language `#1'. Using the pattern for}%
\typeout{** the default language instead.}%
\else
\language=\csname l@#1\endcsname
\fi
#2}}
\providecommand{\BIBdecl}{\relax}
\BIBdecl

\bibitem{alphaGo-2016}
D.~Silver, A.~Huang, C.~J. Maddison, A.~Guez, L.~Sifre, G.~Van Den~Driessche,
  J.~Schrittwieser, I.~Antonoglou, V.~Panneershelvam, M.~Lanctot \emph{et~al.},
  ``Mastering the game of go with deep neural networks and tree search,''
  \emph{nature}, vol. 529, no. 7587, pp. 484--489, 2016.

\bibitem{Badia20Agent57}
A.~P. Badia, B.~Piot, S.~Kapturowski, P.~Sprechmann, A.~Vitvitskyi, D.~Guo, and
  C.~Blundell, ``Agent57: Outperforming the atari human benchmark,''
  \emph{arXiv preprint arXiv:2003.13350}, 2020.

\bibitem{RashidSWFFW18Qmix}
T.~Rashid, M.~Samvelyan, C.~S. de~Witt, G.~Farquhar, J.~N. Foerster, and
  S.~Whiteson, ``{QMIX:} monotonic value function factorisation for deep
  multi-agent reinforcement learning,'' in \emph{ICML}, 2018.

\bibitem{LillicrapHPHETS15DDPG}
T.~P. Lillicrap, J.~J. Hunt, A.~Pritzel, N.~Heess, T.~Erez, Y.~Tassa,
  D.~Silver, and D.~Wierstra, ``Continuous control with deep reinforcement
  learning,'' in \emph{ICLR}, 2016.

\bibitem{PathakAED17}
D.~Pathak, P.~Agrawal, A.~A. Efros, and T.~Darrell, ``Curiosity-driven
  exploration by self-supervised prediction,'' in \emph{ICML}, 2017.

\bibitem{tnnls-9410239}
M.~Li, Z.~Cao, and Z.~Li, ``A reinforcement learning-based vehicle platoon
  control strategy for reducing energy consumption in traffic oscillations,''
  \emph{IEEE Transactions on Neural Networks and Learning Systems}, vol.~32,
  no.~12, pp. 5309--5322, 2021.

\bibitem{KongHYLK21}
L.~Kong, W.~He, W.~Yang, Q.~Li, and O.~Kaynak, ``Fuzzy approximation-based
  finite-time control for a robot with actuator saturation under time-varying
  constraints of work space,'' \emph{IEEE Transactions on Cybernetics},
  vol.~51, no.~10, pp. 4873--4884, 2021.

\bibitem{survey-drl-1}
H.-n. Wang, N.~Liu, Y.-y. Zhang, D.-w. Feng, F.~Huang, D.-s. Li, and Y.-m.
  Zhang, ``Deep reinforcement learning: a survey,'' \emph{FRONT INFORM TECH
  EL}, pp. 1--19, 2020.

\bibitem{survey-drl-2}
Y.~Li, ``Deep reinforcement learning: An overview,'' \emph{arXiv preprint
  arXiv:1701.07274}, 2017.

\bibitem{survey-drl-3}
S.~S. Mousavi, M.~Schukat, and E.~Howley, ``Deep reinforcement learning: an
  overview,'' in \emph{SAI Intelligent Systems Conference}, 2016.

\bibitem{survey-drl-4}
K.~Arulkumaran, M.~P. Deisenroth, M.~Brundage, and A.~A. Bharath, ``Deep
  reinforcement learning: A brief survey,'' \emph{IEEE Signal Processing
  Magazine}, vol.~34, no.~6, pp. 26--38, 2017.

\bibitem{phdthesis:Dann2020StrategicEI}
C.~Dann, ``Strategic exploration in reinforcement learning - new algorithms and
  learning guarantees,'' Ph.D. dissertation, School of Computer Science,
  Carnegie Mellon University, 2020.

\bibitem{survey-madrl-1}
L.~Busoniu, R.~Babuska, and B.~De~Schutter, ``A comprehensive survey of
  multiagent reinforcement learning,'' \emph{TSMC}, vol.~38, no.~2, pp.
  156--172, 2008.

\bibitem{survey-madrl-2}
P.~Hernandez-Leal, B.~Kartal, and M.~E. Taylor, ``A survey and critique of
  multiagent deep reinforcement learning,'' \emph{JAAMAS}, vol.~33, no.~6, pp.
  750--797, 2019.

\bibitem{survey-madrl-3}
K.~Zhang, Z.~Yang, and T.~Ba{\c{s}}ar, ``Multi-agent reinforcement learning: A
  selective overview of theories and algorithms,'' \emph{arXiv preprint
  arXiv:1911.10635}, 2019.

\bibitem{aubret2019survey}
A.~Aubret, L.~Matignon, and S.~Hassas, ``A survey on intrinsic motivation in
  reinforcement learning,'' \emph{arXiv preprint arXiv:1908.06976}, 2019.

\bibitem{BellemareSOSSM16}
M.~G. Bellemare, S.~Srinivasan, G.~Ostrovski, T.~Schaul, D.~Saxton, and
  R.~Munos, ``Unifying count-based exploration and intrinsic motivation,'' in
  \emph{NeurIPS}, 2016.

\bibitem{tnnls-9119863}
C.~Sun, W.~Liu, and L.~Dong, ``Reinforcement learning with task decomposition
  for cooperative multiagent systems,'' \emph{IEEE Transactions on Neural
  Networks and Learning Systems}, vol.~32, no.~5, pp. 2054--2065, 2021.

\bibitem{tnnls-9205265}
J.~Fu, X.~Teng, C.~Cao, Z.~Ju, and P.~Lou, ``Robot motor skill transfer with
  alternate learning in two spaces,'' \emph{IEEE Transactions on Neural
  Networks and Learning Systems}, vol.~32, no.~10, pp. 4553--4564, 2021.

\bibitem{bandit-book-2020}
T.~Lattimore and C.~Szepesvári, \emph{Bandit Algorithms}.\hskip 1em plus 0.5em
  minus 0.4em\relax Cambridge University Press, 2020.

\bibitem{ladosz2022exploration}
P.~Ladosz, L.~Weng, M.~Kim, and H.~Oh, ``Exploration in deep reinforcement
  learning: A survey,'' \emph{Information Fusion}, 2022.

\bibitem{ryan2000intrinsic}
R.~M. Ryan and E.~L. Deci, ``Intrinsic and extrinsic motivations: Classic
  definitions and new directions,'' \emph{Contemporary educational psychology},
  vol.~25, no.~1, pp. 54--67, 2000.

\bibitem{barto2013intrinsic}
A.~G. Barto, ``Intrinsic motivation and reinforcement learning,'' in
  \emph{Intrinsically motivated learning in natural and artificial systems},
  2013.

\bibitem{SuttonB98RLAI}
R.~S. Sutton and A.~G. Barto, \emph{Reinforcement learning - an introduction},
  ser. Adaptive computation and machine learning, 1998.

\bibitem{DBLP:journals/nature/MnihKSRVBGRFOPB15}
V.~Mnih, K.~Kavukcuoglu, D.~Silver, A.~A. Rusu, J.~Veness, M.~G. Bellemare,
  A.~Graves, M.~A. Riedmiller, A.~Fidjeland, G.~Ostrovski, S.~Petersen,
  C.~Beattie, A.~Sadik, I.~Antonoglou, H.~King, D.~Kumaran, D.~Wierstra,
  S.~Legg, and D.~Hassabis, ``Human-level control through deep reinforcement
  learning,'' \emph{Nature}, vol. 518, no. 7540, pp. 529--533, 2015.

\bibitem{HasseltGS16DDQN}
H.~van Hasselt, A.~Guez, and D.~Silver, ``Deep reinforcement learning with
  double q-learning,'' in \emph{AAAI}, 2016.

\bibitem{BellemareDM17C51}
M.~G. Bellemare, W.~Dabney, and R.~Munos, ``A distributional perspective on
  reinforcement learning,'' in \emph{ICML}, 2017.

\bibitem{DabneyRBM18QRDQN}
W.~Dabney, M.~Rowland, M.~G. Bellemare, and R.~Munos, ``Distributional
  reinforcement learning with quantile regression,'' in \emph{AAAI}, 2018.

\bibitem{WangSHHLF16Dueling}
Z.~Wang, T.~Schaul, M.~Hessel, H.~van Hasselt, M.~Lanctot, and N.~de~Freitas,
  ``Dueling network architectures for deep reinforcement learning,'' in
  \emph{ICML}, 2016.

\bibitem{SchaulQAS15PER}
T.~Schaul, J.~Quan, I.~Antonoglou, and D.~Silver, ``Prioritized experience
  replay,'' in \emph{ICLR}, 2016.

\bibitem{Williams92REINFORCE}
R.~J. Williams, ``Simple statistical gradient-following algorithms for
  connectionist reinforcement learning,'' \emph{Machine Learning}, vol.~8, pp.
  229--256, 1992.

\bibitem{SilverLHDWR14DPG}
D.~Silver, G.~Lever, N.~Heess, T.~Degris, D.~Wierstra, and M.~A. Riedmiller,
  ``Deterministic policy gradient algorithms,'' in \emph{ICML}, 2014.

\bibitem{LoweWTHAM17MADDPG}
R.~Lowe, Y.~Wu, A.~Tamar, J.~Harb, P.~Abbeel, and I.~Mordatch, ``Multi-agent
  actor-critic for mixed cooperative-competitive environments,'' in
  \emph{NeurIPS}, 2017.

\bibitem{lai1985asymptotically}
T.~L. Lai and H.~Robbins, ``Asymptotically efficient adaptive allocation
  rules,'' \emph{Advances in applied mathematics}, vol.~6, no.~1, pp. 4--22,
  1985.

\bibitem{DBLP:conf/icml/MnihBMGLHSK16}
V.~Mnih, A.~P. Badia, M.~Mirza, A.~Graves, T.~P. Lillicrap, T.~Harley,
  D.~Silver, and K.~Kavukcuoglu, ``Asynchronous methods for deep reinforcement
  learning,'' in \emph{ICML}, 2016.

\bibitem{FujimotoHM18TD3}
S.~Fujimoto, H.~van Hoof, and D.~Meger, ``Addressing function approximation
  error in actor-critic methods,'' in \emph{ICML}, 2018.

\bibitem{tutorialofbo}
P.~I. Frazier, ``A tutorial on bayesian optimization,'' \emph{arXiv preprint
  arXiv:1807.02811}, 2018.

\bibitem{DBLP:conf/icml/SrinivasKKS10}
N.~Srinivas, A.~Krause, S.~M. Kakade, and M.~W. Seeger, ``Gaussian process
  optimization in the bandit setting: No regret and experimental design,'' in
  \emph{ICML}, 2010.

\bibitem{thompson1933likelihood}
W.~R. Thompson, ``On the likelihood that one unknown probability exceeds
  another in view of the evidence of two samples,'' \emph{Biometrika}, vol.~25,
  no. 3/4, pp. 285--294, 1933.

\bibitem{HausknechtS15aPDDPG}
M.~J. Hausknecht and P.~Stone, ``Deep reinforcement learning in parameterized
  action space,'' in \emph{ICLR}, 2016.

\bibitem{ChandakTKJT19ActionRep}
Y.~Chandak, G.~Theocharous, J.~Kostas, S.~M. Jordan, and P.~S. Thomas,
  ``Learning action representations for reinforcement learning,'' in
  \emph{ICML}, 2019.

\bibitem{Li21HyAR}
B.~Li, H.~Tang, Y.~Zheng, J.~Hao, P.~Li, Z.~Wang, Z.~Meng, and L.~Wang, ``Hyar:
  Addressing discrete-continuous action reinforcement learning via hybrid
  action representation,'' \emph{arXiv preprint arXiv:2109.05490}, 2021.

\bibitem{DBLP:conf/icml/Strens00}
M.~J.~A. Strens, ``A bayesian framework for reinforcement learning,'' in
  \emph{ICML}, 2000.

\bibitem{Brockman2016Gym}
G.~Brockman, V.~Cheung, L.~Pettersson, J.~Schneider, J.~Schulman, J.~Tang, and
  W.~Zaremba, ``Openai gym,'' 2016.

\bibitem{DBLP:conf/nips/OsbandAC18}
I.~Osband, J.~Aslanides, and A.~Cassirer, ``Randomized prior functions for deep
  reinforcement learning,'' in \emph{NeurIPS}, 2018.

\bibitem{BurdaESK19RND}
Y.~Burda, H.~Edwards, A.~J. Storkey, and O.~Klimov, ``Exploration by random
  network distillation,'' in \emph{ICLR}, 2019.

\bibitem{DBLP:conf/iclr/ChoiGMOWNL19}
J.~Choi, Y.~Guo, M.~Moczulski, J.~Oh, N.~Wu, M.~Norouzi, and H.~Lee,
  ``Contingency-aware exploration in reinforcement learning,'' in \emph{ICLR},
  2019.

\bibitem{Badia20NGU}
A.~P. Badia, P.~Sprechmann, A.~Vitvitskyi, D.~Guo, B.~Piot, S.~Kapturowski,
  O.~Tieleman, M.~Arjovsky, A.~Pritzel, A.~Bolt, and C.~Blundell, ``Never give
  up: Learning directed exploration strategies,'' \emph{arXiv preprint
  arXiv:2002.06038}, 2020.

\bibitem{montezuma24rooms}
Atariage, ``Atari 2600 archives: Montezuma's revenge,''
  \url{https://atariage.com/2600/archives/strategy_MontezumasRevenge_Level1.html}.

\bibitem{TesslerGZMM17HieLifelong}
C.~Tessler, S.~Givony, T.~Zahavy, D.~J. Mankowitz, and S.~Mannor, ``A deep
  hierarchical approach to lifelong learning in minecraft,'' in \emph{AAAI},
  2017.

\bibitem{DBLP:conf/cig/KempkaWRTJ16}
M.~Kempka, M.~Wydmuch, G.~Runc, J.~Toczek, and W.~Jaskowski, ``Vizdoom: {A}
  doom-based {AI} research platform for visual reinforcement learning,'' in
  \emph{CIG}, 2016.

\bibitem{BurdaEPSDE19}
Y.~Burda, H.~Edwards, D.~Pathak, A.~J. Storkey, T.~Darrell, and A.~A. Efros,
  ``Large-scale study of curiosity-driven learning,'' in \emph{ICLR}, 2019.

\bibitem{DBLP:conf/icml/KimNKKK19}
Y.~Kim, W.~Nam, H.~Kim, J.~Kim, and G.~Kim, ``Curiosity-bottleneck: Exploration
  by distilling task-specific novelty,'' in \emph{ICML}, 2019.

\bibitem{DBLP:conf/iclr/0001WWZ20}
T.~Wang, J.~Wang, Y.~Wu, and C.~Zhang, ``Influence-based multi-agent
  exploration,'' in \emph{ICLR}, 2020.

\bibitem{osband2019deep}
I.~Osband, B.~Van~Roy, D.~J. Russo, and Z.~Wen, ``Deep exploration via
  randomized value functions.'' \emph{JMLR}, vol.~20, no. 124, pp. 1--62, 2019.

\bibitem{DBLP:conf/icml/OsbandRW16}
I.~Osband, B.~V. Roy, and Z.~Wen, ``Generalization and exploration via
  randomized value functions,'' in \emph{ICML}, 2016.

\bibitem{DBLP:conf/ita/Azizzadenesheli18}
K.~Azizzadenesheli, E.~Brunskill, and A.~Anandkumar, ``Efficient exploration
  through bayesian deep q-networks,'' in \emph{Information Theory and
  Applications Workshop}, 2018.

\bibitem{DBLP:conf/nips/JanzHMHHT19}
D.~Janz, J.~Hron, P.~Mazur, K.~Hofmann, J.~M. Hern{\'{a}}ndez{-}Lobato, and
  S.~Tschiatschek, ``Successor uncertainties: Exploration and uncertainty in
  temporal difference learning,'' in \emph{NeurIPS}, 2019.

\bibitem{DBLP:conf/nips/MetelliLR19}
A.~M. Metelli, A.~Likmeta, and M.~Restelli, ``Propagating uncertainty in
  reinforcement learning via wasserstein barycenters,'' in \emph{NeurIPS},
  2019.

\bibitem{ODonoghueOMM18}
B.~O'Donoghue, I.~Osband, R.~Munos, and V.~Mnih, ``The uncertainty bellman
  equation and exploration,'' in \emph{ICML}, 2018.

\bibitem{DBLP:conf/nips/OsbandBPR16}
I.~Osband, C.~Blundell, A.~Pritzel, and B.~V. Roy, ``Deep exploration via
  bootstrapped {DQN},'' in \emph{NeurIPS}, 2016.

\bibitem{DBLP:conf/nips/CiosekVLH19}
K.~Ciosek, Q.~Vuong, R.~Loftin, and K.~Hofmann, ``Better exploration with
  optimistic actor critic,'' in \emph{NeurIPS}, 2019.

\bibitem{kimin2020sunrise}
K.~Lee, M.~Laskin, A.~Srinivas, and P.~Abbeel, ``{SUNRISE:} {A} simple unified
  framework for ensemble learning in deep reinforcement learning,'' in
  \emph{ICML}, 2021.

\bibitem{OB2I-2021}
C.~Bai, L.~Wang, L.~Han, J.~Hao, A.~Garg, P.~Liu, and Z.~Wang, ``Principled
  exploration via optimistic bootstrapping and backward induction,'' in
  \emph{ICML}, 2021.

\bibitem{DBLP:journals/corr/abs-1711-10789}
T.~M. Moerland, J.~Broekens, and C.~M. Jonker, ``Efficient exploration with
  double uncertain value networks,'' \emph{arXiv preprint arXiv:1711.10789},
  2017.

\bibitem{DBLP:conf/colt/KirschnerK18}
J.~Kirschner and A.~Krause, ``Information directed sampling and bandits with
  heteroscedastic noise,'' in \emph{CoLT}, 2018.

\bibitem{DBLP:conf/icml/MavrinYKWY19}
B.~Mavrin, H.~Yao, L.~Kong, K.~Wu, and Y.~Yu, ``Distributional reinforcement
  learning for efficient exploration,'' in \emph{ICML}, 2019.

\bibitem{zhou2020non}
F.~Zhou, J.~Wang, and X.~Feng, ``Non-crossing quantile regression for
  distributional reinforcement learning,'' in \emph{NeurIPS}, 2020.

\bibitem{DeardenFR98}
R.~Dearden, N.~Friedman, and S.~J. Russell, ``Bayesian q-learning,'' in
  \emph{AAAI}, 1998.

\bibitem{auer2002finite}
P.~Auer, N.~Cesa-Bianchi, and P.~Fischer, ``Finite-time analysis of the
  multiarmed bandit problem,'' \emph{Machine learning}, vol.~47, no. 2-3, pp.
  235--256, 2002.

\bibitem{DBLP:conf/colt/Azizzadenesheli16}
K.~Azizzadenesheli, A.~Lazaric, and A.~Anandkumar, ``Reinforcement learning of
  pomdps using spectral methods,'' in \emph{CoLT}, 2016.

\bibitem{DBLP:journals/jmlr/Auer02}
P.~Auer, ``Using confidence bounds for exploitation-exploration trade-offs,''
  \emph{JMLR}, vol.~3, pp. 397--422, 2002.

\bibitem{DBLP:conf/nips/OsbandRR13}
I.~Osband, D.~Russo, and B.~V. Roy, ``(more) efficient reinforcement learning
  via posterior sampling,'' in \emph{NeurIPS}, 2013.

\bibitem{zanette2020frequentist}
A.~Zanette, D.~Brandfonbrener, E.~Brunskill, M.~Pirotta, and A.~Lazaric,
  ``Frequentist regret bounds for randomized least-squares value iteration,''
  in \emph{AISTATS}, 2020.

\bibitem{DBLP:journals/neco/Dayan93a}
P.~Dayan, ``Improving generalization for temporal difference learning: The
  successor representation,'' \emph{Neural Computation}, vol.~5, no.~4, pp.
  613--624, 1993.

\bibitem{DBLP:conf/nips/BarretoDMHSSH17}
A.~Barreto, W.~Dabney, R.~Munos, J.~J. Hunt, T.~Schaul, D.~Silver, and H.~van
  Hasselt, ``Successor features for transfer in reinforcement learning,'' in
  \emph{NeurIPS}, 2017.

\bibitem{DBLP:journals/siamma/AguehC11}
M.~Agueh and G.~Carlier, ``Barycenters in the wasserstein space,'' \emph{{SIAM}
  J. Math. Anal.}, vol.~43, no.~2, pp. 904--924, 2011.

\bibitem{DBLP:journals/corr/OsbandR15}
I.~Osband and B.~V. Roy, ``Bootstrapped thompson sampling and deep
  exploration,'' \emph{arXiv preprint arXiv:1507.00300}, 2015.

\bibitem{DBLP:conf/iclr/NikolovKBK19}
N.~Nikolov, J.~Kirschner, F.~Berkenkamp, and A.~Krause, ``Information-directed
  exploration for deep reinforcement learning,'' in \emph{ICLR}, 2019.

\bibitem{DBLP:journals/corr/abs-1905-09638}
W.~R. Clements, B.~Robaglia, B.~V. Delft, R.~B. Slaoui, and S.~Toth,
  ``Estimating risk and uncertainty in deep reinforcement learning,''
  \emph{arXiv preprint arXiv:1905.09638}, 2019.

\bibitem{GalG16}
Y.~Gal and Z.~Ghahramani, ``Dropout as a bayesian approximation: Representing
  model uncertainty in deep learning,'' in \emph{ICML}, 2016.

\bibitem{StadieLA15}
B.~C. Stadie, S.~Levine, and P.~Abbeel, ``Incentivizing exploration in
  reinforcement learning with deep predictive models,'' \emph{arXiv preprint
  arXiv:1507.00814}, 2015.

\bibitem{vdm-2020}
C.~Bai, P.~Liu, K.~Liu, L.~Wang, Y.~Zhao, L.~Han, and Z.~Wang, ``Variational
  dynamic for self-supervised exploration in deep reinforcement learning,''
  \emph{IEEE Transactions on Neural Networks and Learning Systems}, 2021.

\bibitem{OhC19}
C.~Oh and A.~Cavallaro, ``Learning action representations for self-supervised
  visual exploration,'' in \emph{ICRA}, 2019.

\bibitem{KimKJLS19}
H.~Kim, J.~Kim, Y.~Jeong, S.~Levine, and H.~O. Song, ``{EMI:} exploration with
  mutual information,'' in \emph{ICML}, 2019.

\bibitem{TangHFSCDSTA17}
H.~Tang, R.~Houthooft, D.~Foote, A.~Stooke, X.~Chen, Y.~Duan, J.~Schulman,
  F.~D. Turck, and P.~Abbeel, ``{\#}exploration: {A} study of count-based
  exploration for deep reinforcement learning,'' in \emph{NeurIPS}, 2017.

\bibitem{OstrovskiBOM17}
G.~Ostrovski, M.~G. Bellemare, A.~van~den Oord, and R.~Munos, ``Count-based
  exploration with neural density models,'' in \emph{ICML}, 2017.

\bibitem{MartinSEH17}
J.~Martin, S.~N. Sasikumar, T.~Everitt, and M.~Hutter, ``Count-based
  exploration in feature space for reinforcement learning,'' in \emph{IJCAI},
  2017.

\bibitem{abs-1905-12621}
G.~Vezzani, A.~Gupta, L.~Natale, and P.~Abbeel, ``Learning latent state
  representation for speeding up exploration,'' \emph{arXiv preprint
  arXiv:1905.12621}, 2019.

\bibitem{MachadoBB20}
M.~C. Machado, M.~G. Bellemare, and M.~Bowling, ``Count-based exploration with
  the successor representation,'' in \emph{AAAI}, 2020.

\bibitem{DBLP:conf/iclr/FoxCL18}
L.~Fox, L.~Choshen, and Y.~Loewenstein, ``{DORA} the explorer: Directed
  outreaching reinforcement action-selection,'' in \emph{ICLR}, 2018.

\bibitem{DBLP:conf/cig/SongCHF20}
Y.~Song, Y.~Chen, Y.~Hu, and C.~Fan, ``Exploring unknown states with action
  balance,'' in \emph{CIG}, 2020.

\bibitem{zhang2021made}
T.~Zhang, P.~Rashidinejad, J.~Jiao, Y.~Tian, J.~Gonzalez, and S.~Russell,
  ``Made: Exploration via maximizing deviation from explored regions,''
  \emph{arXiv preprint arXiv:2106.10268}, 2021.

\bibitem{DBLP:conf/nips/OhGLLS15}
J.~Oh, X.~Guo, H.~Lee, R.~L. Lewis, and S.~P. Singh, ``Action-conditional video
  prediction using deep networks in atari games,'' in \emph{NeurIPS}, 2015.

\bibitem{DBLP:conf/nips/FuCL17}
J.~Fu, J.~D. Co{-}Reyes, and S.~Levine, ``{EX2:} exploration with exemplar
  models for deep reinforcement learning,'' in \emph{NeurIPS}, 2017.

\bibitem{DBLP:journals/corr/abs-1903-07400}
J.~Zhang, N.~Wetzel, N.~Dorka, J.~Boedecker, and W.~Burgard, ``Scheduled
  intrinsic drive: {A} hierarchical take on intrinsically motivated
  exploration,'' \emph{arXiv preprint arXiv:1903.07400}, 2019.

\bibitem{klissarovvariational}
M.~Klissarov, R.~Islam, K.~Khetarpal, and D.~Precup, ``Variational state
  encoding as intrinsic motivation in reinforcement learning,'' in \emph{TARL
  Workshop on ICLR}, 2019.

\bibitem{DBLP:journals/corr/abs-1906-05274}
L.~Lee, B.~Eysenbach, E.~Parisotto, E.~P. Xing, S.~Levine, and
  R.~Salakhutdinov, ``Efficient exploration via state marginal matching,''
  \emph{arXiv preprint arXiv:1906.05274}, 2019.

\bibitem{stanton2018deep}
C.~Stanton and J.~Clune, ``Deep curiosity search: Intra-life exploration
  improves performance on challenging deep reinforcement learning problems,''
  \emph{arXiv preprint arXiv:1806.00553}, 2018.

\bibitem{DBLP:conf/nips/TaoFP20}
R.~Y. Tao, V.~Fran{\c{c}}ois{-}Lavet, and J.~Pineau, ``Novelty search in
  representational space for sample efficient exploration,'' in \emph{NeurIPS},
  2020.

\bibitem{DBLP:conf/iclr/SavinovRVMPLG19}
N.~Savinov, A.~Raichuk, D.~Vincent, R.~Marinier, M.~Pollefeys, T.~P. Lillicrap,
  and S.~Gelly, ``Episodic curiosity through reachability,'' in \emph{ICLR},
  2019.

\bibitem{zhang2021noveld}
T.~Zhang, H.~Xu, X.~Wang, Y.~Wu, K.~Keutzer, J.~E. Gonzalez, and Y.~Tian,
  ``Noveld: A simple yet effective exploration criterion,'' in \emph{Advances
  in Neural Information Processing Systems}, 2021.

\bibitem{DBLP:conf/nips/HouthooftCCDSTA16}
R.~Houthooft, X.~Chen, Y.~Duan, J.~Schulman, F.~D. Turck, and P.~Abbeel,
  ``{VIME:} variational information maximizing exploration,'' in
  \emph{NeurIPS}, 2016.

\bibitem{DBLP:journals/corr/AchiamS17}
J.~Achiam and S.~Sastry, ``Surprise-based intrinsic motivation for deep
  reinforcement learning,'' \emph{arXiv preprint arXiv:1703.01732}, 2017.

\bibitem{DBLP:conf/icml/PathakG019}
D.~Pathak, D.~Gandhi, and A.~Gupta, ``Self-supervised exploration via
  disagreement,'' in \emph{ICML}, 2019.

\bibitem{DBLP:conf/icml/ShyamJG19}
P.~Shyam, W.~Jaskowski, and F.~Gomez, ``Model-based active exploration,'' in
  \emph{ICML}, 2019.

\bibitem{OordKEKVG16}
A.~van~den Oord, N.~Kalchbrenner, L.~Espeholt, K.~Kavukcuoglu, O.~Vinyals, and
  A.~Graves, ``Conditional image generation with pixelcnn decoders,'' in
  \emph{NeurIPS}, 2016.

\bibitem{SchulmanWDRK17PPO}
J.~Schulman, F.~Wolski, P.~Dhariwal, A.~Radford, and O.~Klimov, ``Proximal
  policy optimization algorithms,'' \emph{arXiv preprint arXiv:1707.06347},
  2017.

\bibitem{DBLP:conf/iclr/AlemiFD017}
A.~A. Alemi, I.~Fischer, J.~V. Dillon, and K.~Murphy, ``Deep variational
  information bottleneck,'' in \emph{ICLR}, 2017.

\bibitem{DBLP:conf/nips/IttiB05}
L.~Itti and P.~Baldi, ``Bayesian surprise attracts human attention,'' in
  \emph{NeurIPS}, 2005.

\bibitem{DBLP:conf/nips/Graves11}
A.~Graves, ``Practical variational inference for neural networks,'' in
  \emph{NeurIPS}, 2011.

\bibitem{HorganQBBHHS18APEX}
D.~Horgan, J.~Quan, D.~Budden, G.~Barth{-}Maron, M.~Hessel, H.~van Hasselt, and
  D.~Silver, ``Distributed prioritized experience replay,'' in \emph{ICLR},
  2018.

\bibitem{Kapturowski19R2D2}
S.~Kapturowski, G.~Ostrovski, J.~Quan, R.~Munos, and W.~Dabney, ``Recurrent
  experience replay in distributed reinforcement learning,'' in \emph{ICLR},
  2019.

\bibitem{PlappertHDSC0AA18ParameterSpaceNoise}
M.~Plappert, R.~Houthooft, P.~Dhariwal, S.~Sidor, R.~Y. Chen, X.~Chen,
  T.~Asfour, P.~Abbeel, and M.~Andrychowicz, ``Parameter space noise for
  exploration,'' in \emph{ICLR}, 2018.

\bibitem{FortunatoAPMHOG18NoisyNet}
M.~Fortunato, M.~G. Azar, B.~Piot, J.~Menick, M.~Hessel, I.~Osband, A.~Graves,
  V.~Mnih, R.~Munos, D.~Hassabis, O.~Pietquin, C.~Blundell, and S.~Legg,
  ``Noisy networks for exploration,'' in \emph{ICLR}, 2018.

\bibitem{GarciaF15SRLsurvey}
J.~Garc{\'{\i}}a and F.~Fern{\'{a}}ndez, ``A comprehensive survey on safe
  reinforcement learning,'' \emph{Journal of Machine Learning Research},
  vol.~16, pp. 1437--1480, 2015.

\bibitem{AchiamHTA17CPO}
J.~Achiam, D.~Held, A.~Tamar, and P.~Abbeel, ``Constrained policy
  optimization,'' in \emph{{ICML}}, vol.~70.\hskip 1em plus 0.5em minus
  0.4em\relax {PMLR}, 2017, pp. 22--31.

\bibitem{SchulmanLAJM15TRPO}
J.~Schulman, S.~Levine, P.~Abbeel, M.~I. Jordan, and P.~Moritz, ``Trust region
  policy optimization,'' in \emph{ICML}, 2015.

\bibitem{TesslerMM19RewardCPO}
C.~Tessler, D.~J. Mankowitz, and S.~Mannor, ``Reward constrained policy
  optimization,'' in \emph{{ICLR}}, 2019.

\bibitem{StookeAA20ResponsiveSafety}
A.~Stooke, J.~Achiam, and P.~Abbeel, ``Responsive safety in reinforcement
  learning by {PID} lagrangian methods,'' in \emph{{ICML}}, vol. 119, 2020, pp.
  9133--9143.

\bibitem{BharadhwajKRLSG21Conser}
H.~Bharadhwaj, A.~Kumar, N.~Rhinehart, S.~Levine, F.~Shkurti, and A.~Garg,
  ``Conservative safety critics for exploration,'' in \emph{ICLR}, 2021.

\bibitem{Achiam2019BenchmarkingSE}
J.~Achiam and D.~Amodei, ``Benchmarking safe exploration in deep reinforcement
  learning,'' 2019.

\bibitem{HuntFMHDS21VSRL}
N.~Hunt, N.~Fulton, S.~Magliacane, T.~N. Hoang, S.~Das, and A.~Solar{-}Lezama,
  ``Verifiably safe exploration for end-to-end reinforcement learning,'' in
  \emph{{HSCC}}, 2021, pp. 14:1--14:11.

\bibitem{Dalal2018SafeECAS}
G.~Dalal, K.~Dvijotham, M.~Vecer{\'{\i}}k, T.~Hester, C.~Paduraru, and
  Y.~Tassa, ``Safe exploration in continuous action spaces,'' \emph{arXiv
  preprint arXiv:1801.08757}, 2018.

\bibitem{SaundersSSE18HumanInter}
W.~Saunders, G.~Sastry, A.~Stuhlm{\"{u}}ller, and O.~Evans, ``Trial without
  error: Towards safe reinforcement learning via human intervention,'' in
  \emph{{AAMAS}}, 2018, pp. 2067--2069.

\bibitem{ThananjeyanBRLM20SAVED}
B.~Thananjeyan, A.~Balakrishna, U.~Rosolia, F.~Li, R.~McAllister, J.~E.
  Gonzalez, S.~Levine, F.~Borrelli, and K.~Goldberg, ``Safety augmented value
  estimation from demonstrations {(SAVED):} safe deep model-based {RL} for
  sparse cost robotic tasks,'' \emph{{IEEE} Robotics and Automation Letters},
  vol.~5, no.~2, pp. 3612--3619, 2020.

\bibitem{ThomasLM21NearFuture}
G.~Thomas, Y.~Luo, and T.~Ma, ``Safe reinforcement learning by imagining the
  near future,'' in \emph{NeurIPS}, 2021, pp. 13\,859--13\,869.

\bibitem{TurchettaB016SafeEGP}
M.~Turchetta, F.~Berkenkamp, and A.~Krause, ``Safe exploration in finite markov
  decision processes with gaussian processes,'' in \emph{NeurIPS}, 2016, pp.
  4305--4313.

\bibitem{KarimpanalR00V20TransferableDP}
T.~G. Karimpanal, S.~Rana, S.~Gupta, T.~Tran, and S.~Venkatesh, ``Learning
  transferable domain priors for safe exploration in reinforcement learning,''
  in \emph{{IJCNN}}, 2020, pp. 1--10.

\bibitem{LiptonGLCD16CombatingIF}
Z.~C. Lipton, J.~Gao, L.~Li, J.~Chen, and L.~Deng, ``Combating reinforcement
  learning's sisyphean curse with intrinsic fear,'' \emph{arXiv preprint
  arXiv:1611.01211}, 2016.

\bibitem{FatemiSSK19DeadEnd}
M.~Fatemi, S.~Sharma, H.~van Seijen, and S.~E. Kahou, ``Dead-ends and secure
  exploration in reinforcement learning,'' in \emph{{ICML}}, vol.~97, 2019, pp.
  1873--1881.

\bibitem{Yu2022SafetyEditor}
H.~Yu, W.~Xu, and H.~Zhang, ``Towards safe reinforcement learning with a safety
  editor policy,'' \emph{arXiv preprint arXiv:2201.12427}, 2022.

\bibitem{BajcsyBBTT19Efficient}
A.~Bajcsy, S.~Bansal, E.~Bronstein, V.~Tolani, and C.~J. Tomlin, ``An efficient
  reachability-based framework for provably safe autonomous navigation in
  unknown environments,'' in \emph{{CDC}}, 2019, pp. 1758--1765.

\bibitem{HerbertBGT19Reach}
S.~L. Herbert, S.~Bansal, S.~Ghosh, and C.~J. Tomlin, ``Reachability-based
  safety guarantees using efficient initializations,'' in \emph{{CDC}}, 2019,
  pp. 4810--4816.

\bibitem{Ecoffet19GoExp}
A.~Ecoffet, J.~Huizinga, J.~Lehman, K.~O. Stanley, and J.~Clune, ``Go-explore:
  a new approach for hard-exploration problems,'' \emph{arXiv preprint
  arXiv:1901.10995}, 2019.

\bibitem{Ecoffet20FirstReturn}
E.~Adrien, H.~Joost, L.~Joel, S.~K. O, and C.~Jeff, ``First return, then
  explore,'' \emph{Nature}, vol. 590, no. 7847, pp. 580--586, 2021.

\bibitem{GuoCMFB0L20DTSIL}
Y.~Guo, J.~Choi, M.~Moczulski, S.~Feng, S.~Bengio, M.~Norouzi, and H.~Lee,
  ``Memory based trajectory-conditioned policies for learning from sparse
  rewards,'' in \emph{NeurIPS}, 2020.

\bibitem{ZhaoDZKLX20Potential}
E.~Zhao, S.~Deng, Y.~Zang, Y.~Kang, K.~Li, and J.~Xing, ``Potential driven
  reinforcement learning for hard exploration tasks,'' in \emph{IJCAI}, 2020.

\bibitem{BolanderA11}
T.~Bolander and M.~B. Andersen, ``Epistemic planning for single and multi-agent
  systems,'' \emph{JANCL}, vol.~21, no.~1, pp. 9--34, 2011.

\bibitem{maspwgpzhu}
Z.~Zhu, E.~Biyik, and D.~Sadigh, ``Multi-agent safe planning with gaussian
  processes,'' \emph{arXiv preprint arXiv:2008.04452}, 2020.

\bibitem{eeozsgzerosum}
C.~Martin and T.~Sandholm, ``Efficient exploration of zero-sum stochastic
  games,'' \emph{arXiv preprint arXiv:2002.10524}, 2020.

\bibitem{russo2018tutorial}
D.~J. Russo, B.~Van~Roy, A.~Kazerouni, I.~Osband, and Z.~Wen, ``A tutorial on
  thompson sampling,'' \emph{Foundations and Trends{\textregistered} in Machine
  Learning}, vol.~11, no.~1, pp. 1--96, 2018.

\bibitem{KaufmannCG12}
E.~Kaufmann, O.~Capp{\'{e}}, and A.~Garivier, ``On bayesian upper confidence
  bounds for bandit problems,'' in \emph{AISTATS}, 2012.

\bibitem{LyuA20}
X.~Lyu and C.~Amato, ``Likelihood quantile networks for coordinating
  multi-agent reinforcement learning,'' in \emph{AAMAS}, 2020.

\bibitem{DFAC}
W.~Sun, C.~Lee, and C.~Lee, ``{DFAC} framework: Factorizing the value function
  via quantile mixture for multi-agent distributional q-learning,'' in
  \emph{ICML}, 2021.

\bibitem{DQMIX}
J.~Zhao, M.~Yang, X.~Hu, W.~Zhou, and H.~Li, ``{DQMIX:} {A} distributional
  perspective on multi-agent reinforcement learning,'' \emph{CoRR}, vol.
  abs/2202.10134, 2022.

\bibitem{DBLP:journals/corr/abs-1906-02138}
W.~B{\"{o}}hmer, T.~Rashid, and S.~Whiteson, ``Exploration with unreliable
  intrinsic reward in multi-agent reinforcement learning,'' \emph{arXiv
  preprint arXiv:1906.02138}, 2019.

\bibitem{DBLP:journals/corr/abs-1905-12127}
S.~Iqbal and F.~Sha, ``Coordinated exploration via intrinsic rewards for
  multi-agent reinforcement learning,'' \emph{CoRR}, vol. abs/1905.12127, 2019.

\bibitem{came}
I.-J. Liu, U.~Jain, R.~A. Yeh, and A.~Schwing, ``Cooperative exploration for
  multi-agent deep reinforcement learning,'' in \emph{International Conference
  on Machine Learning}, 2021, pp. 6826--6836.

\bibitem{DBLP:conf/nips/ZhengCWHHCFGZ21}
L.~Zheng, J.~Chen, J.~Wang, J.~He, Y.~Hu, Y.~Chen, C.~Fan, Y.~Gao, and
  C.~Zhang, ``Episodic multi-agent reinforcement learning with curiosity-driven
  exploration,'' in \emph{NeurIPS}, M.~Ranzato, A.~Beygelzimer, Y.~N. Dauphin,
  P.~Liang, and J.~W. Vaughan, Eds., 2021, pp. 3757--3769.

\bibitem{DBLP:conf/nips/DuHFLDT19}
Y.~Du, L.~Han, M.~Fang, J.~Liu, T.~Dai, and D.~Tao, ``{LIIR:} learning
  individual intrinsic reward in multi-agent reinforcement learning,'' in
  \emph{NeurIPS}, 2019.

\bibitem{zheng2018learning}
Z.~Zheng, J.~Oh, and S.~Singh, ``On learning intrinsic rewards for policy
  gradient methods,'' in \emph{NeurIPS}, 2018.

\bibitem{zheng2019can}
Z.~Zheng, J.~Oh, M.~Hessel, Z.~Xu, M.~Kroiss, H.~van Hasselt, D.~Silver, and
  S.~Singh, ``What can learned intrinsic rewards capture?'' in \emph{ICML},
  2020.

\bibitem{DBLP:conf/icml/JaquesLHGOSLF19}
N.~Jaques, A.~Lazaridou, E.~Hughes, {\c{C}}.~G{\"{u}}l{\c{c}}ehre, P.~A.
  Ortega, D.~Strouse, J.~Z. Leibo, and N.~de~Freitas, ``Social influence as
  intrinsic motivation for multi-agent deep reinforcement learning,'' in
  \emph{ICML}, 2019.

\bibitem{DBLP:conf/iclr/ChitnisT0020}
R.~Chitnis, S.~Tulsiani, S.~Gupta, and A.~Gupta, ``Intrinsic motivation for
  encouraging synergistic behavior,'' in \emph{ICLR}, 2020.

\bibitem{DBLP:journals/aamas/CarmelM99}
D.~Carmel and S.~Markovitch, ``Exploration strategies for model-based learning
  in multi-agent systems: Exploration strategies,'' \emph{JAAMAS}, vol.~2,
  no.~2, pp. 141--172, 1999.

\bibitem{DBLP:conf/atal/ChalkiadakisB03}
G.~Chalkiadakis and C.~Boutilier, ``Coordination in multiagent reinforcement
  learning: a bayesian approach,'' in \emph{AAMAS}, 2003.

\bibitem{VerbeeckNPT05ESRL}
K.~Verbeeck, A.~Now{\'{e}}, M.~Peeters, and K.~Tuyls, ``Multi-agent
  reinforcement learning in stochastic single and multi-stage games,'' in
  \emph{ALAMAS}, 2005.

\bibitem{DBLP:conf/ijcai/ChakrabortyCDJ17}
M.~Chakraborty, K.~Y.~P. Chua, S.~Das, and B.~Juba, ``Coordinated versus
  decentralized exploration in multi-agent multi-armed bandits,'' in
  \emph{IJCAI}, 2017.

\bibitem{DBLP:conf/icml/DimakopoulouR18}
M.~Dimakopoulou and B.~V. Roy, ``Coordinated exploration in concurrent
  reinforcement learning,'' in \emph{ICML}, 2018.

\bibitem{DBLP:conf/nips/DimakopoulouOR18}
M.~Dimakopoulou, I.~Osband, and B.~V. Roy, ``Scalable coordinated exploration
  in concurrent reinforcement learning,'' in \emph{NeurIPS}, 2018.

\bibitem{DBLP:conf/atal/000220}
G.~Chen, ``A new framework for multi-agent reinforcement learning - centralized
  training and exploration with decentralized execution via policy
  distillation,'' in \emph{AAMAS}, 2020.

\bibitem{DBLP:conf/nips/MahajanRSW19}
A.~Mahajan, T.~Rashid, M.~Samvelyan, and S.~Whiteson, ``{MAVEN:} multi-agent
  variational exploration,'' in \emph{NeurIPS}, 2019.

\bibitem{OhGSL18SIL}
J.~Oh, Y.~Guo, S.~Singh, and H.~Lee, ``Self-imitation learning,'' in
  \emph{ICML}, ser. Proceedings of Machine Learning Research, vol.~80, 2018,
  pp. 3875--3884.

\bibitem{Salimans18Backward}
T.~Salimans and R.~Chen, ``Learning montezuma's revenge from a single
  demonstration,'' \emph{arXiv preprint arXiv:1812.03381}, 2018.

\bibitem{trott2019keeping}
A.~Trott, S.~Zheng, C.~Xiong, and R.~Socher, ``Keeping your distance: Solving
  sparse reward tasks using self-balancing shaped rewards,'' in \emph{NeurIPS},
  2019, pp. 10\,376--10\,386.

\bibitem{Taiga2020On}
A.~A. Taiga, W.~Fedus, M.~C. Machado, A.~Courville, and M.~G. Bellemare, ``On
  bonus based exploration methods in the arcade learning environment,'' in
  \emph{ICLR}, 2020.

\bibitem{smac}
M.~Samvelyan, T.~Rashid, C.~S. de~Witt, G.~Farquhar, N.~Nardelli, T.~G.~J.
  Rudner, C.~Hung, P.~H.~S. Torr, J.~N. Foerster, and S.~Whiteson, ``The
  starcraft multi-agent challenge,'' pp. 2186--2188, 2019.

\bibitem{FarquharGLWUS20Growing}
G.~Farquhar, L.~Gustafson, Z.~Lin, S.~Whiteson, N.~Usunier, and G.~Synnaeve,
  ``Growing action spaces,'' in \emph{ICML}, 2020.

\bibitem{Xiong18PDQN}
J.~Xiong, Q.~Wang, Z.~Yang, P.~Sun, L.~Han, Y.~Zheng, H.~Fu, T.~Zhang, J.~Liu,
  and H.~Liu, ``Parametrized deep q-networks learning: Reinforcement learning
  with discrete-continuous hybrid action space,'' \emph{arXiv preprint
  arXiv:1810.06394}, 2018.

\bibitem{FuTHLCF19MAHybrid}
H.~Fu, H.~Tang, J.~Hao, Z.~Lei, Y.~Chen, and C.~Fan, ``Deep multi-agent
  reinforcement learning with discrete-continuous hybrid action spaces,'' in
  \emph{IJCAI}, 2019.

\bibitem{LaskinSA20CURL}
M.~Laskin, A.~Srinivas, and P.~Abbeel, ``{CURL:} contrastive unsupervised
  representations for reinforcement learning,'' in \emph{{ICML}}, ser.
  Proceedings of Machine Learning Research, vol. 119, 2020, pp. 5639--5650.

\bibitem{YaratsFLP21ProtoRL}
D.~Yarats, R.~Fergus, A.~Lazaric, and L.~Pinto, ``Reinforcement learning with
  prototypical representations,'' in \emph{{ICML}}, ser. Proceedings of Machine
  Learning Research, vol. 139, 2021, pp. 11\,920--11\,931.

\bibitem{db-2021}
C.~Bai, L.~Wang, L.~Han, A.~Garg, J.~Hao, P.~Liu, and Z.~Wang, ``Dynamic
  bottleneck for robust self-supervised exploration,'' \emph{Advances in Neural
  Information Processing Systems}, vol.~34, 2021.

\bibitem{GhoshGL19Actionable}
D.~Ghosh, A.~Gupta, and S.~Levine, ``Learning actionable representations with
  goal conditioned policies,'' in \emph{{ICLR}}, 2019.

\bibitem{mirowski2018learning}
P.~Mirowski, M.~Grimes, M.~Malinowski, K.~M. Hermann, K.~Anderson,
  D.~Teplyashin, K.~Simonyan, A.~Zisserman, R.~Hadsell \emph{et~al.},
  ``Learning to navigate in cities without a map,'' in \emph{NeurIPS}, 2018.

\bibitem{savinov2018semi}
N.~Savinov, A.~Dosovitskiy, and V.~Koltun, ``Semi-parametric topological memory
  for navigation,'' in \emph{ICLR}, 2018.

\bibitem{VaswaniSPUJGKP17Transformer}
A.~Vaswani, N.~Shazeer, N.~Parmar, J.~Uszkoreit, L.~Jones, A.~N. Gomez,
  L.~Kaiser, and I.~Polosukhin, ``Attention is all you need,'' in
  \emph{Advances in Neural Information Processing Systems}, 2017, pp.
  5998--6008.

\bibitem{kim2020active}
K.~H. Kim, M.~Sano, J.~De~Freitas, N.~Haber, and D.~Yamins, ``Active world
  model learning in agent-rich environments with progress curiosity,'' in
  \emph{ICML}, 2020.

\bibitem{machado2018revisiting}
M.~C. Machado, M.~G. Bellemare, E.~Talvitie, J.~Veness, M.~Hausknecht, and
  M.~Bowling, ``Revisiting the arcade learning environment: Evaluation
  protocols and open problems for general agents,'' \emph{Journal of Artificial
  Intelligence Research}, vol.~61, pp. 523--562, 2018.

\bibitem{jin2018q}
C.~Jin, Z.~Allen-Zhu, S.~Bubeck, and M.~I. Jordan, ``Is q-learning provably
  efficient?'' in \emph{NeurIPS}, 2018.

\bibitem{OPPO-2020}
Q.~Cai, Z.~Yang, C.~Jin, and Z.~Wang, ``Provably efficient exploration in
  policy optimization,'' in \emph{ICML}, 2020.

\bibitem{jin2020provably}
C.~Jin, Z.~Yang, Z.~Wang, and M.~I. Jordan, ``Provably efficient reinforcement
  learning with linear function approximation,'' in \emph{CoLT}, 2020.

\bibitem{auer2009near}
P.~Auer, T.~Jaksch, and R.~Ortner, ``Near-optimal regret bounds for
  reinforcement learning,'' in \emph{NeurIPS}, 2009.

\bibitem{bound-2016}
T.~Lattimore, ``Regret analysis of the anytime optimally confident ucb
  algorithm,'' \emph{arXiv preprint arXiv:1603.08661}, 2016.

\bibitem{lipton2018bbq}
Z.~C. Lipton, X.~Li, J.~Gao, L.~Li, F.~Ahmed, and L.~Deng, ``Bbq-networks:
  Efficient exploration in deep reinforcement learning for task-oriented
  dialogue systems,'' in \emph{AAAI}, 2018.

\bibitem{tagasovska2018single}
N.~Tagasovska and D.~Lopez-Paz, ``Single-model uncertainties for deep
  learning,'' \emph{arXiv preprint arXiv:1811.00908}, 2018.

\bibitem{thudumu2020comprehensive}
S.~Thudumu, P.~Branch, J.~Jin, and J.~J. Singh, ``A comprehensive survey of
  anomaly detection techniques for high dimensional big data,'' \emph{Journal
  of Big Data}, vol.~7, no.~1, pp. 1--30, 2020.

\bibitem{ng1999policy}
A.~Y. Ng, D.~Harada, and S.~Russell, ``Policy invariance under reward
  transformations: Theory and application to reward shaping,'' in \emph{ICML},
  1999.

\bibitem{beyer2019mulex}
L.~Beyer, D.~Vincent, O.~Teboul, S.~Gelly, M.~Geist, and O.~Pietquin,
  ``{MULEX:} disentangling exploitation from exploration in deep {RL},''
  \emph{arXiv preprint arXiv:1907.00868}, 2019.

\bibitem{zhang2019learning}
Y.~Zhang, W.~Yu, and G.~Turk, ``Learning novel policies for tasks,'' in
  \emph{ICML}, 2019.

\bibitem{FoersterFANW18COMA}
J.~N. Foerster, G.~Farquhar, T.~Afouras, N.~Nardelli, and S.~Whiteson,
  ``Counterfactual multi-agent policy gradients,'' in \emph{AAAI}, 2018.

\bibitem{NguyenNVNDL20multiobj}
T.~T. Nguyen, N.~D. Nguyen, P.~Vamplew, S.~Nahavandi, R.~Dazeley, and C.~P.
  Lim, ``A multi-objective deep reinforcement learning framework,''
  \emph{EAAI}, vol.~96, p. 103915, 2020.

\bibitem{Bura21SEProvable}
A.~Bura, A.~HasanzadeZonuzy, D.~M. Kalathil, S.~Shakkottai, and J.~Chamberland,
  ``Safe exploration for constrained reinforcement learning with provable
  guarantees,'' \emph{arXiv preprint arXiv:2112.00885}, 2021.

\bibitem{CiosekFTHT20FittingPrior}
K.~Ciosek, V.~Fortuin, R.~Tomioka, K.~Hofmann, and R.~E. Turner, ``Conservative
  uncertainty estimation by fitting prior networks,'' in \emph{ICLR}, 2020.

\bibitem{nix1994estimating}
D.~A. Nix and A.~S. Weigend, ``Estimating the mean and variance of the target
  probability distribution,'' in \emph{ICNN}, vol.~1, 1994, pp. 55--60.

\bibitem{kalweit2017uncertainty}
G.~Kalweit and J.~Boedecker, ``Uncertainty-driven imagination for continuous
  deep reinforcement learning,'' in \emph{CoRL}, 2017, pp. 195--206.

\bibitem{sekar2020planning}
R.~Sekar, O.~Rybkin, K.~Daniilidis, P.~Abbeel, D.~Hafner, and D.~Pathak,
  ``Planning to explore via self-supervised world models,'' in \emph{ICML},
  2020.

\bibitem{hafner2019dream}
D.~Hafner, T.~Lillicrap, J.~Ba, and M.~Norouzi, ``Dream to control: Learning
  behaviors by latent imagination,'' in \emph{ICLR}, 2020.

\bibitem{pacchiano2020optimism}
A.~Pacchiano, P.~Ball, J.~Parker-Holder, K.~Choromanski, and S.~Roberts, ``On
  optimism in model-based reinforcement learning,'' \emph{arXiv preprint
  arXiv:2006.11911}, 2020.

\bibitem{agrawal2017posterior}
S.~Agrawal and R.~Jia, ``Posterior sampling for reinforcement learning:
  worst-case regret bounds,'' in \emph{Advances in Neural Information
  Processing Systems}, 2017, pp. 1184--1194.

\bibitem{curi2020efficient}
S.~Curi, F.~Berkenkamp, and A.~Krause, ``Efficient model-based reinforcement
  learning through optimistic policy search and planning,'' \emph{Advances in
  Neural Information Processing Systems}, vol.~33, 2020.

\bibitem{ball2020ready}
P.~Ball, J.~Parker-Holder, A.~Pacchiano, K.~Choromanski, and S.~Roberts,
  ``Ready policy one: World building through active learning,'' \emph{arXiv
  preprint arXiv:2002.02693}, 2020.

\bibitem{HER-2017}
M.~Andrychowicz, F.~Wolski, A.~Ray, J.~Schneider, R.~Fong, P.~Welinder,
  B.~McGrew, J.~Tobin, P.~Abbeel, and W.~Zaremba, ``Hindsight experience
  replay,'' in \emph{NeurIPS}, 2017.

\bibitem{HPG-2019}
P.~Rauber, F.~Mutz, and J.~Schmidhuber, ``Hindsight policy gradients,'' in
  \emph{ICLR}, 2019.

\bibitem{vrl-2018}
A.~V. Nair, V.~Pong, M.~Dalal, S.~Bahl, S.~Lin, and S.~Levine, ``Visual
  reinforcement learning with imagined goals,'' in \emph{NeurIPS}, 2018.

\bibitem{dher-2019}
M.~Fang, C.~Zhou, B.~Shi, B.~Gong, J.~Xu, and T.~Zhang, ``Dher: Hindsight
  experience replay for dynamic goals,'' in \emph{ICLR}, 2019.

\bibitem{cer-2019}
H.~Liu, A.~Trott, R.~Socher, and C.~Xiong, ``Competitive experience replay,''
  in \emph{ICLR}, 2019.

\bibitem{Selfplay-2018}
S.~Sukhbaatar, I.~Kostrikov, A.~Szlam, and R.~Fergus, ``Intrinsic motivation
  and automatic curricula via asymmetric self-play,'' in \emph{ICLR}, 2018.

\bibitem{mep-2019}
R.~Zhao, X.~Sun, and V.~Tresp, ``Maximum entropy-regularized multi-goal
  reinforcement learning,'' in \emph{ICML}, 2019.

\bibitem{curr-2019}
M.~Fang, T.~Zhou, Y.~Du, L.~Han, and Z.~Zhang, ``Curriculum-guided hindsight
  experience replay,'' in \emph{NeurIPS}, 2019.

\bibitem{HGG-2019}
Z.~Ren, K.~Dong, Y.~Zhou, Q.~Liu, and J.~Peng, ``Exploration via hindsight goal
  generation,'' in \emph{NeurIPS}, 2019.

\bibitem{guo2019directed}
Z.~D. Guo and E.~Brunskill, ``Directed exploration for reinforcement
  learning,'' \emph{arXiv preprint arXiv:1906.07805}, 2019.

\bibitem{salge2014empowerment}
C.~Salge, C.~Glackin, and D.~Polani, ``Empowerment--an introduction,'' in
  \emph{Guided Self-Organization: Inception}, 2014.

\bibitem{gregor2016variational}
K.~Gregor, D.~J. Rezende, and D.~Wierstra, ``Variational intrinsic control,''
  \emph{arXiv preprint arXiv:1611.07507}, 2016.

\bibitem{eysenbach2018diversity}
B.~Eysenbach, A.~Gupta, J.~Ibarz, and S.~Levine, ``Diversity is all you need:
  Learning skills without a reward function,'' in \emph{ICLR}, 2019.

\bibitem{achiam2018variational}
J.~Achiam, H.~Edwards, D.~Amodei, and P.~Abbeel, ``Variational option discovery
  algorithms,'' \emph{arXiv preprint arXiv:1807.10299}, 2018.

\bibitem{pong2019skew}
V.~H. Pong, M.~Dalal, S.~Lin, A.~Nair, S.~Bahl, and S.~Levine, ``Skew-fit:
  State-covering self-supervised reinforcement learning,'' in \emph{ICML},
  2020.

\bibitem{sharma2019dynamics}
A.~Sharma, S.~Gu, S.~Levine, V.~Kumar, and K.~Hausman, ``Dynamics-aware
  unsupervised discovery of skills,'' in \emph{ICLR}, 2020.

\bibitem{campos2020explore}
V.~Campos, A.~Trott, C.~Xiong, R.~Socher, X.~Giro-i Nieto, and J.~Torres,
  ``Explore, discover and learn: Unsupervised discovery of state-covering
  skills,'' in \emph{ICML}, 2020.

\bibitem{HaarnojaTAL17SQL}
T.~Haarnoja, H.~Tang, P.~Abbeel, and S.~Levine, ``Reinforcement learning with
  deep energy-based policies,'' in \emph{ICML}, ser. Proceedings of Machine
  Learning Research, vol.~70, 2017, pp. 1352--1361.

\bibitem{GangwaniLP19SIDP}
T.~Gangwani, Q.~Liu, and J.~Peng, ``Learning self-imitating diverse policies,''
  in \emph{ICLR}, 2019.

\bibitem{LiuRLP17SVPG}
Y.~Liu, P.~Ramachandran, Q.~Liu, and J.~Peng, ``Stein variational policy
  gradient,'' in \emph{UAI}, 2017.

\bibitem{Guo19ExpSIL}
Y.~Guo, J.~Choi, M.~Moczulski, S.~Bengio, M.~Norouzi, and H.~Lee, ``Efficient
  exploration with self-imitation learning via trajectory-conditioned policy,''
  \emph{CoRR}, vol. abs/1907.10247, 2019.

\bibitem{DAI20ESIL}
T.~Dai, H.~Liu, and A.~A. Bharath, ``Episodic self-imitation learning with
  hindsight,'' \emph{CoRR}, vol. abs/2011.13467, 2020.

\bibitem{Tang20SILQ}
Y.~Tang, ``Self-imitation learning via generalized lower bound q-learning,'' in
  \emph{NeurIPS}, 2020.

\bibitem{Chen20SILSR}
\BIBentryALTinterwordspacing
Z.~Chen and M.~Lin, ``Self-imitation learning in sparse reward settings,''
  \emph{CoRR}, vol. abs/2010.06962, 2020. [Online]. Available:
  \url{https://arxiv.org/abs/2010.06962}
\BIBentrySTDinterwordspacing

\bibitem{FerretPG21SIAL}
J.~Ferret, O.~Pietquin, and M.~Geist, ``Self-imitation advantage learning,'' in
  \emph{{AAMAS}}, 2021, pp. 501--509.

\bibitem{Ning21CoIL}
K.~Ning, H.~Xu, K.~Zhu, and S.~Huang, ``Co-imitation learning without expert
  demonstration,'' \emph{CoRR}, vol. abs/2103.14823, 2021.

\end{thebibliography}
\end{document}